\documentclass[11pt]{article}
\usepackage[utf8]{inputenc}
\usepackage[margin=1in]{geometry}

\usepackage{microtype}
\usepackage{graphicx}
\usepackage{subfigure}
\usepackage{booktabs} 
\usepackage{natbib}
\setcitestyle{open={(},close={)}}

\usepackage[colorlinks=true,citecolor=blue]{hyperref}

\usepackage{amsmath,amsthm,amsfonts,amssymb,mathdots,array,mathrsfs,bm,bbm,stmaryrd,graphicx,subfigure,xcolor}
\usepackage{algorithm,algorithmic}
\usepackage{breakcites}

\usepackage[T1]{fontenc}
\usepackage{enumerate}
\usepackage{inputenc}

\usepackage{graphicx} 
\usepackage{subfigure}

\usepackage{booktabs,balance}
\usepackage{rotating}
\usepackage{boldline}
\usepackage{makecell}
\usepackage{multirow}
\usepackage{balance}

\usepackage{tikz}

\newtheorem{theorem}{Theorem}[section]

\newtheorem{lemma}[theorem]{Lemma}

\usepackage{xcolor}         
\usepackage{amssymb, amsmath}
\usepackage{algorithm}
\usepackage{algorithmic}
\usepackage{graphicx}
\usepackage{subfigure}

\newcommand{\one}{\mathbf{1}}

\newcommand{\calA}{\mathcal{A}}

\newcommand{\bma}{\begin{matrix*}[r]}
	\newcommand{\ema}{\end{matrix*}}

\begin{document}

\title{\bf On Penalization in Stochastic Multi-armed Bandits}
\author{\vspace{0.5in}\\\textbf{Guanhua Fang} and \textbf{Ping Li} \\
Cognitive Computing Lab\\
Baidu Research\\
10900 NE 8th St. Bellevue, WA 98004, USA\\
  \texttt{\{fanggh2018,\ pingli98\}@gmail.com}\vspace{0.5in} \\
  \textbf{Gennady Samorodnitsky} \\
  Cornell University\\
  616 Thurston Ave. Ithaca, NY 14853, USA\\
  \texttt{gs18@cornell.edu}
}

\date{}
\maketitle

\begin{abstract}\vspace{0.3in}
\noindent\footnote{The work of Gennady Samorodnitsky was conducted as a consulting researcher at Baidu Research -- Bellevue.}We study an important variant of the stochastic multi-armed bandit (MAB) problem, which takes penalization into consideration. Instead of directly maximizing cumulative expected reward, we need to balance between the total reward and fairness level. In this paper, we present some new insights in MAB and formulate the problem in the penalization framework, where rigorous penalized regret can be well defined and more sophisticated regret analysis is possible. Under such a framework, we propose a hard-threshold UCB-like algorithm, which enjoys many merits including asymptotic fairness, nearly optimal regret, better tradeoff between reward and fairness. Both gap-dependent and gap-independent regret bounds have been established. Multiple insightful comments are given to illustrate the soundness of our theoretical analysis. Numerous experimental results corroborate the theory and show the superiority of our method over other existing methods.
\end{abstract}

\newpage

\section{Introduction}

The multi-armed bandit (MAB) problem is a classical framework for sequential decision-making in uncertain environments. Starting with the seminal work of~\citet{robbins1952some}, over the years, a significant body of work has been developed to address both theoretical aspects and practical applications of this problem.
In a traditional stochastic multi-armed bandit (MAB) problem~\citep{lai1985asymptotically, auer2002finite, vermorel2005multi,streeter2006asymptotically, bubeck2012regret, nishihara2016no, bhatt2022extreme}, a learner has access to $K$ arms and pulling arm $k$ generates a stochastic reward for the principal from an unknown distribution $F_k$ with an unknown expected reward $\mu_k$.
If the mean rewards were known as prior information, the learner could just repeatedly pull the best arm given by $k^{\ast} = \arg\max_k \mu_k$.
However, the learner has no such knowledge of the reward of each arm. Hence, one should use some learning algorithm $\pi$ which operates in rounds, pulls arm $\pi_t \in \{1,\ldots,K\}$ in round $t$, observes the stochastic reward generated from reward distribution $F_{\pi_t}$, and uses that information to learn the best arm over time. 
The performance of learning algorithm $\pi$ is evaluated based on its cumulative regret over time horizon $T$, defined as
\begin{eqnarray}
\bar R_{\pi}(T) = \mu_{k^{\ast}} T - \sum_{t = 1}^T \mathbb E\mu_{\pi_t}. 
\end{eqnarray}
To achieve the minimum regret, a good learner should make a balance between exploration (pulling different arms to get more information of reward distribution of each arm) and exploitation (pulling the arm currently believed to have the highest reward).

In addition to the classical MAB problems, many variations of the MAB framework have been extensively studied in the literature recently. 
Various papers study MAB problems with additional constraints which include bandits with knapsack constraints~\citep{badanidiyuru2013bandits}, bandits with budget constraints~\citep{xia2015thompson}, sleeping bandits~\citep{kleinberg2010regret, chatterjee2017analysis}, etc. 
Except these, there is a huge research interest in fairness within machine learning field.
Fairness has been a hot topic of many recent application tasks, including classification~\citep{zafar2017fairness2,zafar2017fairness,agarwal2018reductions, Roh21fairbatch}, regression~\citep{berk2017convex, rezaei2020fairness}, recommendation~\citep{celis2018ranking, singh2018fairness, beutel2019fairness, WangBSJ21Fairness}, resource allocation~\citep{baruah1996proportionate, talebi2018learning, li2020fair}, Markov decision process~\citep{khan2019s}, etc.
There are two popular definitions of fairness in the MAB literature. 
1).
The fairness is introduced into the bandit learning framework by saying that it is unfair to preferentially choose one arm over another if the chosen arm has lower expected reward than the unchosen arm~\citep{joseph2016fairness}.
In other words, the learning algorithm cannot favor low-reward arms.
2). The fairness is introduced such that the algorithm needs to ensure that uniformly (i.e., at the end of every round) each arm is pulled at least a pre-specified fraction of times~\citep{patil2020achieving}.
In other words, it imposes an additional constraint to prevent the algorithm from playing low-reward arms too few times.

Apart from above traditional definitions of fairness, we adopt a new perspective in this paper. That is, 
in addition to maximizing the cumulative expected reward, it also allows the user to specify how ``hard"  or how ``soft" the fairness requirement on each arm is.
We emphasize that we aim to seek a better trade-off between total reward and fairness requirement instead of meeting strict fairness constraints.   
This perspective is especially useful by considering the following applications. 
In finance, a company not only wants to maximize their profits but also to have a healthy market share~\citep{szymanski1993analysis, genchev2012effects}. 
But a realistic question is that the company cannot invest on every products. It needs to balance between profits and market share. 
In management, the supplier usually cannot meet the demand of every retailer. He/she needs to make the distribution in the most profitable way~\citep{yang2011solving, adida2014effect}. 
In this view, it is not always easy even to formulate the problem and to introduce an appropriate notion of regret. 
We thus propose a new formulation of fairness MAB by introducing penalty term $A_k \max(\tau_k T - N_k(T),0)$, where $A_k$, $\tau_k$ are the penalty rate and fairness fraction for arm $k$ and $N_k(T)$ is the number of times pulling arm $k$. 
Hence it gives penalization when the algorithm fails to meet the fairness constraint and penalty term is proportional to the gap between pulling number and its required level. 
To solve this regularized MAB problem, 
we also propose a hard-threshold upper confidence bound (UCB) algorithm. 
It is similar to the classical UCB algorithm but adds an additional term to encourage the learner to favor those arms whose pulling numbers are below the required level at each round.
The advantage of our approach is that it allows the user to distinguish , if desired, between arms for which is more important to sample an arm with required frequency and those arms for which it is less important to do so.

To the best of our knowledge, there is no work on mathematical framework of stochastic MAB with regularization term in the literature. In this paper, we provide a relatively complete theory for the penalized MAB. We rigorously formalize the penalized regret which can be used for evaluating the performance of learning algorithm under fairness constraints. 
On theoretical side, the hard-threshold UCB algorithm is proved to achieve asymptotic fairness when a large penalty rate is chosen.
The algorithm is shown to obtain $O(\log T)$ regret when the sub-optimality gap is assumed to be fixed. 
Additionally, the characterization of fluctuation of non-fairness level, $\max_{1 \leq t \leq T} \max(\tau_k t - N_k(t), 0)$ is also given. Its magnitude is also shown to be $O(\log T)$. 
Moreover, we establish a nearly-optimal gap-free regret bound of proposed method and provide insights on how hard-threshold based UCB index works. 
We also point out that the analysis of proposed hard-threshold UCB algorithm is much harder than the classical UCB due to the existence of interventions between different sub-optimal arms. 
On numerical side, the experimental results confirm our theory and show that the performance of the proposed algorithm is better than other popular methods. Our method achieves a better trade-off between reward and fairness.

\vspace{0.1in}
\noindent\textbf{Notations}.
For real number $x$, $(x)_{+}$ stands for $\max\{0, x\}$; $\lfloor x \rfloor$ is the largest integer smaller or equal to $x$.
For integer $n$, we use $[n]$ to represent the set $\{1, \ldots, n\}$.
We say $a = O(b)$; $a = \Omega(b)$ or $a = \Theta(b)$ if there exists a constant $C$ such that $a \leq C b$; $a \geq b / C$ or 
$b/C \leq a \leq C b$.
The symbols $\mathbb E$ and $\mathbb P(\cdot)$ denote generic expectation and probability under a probability measure that may be determined from the context. 
We let $\pi$ be a generic policy / learning algorithm.

\section{The Penalized Regret}

Consider a stochastic multi-armed bandit problem with $K$ arms and
unknown expected rewards $\mu_1,\ldots, \mu_K$ associated with these
arms. The notion of fairness we introduce consists of proportions
$\tau_k\geq 0, \,  k=1,\ldots, K$ with $\tau_1+\cdots +\tau_K <
1$. 
We use $T \in \{1,2, \ldots,\}$ to denote the time horizon and 
$N_{k,\pi}(t)$ to denote the number of times that arm $k$ has been pulled by time $t \in [T]$ using policy $\pi$.
For notational simplicity, we may write $N_{k,\pi}(t)$ as $N_k(t)$.
It is desired to pull arm $k$ at least at the uniform rate of $\tau_k$, $k=1,\ldots, K$. 
In other words, the learner should obey the constraint that 
$N_k(t) \geq \tau_k t$ for any $t \in [T]$.
Thus a good policy aims to solve the following optimization problem,  
\begin{eqnarray}\label{eqn:problem1}
 &&\arg\max_{\pi} \mathbb E\sum_k \mu_k N_{k,\pi}(T), \nonumber \\
 &&\text{subject to}~ N_{k,\pi}(t) \geq \tau_k t \ \text{for all $k$ and $t$.} 
\end{eqnarray}
Instead of directly working with such a constrained bandit problem, we consider a penalization problem. That is, one gets penalized if the arm is not pulled
sufficiently often. To reflect this, we introduce the following design 
problem.
Let $S_{\pi}(T)$ be the sum of the rewards obtained by time $t$ under policy $\pi$, i.e., $S_{\pi}(T) = \sum_{t=1}^T r_{\pi_t}$ where $\pi_t$ is the arm index chosen by policy $\pi$ at time $t \in [T]$ and $r_{\pi_t}$ is the corresponding reward.
Then the penalized total reward is defined as
\begin{equation} \label{e:penalized.total}
S_{\rm pen, \pi}(T) = S_{\pi}(T)-\sum_{k=1}^K A_k\bigl(\tau_kT-N_{k, \pi}(T)\bigr)_+,
\end{equation}
where $A_1,\ldots, A_K$ are known nonnegative penalty rates.
Our goal is to design a learning algorithm to make the expectation of $S_{\rm pen, \pi}(T) $ as large as possible. 
By taking the expectation, we have
\begin{eqnarray} 
 \mathbb E[S_{\rm pen, \pi}(T)] 
= \sum_{k=1}^K \mu_k \mathbb E[N_{k,\pi}(t)] -\sum_{k=1}^K A_k \mathbb E [\bigl(\tau_kT-N_{k,\pi}(T)\bigr)_+], \label{e:expected.penrev}
\end{eqnarray}
which is the penalized reward achieved by policy $\pi$ and we would like to maximize it over $\pi$. 
Now we are ready to introduce the \textit{penalized regret} function, which is the core for the regret analysis.

To derive the new regret, we first note that maximizing $\mathbb E[S_{\rm pen, \pi}(T)]$ is the same as minimizing the following loss function,
\begin{align} 
L(T)&=\mu^*T - \mathbb E[S_{\rm pen, \pi}(T)] \notag \\
&=\sum_{k=1}^K \Bigl[ \Delta_k \mathbb E[N_k(t)] +
A_k \mathbb E [\bigl(\tau_kT-N_k(T)\bigr)_+] \Bigr], \label{e:minus.rev}
\end{align} 
where we denote  
\[
\mu^*=\max_{k=1,\ldots, K}\mu_k, \ \Delta_k=\mu^*-\mu_k,\, k=1,\ldots,
K.
\]

In order to find the minimum possible value of $L(T)$, let us understand what a prophet
(who knows the expected rewards $\mu_1,\ldots, \mu_K$) would
do. Clearly, a prophet (who, in addition, is not constrained by integer value) would solve the following optimization problem,
\begin{align}
& \min_{x_1, \ldots, x_K} \sum_{k=1}^K \Bigl[ \Delta_k x_k +
A_k \mathbb E\bigl(\tau_kT-x_k\bigr)_+ \Bigr], \notag \\
& \text{subject to} ~  \sum_{k=1}^K x_k=T, \ x_k\geq 0, \, k=1,\ldots, K \notag,
\end{align}
and pull arm $k$ for $x_k$ times ($k=1,\ldots, K$). By denoting $y_k=x_k/T,\, k \in [K]$, we transform this
problem into
\begin{align} \label{e:opt.problem.norm}
 & \min_{y_1, \ldots, y_K} \sum_{k=1}^K \Bigl[ \Delta_k y_k +
A_k \bigl(\tau_k-y_k\bigr)_+ \Bigr], \notag \\
&\text{subject to} ~ \sum_{k=1}^K y_k=1, \ y_k\geq 0, \, k=1,\ldots, K.
\end{align}
We will solve the problem (\ref{e:opt.problem.norm}) by finding $y_1,\ldots, y_K$ that satisfy the constraints and that minimize simultaneously each term in the objective function.

\newpage

It is not hard to observe the following facts.
\begin{enumerate}
	\item For $A\geq 0$, function $y\mapsto A(\tau-y)_+$ achieves
	its minimum value of 0 for $y\geq\tau$.
	\item For $A\geq \Delta>0$, function $y\mapsto \Delta
	y+A(\tau-y)_+$ achieves its minimum of $\Delta\tau$ at
	$y=\tau$.
	\item For $\Delta>A\geq 0$, function $y\mapsto \Delta
	y+A(\tau-y)_+$ achieves its minimum of $A\tau$ at 
	$y=0$.
\end{enumerate}
As a result, we introduce the following three sets
\begin{align*}
\calA_{\rm opt} &= \bigl\{ k \in [K]:\, \mu_k=\mu^*\bigr\}, \\
\calA_{\rm cr} &= \bigl\{ k \in [K]:\,  A_k\geq \Delta_k>0\bigr\}, \\
\calA_{\rm non-cr}& = \bigl\{ k \in [K]:\,    \Delta_k>A_k\bigr\},
\end{align*}
where $\calA_{\rm opt}$ consists of all optimal arms, $\calA_{\rm cr}$ consists of sub-optimal arms with penalty rate larger than (or equal to) the sub-optimal gap and $\calA_{\rm non-cr}$ includes sub-optimal arms with penalty rate smaller than the sub-optimal gap.
Therefore an optimal solution to the problem  (\ref{e:opt.problem.norm})
can be constructed as follows. Let $k^*$ be an arbitrary arm in
$\calA_{\rm opt}$, and choose 
\begin{equation} \label{e:opt.sol}
y_k=\left\{
\begin{array}{l}
1-\sum_{j\in \calA_{\rm cr} \cup (\calA_{\rm opt}\setminus \{k^*\})} \tau_j, \quad k=k^*, \\ 
\tau_k,  \quad k\in \calA_{\rm cr} \cup (\calA_{\rm opt}\setminus \{k^*\}), \\
0, \quad k\in \calA_{\rm non-cr}.
\end{array}
\right.
\end{equation}

Hence a prophet would choose (modulo rounding) in
(\ref{e:minus.rev})
\begin{equation} \label{e:opt.policy}
N_k(T)=\left\{
\begin{array}{l}
\left(1-\sum_{j\in \calA_{\rm cr} \cup (\calA_{\rm opt}\setminus \{k^*\})} \tau_j\right)T, \quad k =k^*,\\
\tau_kT, \quad k\in \calA_{\rm cr}\cup (\calA_{\rm opt}\setminus \{k^*\}), \\
0, \quad k\in \calA_{\rm non-cr}, 
\end{array}
\right.
\end{equation}
leading to the following optimal value of $L(T)$,
\begin{align} 
L^*(T) &=  \sum_{k\in \calA_{\rm cr} } \Delta_k\tau_kT +\sum_{k\in
	\calA_{\rm non- cr} } A_k\tau_kT \notag \\
&= \left(\sum_{k=1}^K \min(\Delta_k,A_k) \tau_k \right)T. \label{e:optimal.penrev}
\end{align}

Given an arbitrary algorithm $\pi$, we can therefore define the penalized regret as
\begin{align} \label{e:regret}
& R_\pi(T) 
=  L_\pi(T)-L^*(T) \notag \\
& = \sum_{k\in \calA_{\rm opt} }  A_k
\mathbb E\bigl( \tau_kT-N_{k,\pi}(T)\bigr)_+  \notag  \\
+& \sum_{k\in \calA_{\rm cr} }  \Bigl[ \Delta_k \mathbb E \bigl(
N_{k, \pi}(T)-\tau_kT\bigr) +  A_k
\mathbb E\bigl( \tau_kT-N_{k,\pi}(T)\bigr)_+\Bigr] \notag \\
+& \sum_{k\in \calA_{\rm non-cr} }  \Bigl[  \Delta_k \mathbb E N_{k, \pi}(T) +
A_k\Bigl(\mathbb E \bigl( \tau_kT - N_{k,\pi}(T)\bigr)_+
-\tau_kT\Bigr)\Bigr].
\end{align}

\newpage

\section{A Hard-Threshold UCB Algorithm} \label{sec:UCB}

We now introduce a UCB-like algorithm which aims to achieve the minimum penalized regret described in the previous section. We assume that all rewards take values in the
interval $[0,1]$. 
We denote by $X_n^{(k)}$ the reward obtained after pulling arm $k$ for
the $n$th time, $k \in [K]$, $n=1,2,\ldots$. Let
\begin{equation} 
\hat m_k(n)= \frac{1}{N_n(k)}\sum_{i=1}^{N_n(k)} X_i^{(k)}, \ k=1,\ldots,
K, \ n=1,2,\ldots, \nonumber
\end{equation}
and introduce the following index: for $\ k=1,\ldots,
K, \ n=1,2,\ldots$ set 
\begin{equation} \label{e:index}
i_k(n) = \hat m_k(n-1)+ A_k\one \bigl( N_k(n-1)<\tau_kn\bigr) +
\sqrt{ \frac{2\log n}{N_k(n-1)}}. 
\end{equation} 
The algorithm proceeds as follows. It starts by pulling each arm
once. Then at each subsequent step, we pull an arm with the highest value
of the index $ i_k(n)$. 
In equation \ref{e:index}, there is an additional term
$A_k \mathbf 1(N_k(n-1) < \tau_k n)$ compared with classical UCB algorithm.
It takes the hard threshold form. Once the number of times that arm $k$ has been pulled before time $n$ is less than the \textit{fairness level} ($\tau_k n$) at round $n$, penalty rate $A_k$ will be added to the UCB index. 
In other words, the proposed algorithm favors those arms which does not meet the fairness requirement.  
The detailed implementation is given in Algorithm \ref{alg:proposed}.

\vspace{0.2in}

\begin{algorithm}[ht!]
	\caption{Hard-Threshold UCB Algorithm.}
	\label{alg:proposed}
	\begin{algorithmic}[1]
		\STATE {\bfseries Input.}\  
		Number of arms $K$, fairness proportions $\tau_k$'s, penalty rates $A_k$'s, time horizon $T$.
		\STATE {\bfseries Output.}\  
		Cumulative reward, the number of times that each arm is played ($N_k(T)$, $k \in \{1, \ldots, K\}$.)
		\STATE {\bfseries Initialization.}\  \\
		For each $k \in \{1, \ldots, K\}$, we set initial count
		$N_k = 0$ and arm-specific cumulative reward $R_k = 0$. 
		\WHILE{$n \leq T$}
		\STATE If $n \leq K$, we choose $k_n = n$.
		\STATE If $n > K$, we choose $k_n = \arg\max_k i_k(n)$.
		\STATE We observe reward $r_n$. We update count $N_{k_n} = N_{k_n} + 1$
		and update reward $R_{k_n} = R_{k_n} + r_n$. 
		\STATE We update hard-threshold index for each arm $k \in \{1,\ldots,K\}$ by calculating
		\begin{align}
		 i_k(n+1) 
		= R_k / N_k + A_k \mathbf 1(N_k < \tau_k (n+1)) + \sqrt{\frac{2 \log n}{N_k}}.  \nonumber 
		\end{align}
		\STATE Increase time index $n$ by 1.
		\ENDWHILE
		\STATE Return vector $(N_k)$.
	\end{algorithmic}
\end{algorithm}

\newpage

We compare the proposed methods with related existing methods.

\noindent \textit{Learning with Fairness Guarantee (LFG,~\citep{li2019combinatorial})}.
It is implemented via following steps.
\begin{itemize}
	\item For each round $n$, we compute the index for each arm,
	$\bar i_k(n) = \min \{\hat m_k(n-1) + \sqrt{\frac{2 \log n}{N_k(n-1)}}, 1\}$
	and compute queue length for each arm,
	$Q_k(n) = \max\{Q_k(n-1) + \tau_k - \mathbf 1\{\textrm{arm $k$ is pulled}\}, 0 \}$.
	\item The learner plays the arm which maximizes
	$Q_k(n) + \eta_0 w_k i_k(n)$
	and receives the corresponding reward, where $\eta_0$ is the tuning parameter and $w_k$ is the known weight. Without loss of generality, we assume $w_k \equiv 1$ by treating each arm equally when we have no additional information.
\end{itemize}

\noindent \textit{Fair-Learn (Flearn,~\citep{patil2020achieving})}.
Its main procedure is given as below.
\begin{itemize}
	\item For each round $n$, we compute set $\mathcal A(n)$,
	$\mathcal A(n) := \{k: \tau_k (n - 1) - N_k(n-1) > \alpha\}$,
	which contains those arms which are not fair at round $n$ at level.
	
	\item If $\mathcal A(n) \neq \emptyset$, we play arm which maximizes $\tau_k (n-1) - N_k(n-1)$.
	Otherwise, we play arm which maximizes 
	$\hat m_k(n-1) + \sqrt{\frac{2 \log n}{N_k(n-1)}}$.
\end{itemize}

Fair-learn method can enforce each arm $k$ should be played at proportion level $\tau_k$ only when $\alpha = 0$. 
LFG method does not guarantee the asymptotic fairness when $\eta_0 > 0$. 
Neither of these methods can well balance between total rewards and fairness constraint as our method does. 
See experimental section for more explanations.

\section{Theoretical Analysis of the Hard-threshold UCB algorithm} \label{sec:theory}

In this section, we present theoretical results for the  hard-threshold UCB algorithm introduced in Section \ref{sec:UCB}. Throughout this section, we need to introduce additional notation and concepts.
We say $\tau_k = \tilde \Omega(1)$ if it is a positive constant which is independent of $T$. 
We assume that there exists a positive constant $c_0$ such that $\sum_k \tau_k \leq 1 - c_0$ and $T$ is much larger than $K$.
The penalty rates $A_k$'s are assumed to be known fixed constants.  
The expected reward $\mu_k$ ($k \in [K]$) is assumed between 0 and 1. Hence sub-optimality gap $\Delta_k$ is between 0 and 1 as well.

\vspace{0.1in}
\noindent \textbf{Asymptotic Fairness}.
Given the large penalty rates, the proposed algorithm can guarantee the asymptotic fairness for any arm $k \in [K]$. 
In other words, the algorithm can guarantee that the number of times that arm $k$ has been pulled up to time $T$ is at least $\lfloor \tau_k T\rfloor$ with high probability.
\begin{theorem}\label{thm:fairness}
	If $A_k - \Delta_k \geq \min\{\sqrt{\frac{32 \log T}{\tau_k T}},1\}$ and $\tau_k = \tilde \Omega(1)$ for all $k$, we have $N_k(T) \geq \lfloor \tau_k T \rfloor $ for any $k$ with probability going to 1 as $T \rightarrow \infty$.
\end{theorem}
Theorem \ref{thm:fairness} tells us that the proposed algorithm treats every arm fairly when the penalty rate dominates the sub-optimality gap.  
In other words, when the penalization rate is large enough, it is desired to meet the fairness requirement rather than exploiting the arm with the highest reward.

\subsection{Regret Analysis: Upper Bounds}

In this section, we provide upper bounds on the penalized regret defined in \eqref{e:regret} under two scenarios. (i) We establish the gap-dependent bound when the sub-optimality $\Delta_k$'s are fixed constants. (ii) We prove the gap-independent bound when $\Delta_k$'s vary within the interval $[0,1]$.  

\newpage

\begin{theorem}\label{thm:dep:upper} (\textit{Gap-dependent Upper bound}.)
	Assume that 
	\begin{itemize}
	    \item[(i)] $A_{k} - \Delta_{k} \geq c_a$ ($c_a$ is some fixed positive constant such that $c_a > \sqrt{16 c_0^2 K \log T / T}$) holds for any arm $k \in \mathcal A_{opt} \cup \mathcal A_{\text{cr}}$;
	    \item[(ii)] $c_a /4 > \Delta_{k} - A_k \geq \min\{\sqrt{\frac{8 K \log T}{c_0^2 T}}, \sqrt{\frac{\log (c_0 T)}{\tau_k c_0 T}} \}$ holds for any $k \in \mathcal A_{\text{non-cr}}$.
	\end{itemize}
	 We then have the following results. 
	
	\begin{itemize}
	    \item[a.] 
	    	For any $k \in \mathcal A_{\text{opt}} \cup \mathcal A_{\text{cr}}$, it holds
	$\mathbb E[ (\tau_{k} T - N_{k}(T))_{+} ] = O(1)$.
	    \item[b.] 
	    For any $k \in \mathcal A_{\text{cr}}$, it holds
	$\mathbb E [N_{k}(T)] \leq \max\{\frac{8 \log T}{\Delta_{k}^2}, \tau_{k} T\} + O(1)$.
	    \item[c.] 
	    For any arm $k \in \mathcal A_{\text{non-cr}}$, it holds 
	$ \mathbb E[N_{k}(T)] \leq \max\{\min\{ \frac{8 \log T}{(\Delta_{k} - A_{k})^2}, \tau_{k} T \}, \frac{8 \log T}{\Delta_k^2}\} + O(1)$. 
	\item[d.]
	The regret is bounded via
	\begin{align}\label{eq:dep:upper}
	 R_{\pi}(T) 
	\leq& \sum_{k \in \mathcal A_{\text{non-cr}}} \max\{\min\{\frac{8 \log T}{\Delta_k - A_k}, (\Delta_k - A_k) \tau_k T \},\frac{8 \log T}{\Delta_k}\} \notag \\
	& + \sum_{k \in \mathcal A_{\text{cr}}} (\frac{8 \log T}{\Delta_k} - \tau_k T)_{+} + O(K).
	\end{align}
	\end{itemize}
\end{theorem}

Here we summarize some important implications from Theorem \ref{thm:dep:upper}.
It tells us that the number of times that each arm $k$ in critical set $\mathcal A_{\text{cr}}$ is played is at least around  fairness requirement $\tau_k T$ when the  penalty rate is larger than the sub-optimality gap by some constant. 
On the other hand, for each arm $k$ in non-critical set $\mathcal A_{\text{non-cr}}$, it could be played less than fairness requirement when sub-optimality gap substantially dominates the penalty rate. 
The total penalized regret has order of $\log T$ and is hence nearly not improvable.
In particular, if we set penalty rate of each arm to be equal (i.e., $A_k \equiv A$), then our theorem implies that we can always choose a suitable tuning parameter $A$ such that the algorithm guarantees the fairness of top-$k$ best arms once there is a positive gap between $\Delta_{(k)}$ and $\Delta_{(k+1)}$
(where $\Delta_{(k)}$ is the $k$-th element of $\{\Delta_k\}$'s in the ascending order.)
In addition, when $A_k \equiv 0$ and it degenerates to the classical settings, then all arms become non-critical arms and our bound reduces to $O(\sum_k \frac{8 \log T}{\Delta_k})$ which matches the existing result~\citep{auer2002finite}.
This explains the optimality of inequality \eqref{eq:dep:upper}.

\vspace{0.1in}
\noindent \textbf{Maximal Inequality}.
In Theorem \ref{thm:dep:upper} above we have shown that 
$\mathbb E[(\tau_k T - N_k(T))_{+}] = O(1)$ for any $k \in \mathcal A_{\text{opt}} \cup \mathcal A_{\text{cr}}$ under mild conditions on $\Delta_k$'s. In the result below, we derive a maximal inequality for the \textit{non-fairness level}, $(\tau_k t - N_k(t))_{+}$, $t\in [T]$.

\vspace{0.1in}
\noindent
\begin{theorem}\label{thm:maximal}
	Order the $K$ arms in such a way that 
	\[A_{k_1} + \mu_{k_1} \geq \ldots \geq A_{k_j} + \mu_{k_j} \geq \ldots \geq A_{k_K} + \mu_{k_K}.\]
	Then for any arm $k_j \in \mathcal A_{opt} \cup \mathcal A_{cr}$, we have
	\[ \mathbb E[\max_{1 \leq t \leq T}(\tau_{k_j}t - N_{k_j}(t))_{+}] \leq a_j \log T + O(1),\]
	where the coefficient $a_j$ is defined as
	\begin{eqnarray} 
	a_j &=& 8\sum_{d=1}^j (j-d+1) \big\{ \sum_{m=1}^{d-1} \frac{1}{(\mu_{k_d}+A_{k_d}-\mu_{k_m})^2} \nonumber \\ 
	&&+ \sum_{m=d+1}^K
	\frac{1}{(\mu_{k_d}+A_{k_d}-\mu_{k_m}-A_{k_m})^2}\big\}. 
	\end{eqnarray} 
\end{theorem}
Theorem \ref{thm:maximal} guarantees the almost any-round fairness for all arms $k \in \mathcal A_{\text{opt}} \cup \mathcal A_{\text{cr}}$ up to a $O(\log T)$ difference.
Therefore, our method can be also treated as a good solution to classical fairness MABs with paying only $O(\log T)$ prices.

\vspace{0.1in}
\noindent\textbf{Gap-independent Upper bound}.
We now switch to establishing a gap-independent upper bound. The key challenge lies in handling the term $\mathbb E[(\tau_k T - N_k(T))_{+}]$ for $k \in \mathcal A_{\text{opt}} \cup \mathcal A_{\text{cr}}$.
It is easy to see that 
$(A_k - \Delta_k) \mathbb E[(\tau_k T - N_k(T))_{+}]$ can be trivially lower bounded by zero. 
The question is how sharp upper bound we can derive for this term.
The solution relies on the following observations.

\begin{itemize}
    \item
    \noindent\textit{Observation 1}.~

If $A_k - \Delta_k \leq \sqrt{\frac{32 K \log T}{T}}$, then $(A_k - \Delta_k) \mathbb E[(\tau_k T - N_k(T))_{+}] \leq \tau_k \sqrt{32 K T \log T}$.
    \item 
    \noindent\textit{Observation 2}.~
    \begin{lemma}\label{lem:independent}
	If arm $k$ satisfies that $A_k - \Delta_k \geq  \sqrt{\frac{32 K \log T}{T}}$ and 
	$\tau_k = \tilde \Omega(1)$,
	then we have \[\mathbb E[(\tau_k T - N_k(T))_{+}] = O(\tau_k K \frac{32 \log T}{(A_k - \Delta_k)^2}).\]
\end{lemma}
\end{itemize}

Based on above two observations, we have the following regret bound which is free of $\Delta_k$'s. 
\begin{theorem}\label{thm:indep:upper}
	When $\tau_k = \tilde \Omega(1)$, it holds that 
	\begin{align}\label{eq:indep:uppper}
	R_{\pi}(T) &\leq  8 \sqrt{T \log T}(\sum_k \sqrt{\tau_k}) + 8 \sqrt{(1 - \tau_{min}) K T \log T}  \notag \\
	& + C \sqrt{K T \log T},
	\end{align}
	where $\tau_{min} = \min_k \tau_k$ and $C$ is a universal constant.
\end{theorem}
The first term in (\ref{eq:indep:uppper}) is for 
$A_k(\mathbb E(\tau_k T - N_k(T))_{+} )$ with $k \in \mathcal A_{\text{non-cr}}$. The second term gives a bound for 
$\Delta_k \mathbb E(N_k(T) - \tau_k T)$ for $k \in [K]$.
The third term in (\ref{eq:indep:uppper}) is for bounding $(A_k - \Delta_k) \mathbb E(\tau_k T - N_k(T))_{+}$ for $k \in \mathcal A_{\text{opt}} \cup \mathcal A_{\text{cr}}$.

\vspace{0.1in}
\noindent\textbf{Technical Challenges}.
Since the index $i_k(n)$ is a discontinuous function of $N_k$, this brings additional difficulties in analyzing the regret bound. 
The most distinguished feature from the classical regret analysis is that we cannot analyze term $N_k(T)$ separately for each sub-optimal gap $k$. 
In fact, the optimal arm ($\arg\max_k \mu_k$) is fixed for all rounds in the classical setting. In contrast, the ``optimal arm" ($\arg\max_k \mu_k + A_k \mathbf 1\{N_k < \tau_k n\}$) varies as the algorithm progresses in our framework.
Due to such interventions among different arms, term $(\tau_k T - N_k(T))_{+}$ should be treated carefully.

\subsection{Regret Analysis: Continued}

\noindent \textbf{Tightness of Gap-dependent Upper Bound}. 

\noindent In this part, we first show that the bound given in inequality (\ref{eq:dep:upper}) is tight. To see this, the results are stated in the following theorems. 

\newpage

\begin{theorem}\label{thm:lower:dep1}
	There exists a bandit setting for which the regret of proposed algorithm has the following lower bound,
	$
	R_{\pi}(T) \geq \sum_{k \in \mathcal A_{\text{non-cr}},\tau_k>0} \frac{\log T}{\Delta_k - A_k}$.
\end{theorem}

\medskip
\begin{theorem}\label{thm:lower:dep2}
	There exists a bandit setting for which the regret of proposed algorithm has the following lower bound,
	$
	R_{\pi}(T) \geq \sum_{k \in \mathcal A_{\text{cr}}} \Delta_k (\frac{\log T}{\Delta_k^2} - \tau_k T)$.
\end{theorem}

Theorem \ref{thm:lower:dep1} says that the term $\log T / (\Delta_k - A_k)$ is nearly optimal up a multiplicative constant 8 for any arm in the non-critical set. Similarly, Theorem \ref{thm:lower:dep2} tells us that $(\frac{8 \log T}{\Delta_k} - \tau_k T)_{+}$ is also nearly optimal for arms in the critical set. Therefore, Theorem \ref{thm:dep:upper} gives a relatively sharp gap-dependent upper bound.
It is almost impossible to improve the regret bound analysis for our proposed hard-threshold UCB algorithm in the instance-dependent scenario.

\vspace{0.1in}
\noindent\textbf{Gap-independent Lower Bound}. \noindent We also obtain a gap-independent lower bound as follows. 
\begin{theorem}\label{thm:indep:lower}
	Let $K > 1$ and $T$ be a large integer. Penalty rates $A_1, A_2, \ldots, A_K$ are fixed positive constants. Assume that the 
	fairness parameters $\tau_1, \ldots, \tau_K \in [0,1]$ with $\sum_k \tau_k < 1$.
	Then, for any policy $\pi$, there exists a mean vector 
	$\boldsymbol \mu = (\mu_1, \ldots, \mu_K)$ such that 
	\[R_{\pi}(T) \geq C (1 - \sum_{k}\tau_{k}) \sqrt{(K - 1) T},\]
	where $C$ is a universal constant which is free of
	$A_k$, $\tau_k$'s.
\end{theorem}

By comparing Theorems \ref{thm:indep:upper} and \ref{thm:indep:lower}, 
the orders of gap-independent upper and lower bounds match with each other if we ignore a $\log T$ factor. 
This indicates that our hard-threshold UCB algorithm is nearly optimal even without knowledge of sub-optimality gaps.

\section{Comments on the Hard-Threshold UCB Algorithm}\label{sec:comment}

\noindent \textbf{On hard threshold}.
In the proposed algorithm, we use a hard-threshold term $A_k \mathbf 1(N_k(n-1) < \tau_k n)$ in constructing a  UCB-like index $i_k(n)$. 
A natural question is whether we can use a soft-threshold index by defining
\begin{align}
\tilde i_k(n) &= \hat m_k(n-1)+ A_k \frac{\max ( \tau_kn - N_k(n-1), 0 )}{ \tau_k n}+
\sqrt{ \frac{2\log n}{N_k(n-1)}}? \nonumber 
\end{align}
The answer is negative in the sense that $\tilde i_k(n)$ becomes a continuous function of $N_k$ and does not have a jump point at the critical value $\tau_k n$. 
Hence term $\frac{\max ( \tau_kn - N_k(n-1), 0 )}{ \tau_k n}$ does not give sufficient penalization to those arms $k$ which are below the fairness proportion $\tau_k$. 
Hence, a soft-threshold UCB-like index fails to guarantee the asymptotic fairness and nearly-optimal penalized regret. 

\vspace{0.1in}
\noindent\textbf{Comparison with ETC method}.~
Explore-Then-Commit method (ETC) is one of popular algorithm in MAB literature~\citep{perchet2016batched, jin2021double}.
It consists of an exploration phase followed by an exploitation phase
It has been shown in~\citet{garivier2016explore} that ETC is suboptimal in the asymptotic sense as the horizon grows, and thus, is worse than fully sequential strategies (e.g., UCB-type methods).
More specifically, the issue of ETC is that the performance relies on the choice of length of exploration phase. For instance, we denote this length as $m$. Then optimal $m$ should be both $\Delta$- and $T$-dependent. However, we have no knowledge of sub-optimality gap $\Delta$ in practice. Thus, we can only choose $m$ depending on $T$. This  leads to the worst case bound of order $O(T^{2/3})$, which is worse than $O(T^{1/2})$. The latter  is achieved by our proposed method.

\vspace{0.1in}
\noindent\textbf{When $\tau_k$ is not constant}.  
In our theoretical analysis, we only consider the case that $\tau_k = \tilde \Omega(1)$ for ease of presentation. 
With slight modifications of proof, the current results could also apply when threshold $\tau_k$ is dependent on time horizon $T$ with $\tau_k(T) = 1/T^b$, where $(0 < b < 1)$.

\vspace{0.1in}
\noindent\textbf{Connections to Statistical Literature}. Our current framework shares similarities with LASSO problem~\citep{tibshirani1996regression, donoho1996density, chen1998atomic, zhao2006model, zou2006adaptive} in linear regression models.
Both of them introduces the penalization terms to enforce the solution to obey fairness constraints / sparsity to some degree.
In our penalized MAB framework, whether an arm $k$ is played at least $\tau_k T$ times or not depends on the penalty rate $A_k$ and the sub-optimality gap $\Delta_k$. 
Similarly, in the LASSO framework, whether a coefficient is to be estimated as zero depends on the penalty parameter and its true coefficient value.  
Such sparsity features make our method more interpretable. 

\vspace{0.1in}
\noindent\textbf{Differences between Linear Bandits with Regularization}.
There is a line of literature on topic, regularized linear bandit with regularization terms,
including LASSO bandits~\citep{kim2019doubly,bastani2020online}, MCP bandits~\citep{wang2018minimax}, etc.
They are all about solving a penalized least-square estimate.
More formally, it seeks to minimize the following regret function,
$R_{\pi}(T) = \sum_{t=1}^T \mathbb E[\max_{i \in [K]}R_i(x_t) - R_{\pi_t}(X_t)]$,
where reward of each arm $i$ assumes a linear structure, i.e.,
$R_i(x) = x^T \beta_i + \epsilon$, and
$\beta_i$ is a parameter vector which is assumed to be sparse.

The above formulation is significantly different from what we do in this work. Firstly, in our penalization framework, we add regularization term $A_k (\tau_k T - N_{k,\pi}(T))_{+}$'s directly to the regret function. 
In contrast, the literature mentioned above only enforce the sparsity in the algorithms not in the objective function.
Secondly, our goal here is intrinsically different. We aim to maximize reward the under the fairness constraints, while linear bandits aim to infer the sparse reward structure.
Third, note that penalty $A_k (\tau_k T - N_{k,\pi}(T))_{+}$ term here is random and dynamic as $t$ goes on,
it cannot be viewed as a trivial extension of ridge regression or LASSO/MCP regression in linear bandit setting, where the penalty term is a function of unknown deterministic parameter $\beta_i$'s.

\vspace{0.1in}
\noindent\textbf{Practical Choice of $A_k$'s }.
In practical problems, choosing suitable penalty rates $A_k$'s is important and useful.
Here we present two possible solutions.
\begin{itemize}
	\item [1.] 
	One of our suggestions is to choose all $A_k$'s to be equal and set $A_k \equiv A$, where $A$ can be determined by how many arms one wants to fully exploit. For example, $A = \inf_a \{a | \sharp\{k: a \geq \Delta_k \} \leq (K_{exploit} - 1)\}$ when an investor only want to invest on $K_{exploit}$ products (arms). 
	Sub-optimality gaps $\Delta_k$'s can be roughly estimated by prior expert knowledge.
	\item [2.] 
	The other suggestion is to choose $A_k \equiv A$ with
	$A = \inf_a \{a | \sum_{k: a \geq \Delta_k } \tau_k \leq \text{tol}\}$.
	In other word, we do not want $A$ to be too large and guarantee that 
	unfairness level is no larger than pre-defined level $\text{tol}$.
	Again, 
	sub-optimality gaps $\Delta_k$'s can be obtained by prior expert knowledge.
\end{itemize}

\vspace{0.1in}
\noindent\textbf{Highlight of the proof}. 
The technical challenge lies in handling the term $(\tau_k T - N_k(T))_{+}$. 
(1) Our main task in proving upper bound is to show that, for any $k \in \mathcal A_{cr}$, $(\tau_k T - N_k(T))_{+}$ is $O(1)$ for  in gap-dependent setting and it is $ O(T^{1/2})$ for gap-independent setting.
Unlike the classical UCB algorithm analysis, we cannot bound $N_k(T)$ separately for each arm $k$.
Instead, we need to study the relationship between any pair of critical arms and the relationship between critical arm and non-critical arm.
A key step is to find a stopping time $n_1$ such that any arm $k \in \mathcal A_{cr}$ satisfies $N_k(n_1) \geq \tau_k n_1$. 
Therefore, between rounds $n_1$ and $T$, the behavior of $(\tau_k T - N_k(T))_{+}$ can be well controlled. 
(2) In proving maximal inequality, we need to order $K$ arms according to the values of $\mu_{k} + A_k$. 
Then the bound of $\max_{1 \leq t \leq T} (\tau_k t - N_k(T))_{+}$ can be obtained by a recursive formula (see \eqref{e:deficit}) starting from $k = k_1$ to $k = k_K$, where 
$k_1 := \arg\max \{\mu_k + A_k\}$ and $k_K := \arg\min \{\mu_k + A_k\}$.

\section{Experiment Results}

In this section, multiple experimental results are provided to support our theoretical analysis. In particular, we illustrate that the proposed hard-threshold UCB algorithm achieves the lowest penalized regret, proposed method can be viewed as analogy of LASSO for best arm selection, and it also returns the highest reward given the same unfairness level compared with baselines.

\subsection{Setting Descriptions}
	\noindent \textbf{Setting 1.} We investigate the relationship between cumulative penalized regret and total time horizon ($T$) under three algorithms (proposed method, LFG, and Flearn).
	The parameters are constructed as follows.
	The number of arms ($K$) is set to be 5 or 20. 
	The total time horizon ($T$) varies from 500 to 16000. 
	The fairness proportion $\tau_k$ of each arm is set to be
	$\tau_k = \tau / K$ with $\tau \in \{0.2, 0.4, 0.8\}$.
	The penalty rate $A_k$ is constructed as $A_k \equiv (\max_k \mu_k - \min_k \mu_k)/2$.
	Each entry of the mean reward vector $(\mu_k)$ is randomly generated between $[0,1]$. 
	The reward distribution of each arm is a Gaussian distribution, e.g., $N(\mu_k, \frac{1}{K^2})$. 
	For Flearn algorithm, we take tuning parameter $\alpha = 0$. For LFG algorithm, we take $\eta_0 = \sqrt{T}$.
	Each case is replicated for 50 times.

\vspace{0.2in}

	\noindent \textbf{Setting 2.}  We investigate the path of unfairness level ($(\tau_k T - N_k(T))_{+}$) of each arm when the tuning parameter varies.
	The parameters of two cases are constructed as follows. 

\vspace{0.1in}

	Case 1: $K = 8, T = 10000$; $(\mu_1, \ldots, \mu_8) = (0.9, 0.7, 0.6, 0.5, 0.4, 0.3, 0.2, 0.1)$; $\tau_1 = \ldots = \tau_8 = \frac{1}{2K}$. 
	The reward distribution of each arm is a Gaussian distribution, e.g., $N(\mu_k, \frac{1}{K^2})$.

\vspace{0.1in}
	
	Case 2: $K = 8, T = 10000$; $(\mu_1, \ldots, \mu_8) = (0.95, 0.7, 0.65, 0.6, 0.2, 0.15, 0.1, 0.05)$; $\tau_1 = \ldots = \tau_4 = 0.8 \frac{1}{K}$ and 
	$\tau_5 = \ldots = \tau_8 = 0.4 \tau_1$. 
	Again, the reward distribution of each arm is $N(\mu_k,\frac{1}{K^2})$.

\vspace{0.1in}
	
	The penalty rates $A_k \equiv \eta$, where we call $\eta$ is the scale parameter which takes value between 0 and 1.
	For Flearn algorithm, the tuning parameter $\alpha = (1 - \eta) \tau_1 T$ with $\eta$ varying from 0 to 1.
	For LFG algorithm, the tuning parameter $\eta_0 = (1 - \eta) T$ with $\eta \in (0,1]$.
	When scale parameter $\eta \rightarrow 1$, three algorithms will prefer to exploit the arm with highest reward and pay less attention to the fairness.
	On the other hand, $\eta \rightarrow 0$, three algorithms tend to treat the fairness as the priority. 
	
\vspace{0.2in}

	\noindent \textbf{Setting 3.} 
	We investigate the relationship between total expected reward ($\sum_{t=1}^T \mu_{\pi_t}$) and unfairness level ($\sum_{k \in [K]} (\tau_k T - N_k(T))_{+}$) for three algorithms by utilizing MoiveLens 1M data set \footnote{\url{https://grouplens.org/datasets/movielens/}}.	 
	It contains one million ratings from 6000 users on 4000 movies with each rating taking discrete values in $\{1, 2, 3, 4, 5\}$.
	The procedure for pre-processing the dataset are described as follows.
	We extract those movies with number of rating records greater than $m_0$. 
	For each of remaining movies, we view it as an action and we can compute its empirical rating probability as the corresponding reward distribution (i.e., five-category distribution).
	Different values of $m_0$ are considered in our experiments, specifically, $m_0 \in \{2500, 2000, 1250\}$. As a result, there are 13, 31 and 118 actions (movies), respectively.

\newpage

\noindent \textbf{Setting 4.} Moreover, we examine the relationship between number of times that non-critical arm $k$ has been pulled at $T$ (=20000) rounds and the inverse gap $1/(\Delta_k - A_k)^2$. 
In particular, we construct the following three parameter settings ($\tau_k \equiv 1/20$) with $K = 9$ arms.

\vspace{0.1in}

Case 1: $\boldsymbol \mu = (0.9, 0.8, 0.7, 0.6, 0.6, 0.4, 0.3, 0.2, 0.1)$; $A_k \equiv 0.45$. 

Case 2: $\boldsymbol \mu = (0.95, 0.8, 0.7, 0.6, 0.6, 0.4, 0.3, 0.2, 0.1)$; $A_k \equiv 0.41$.  

Case 3: $\boldsymbol \mu = (0.9, 0.8, 0.7, 0.6, 0.6, 0.425, 0.4, 0.375, 0.35)$; $A \equiv 0.45$.  

\vspace{0.1in}

Similarly, we examine the relationship between number of times that critical arm $k$ has been pulled at $T$ (=20000) rounds and the inverse gap $1 / \Delta_k^2$.
We set $K = 9$, $\tau_k \equiv 1/20$, $A_k \equiv 0.45$.

\vspace{0.1in}

Case 1: $\boldsymbol \mu = (0.9, 0.86, 0.84, 0.82, 0.6, 0.4, 0.3, 0.2, 0.1)$.

Case 2: $\boldsymbol \mu = (0.9, 0.85, 0.84, 0.83, 0.82, 0.4, 0.3, 0.2, 0.1)$.

Case 3: $\boldsymbol \mu = (0.9, 0.8, 0.7, 0.6, 0.5, 0.4, 0.3, 0.2, 0.1)$.

\vspace{0.2in}
\noindent \textbf{Setting 5.} We additionally investigate the relationship between total expected reward ($\sum_{t=1}^T \mu_{\pi_t}$) and unfairness level ($\sum_{k \in [K]} (\tau_k T - N_k(T))_{+}$) for three algorithms under various choices of tuning parameter. 
The parameters are given as follows.
We set $K \in \{5, 20\}$ and $\tau_k \equiv \tau/K$ with $\tau \in \{0,2 0.5\}$. 
Each element in mean reward vector $(\mu_k)$ is generated between 0 and 1.
Moreover, we generate the reward from three different distributions, (i) Gaussian $N(\mu_k, \frac{1}{K^2})$, 
(ii) Beta $Beta(\mu_k, 1 - \mu_k)$, (iii) Bernoulli $Bern(1, \mu_k)$.

\subsection{Result Interpretations}

\begin{figure*}[h!]

\vspace{-0.1in}

	\mbox{\hspace{-0.2in}
		\includegraphics[width=2.35in]{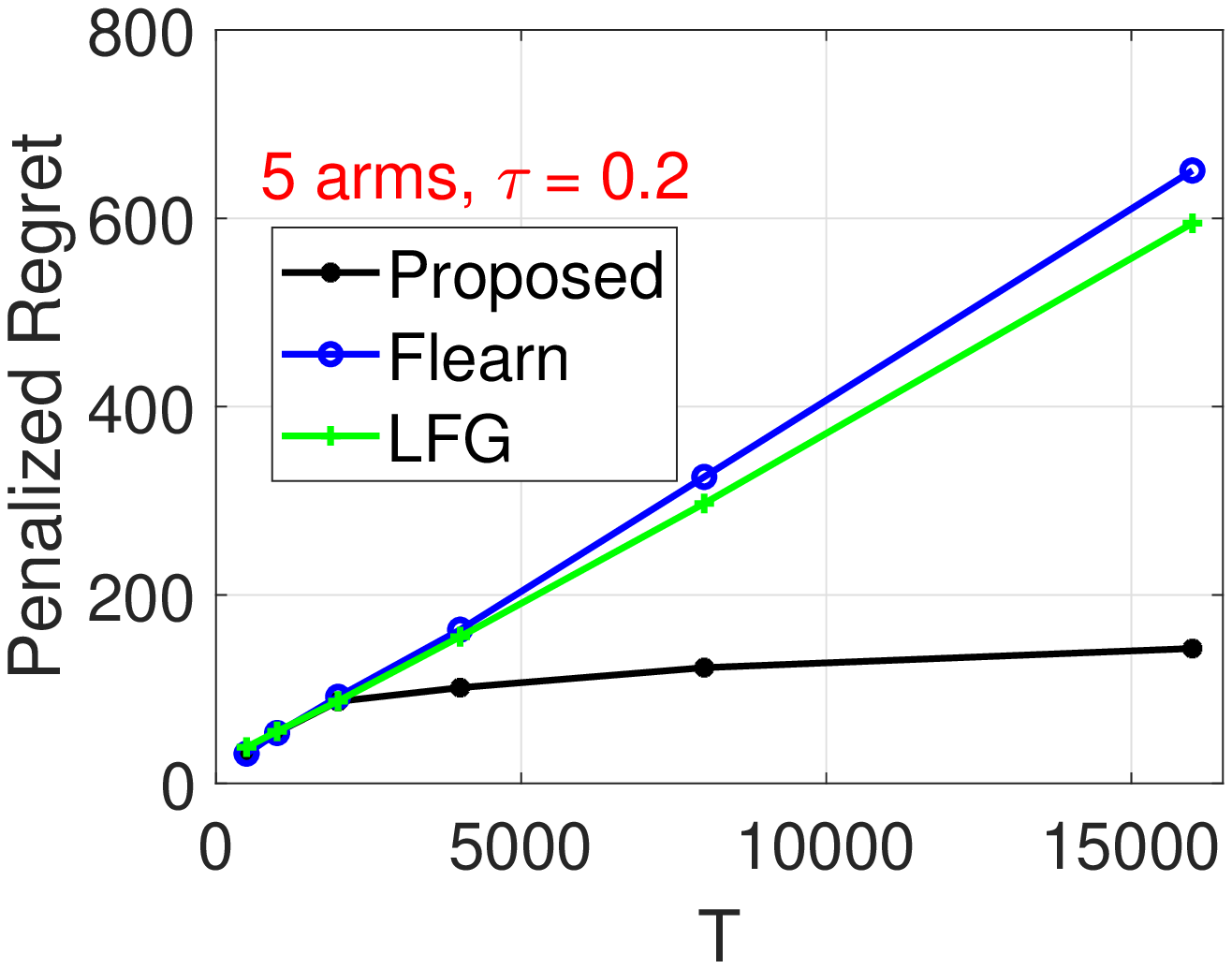}\hspace{-0.12in}
		\includegraphics[width=2.35in]{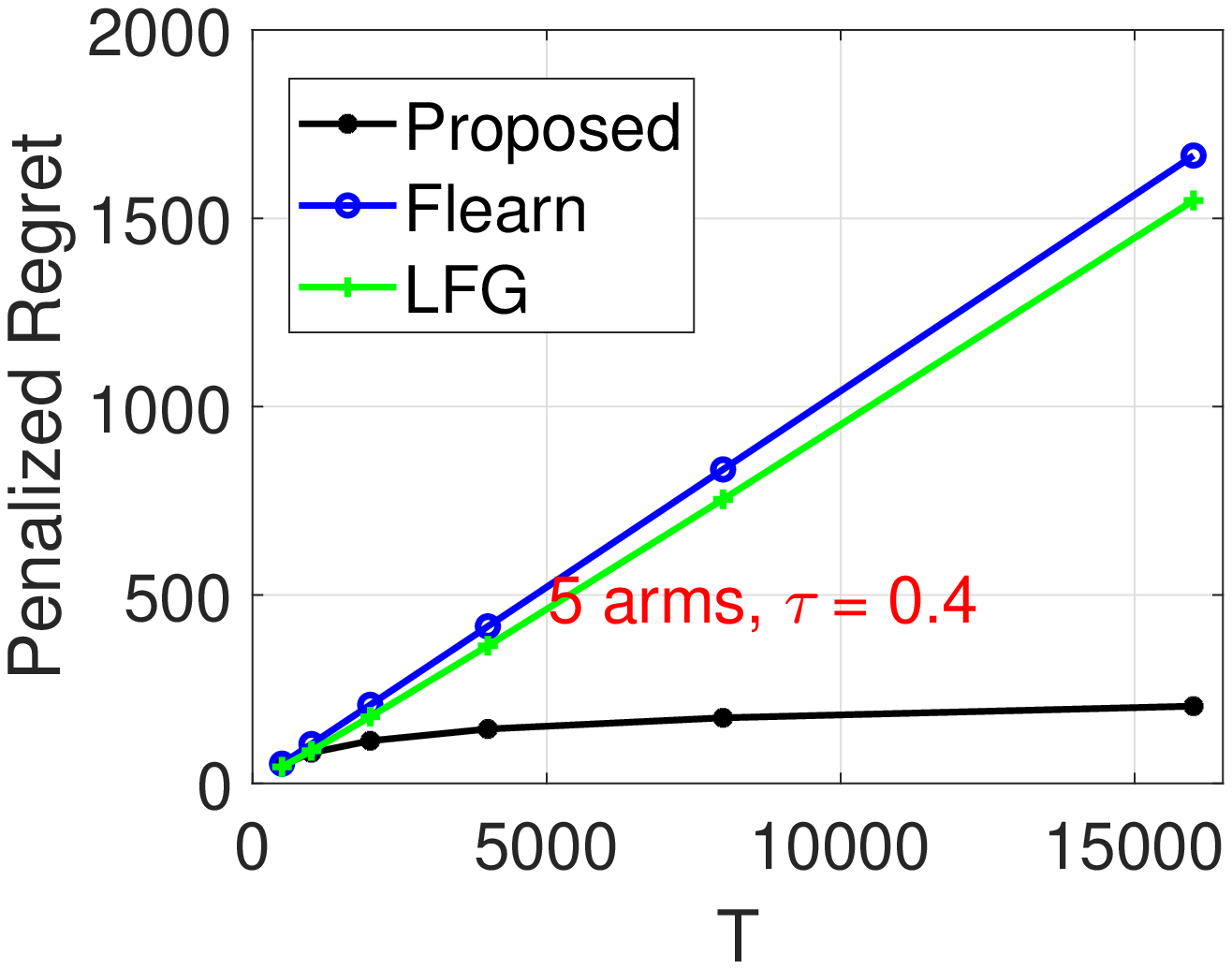}\hspace{-0.12in}
		\includegraphics[width=2.35in]{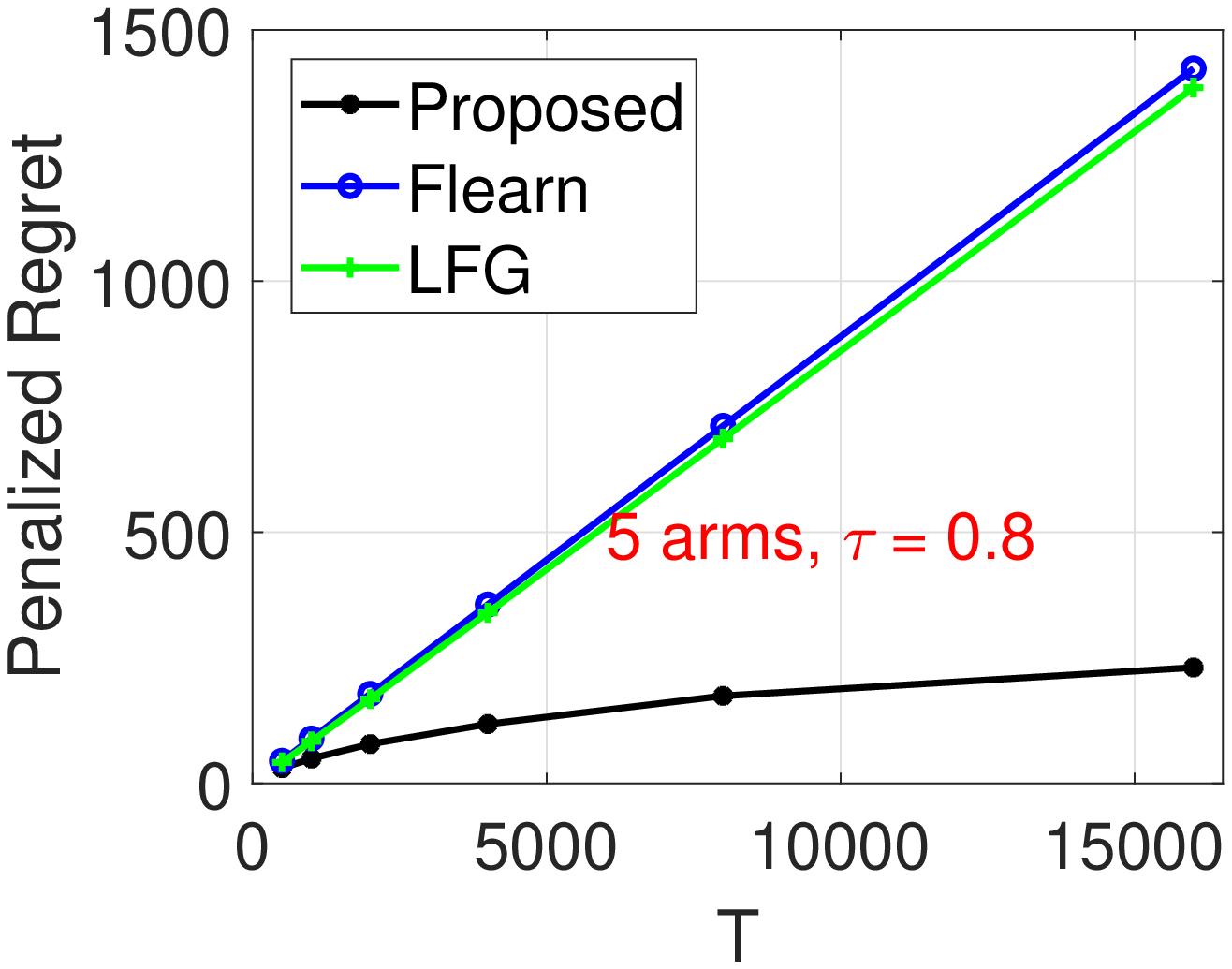}
	}
	\mbox{\hspace{-0.2in}
		\includegraphics[width=2.35in]{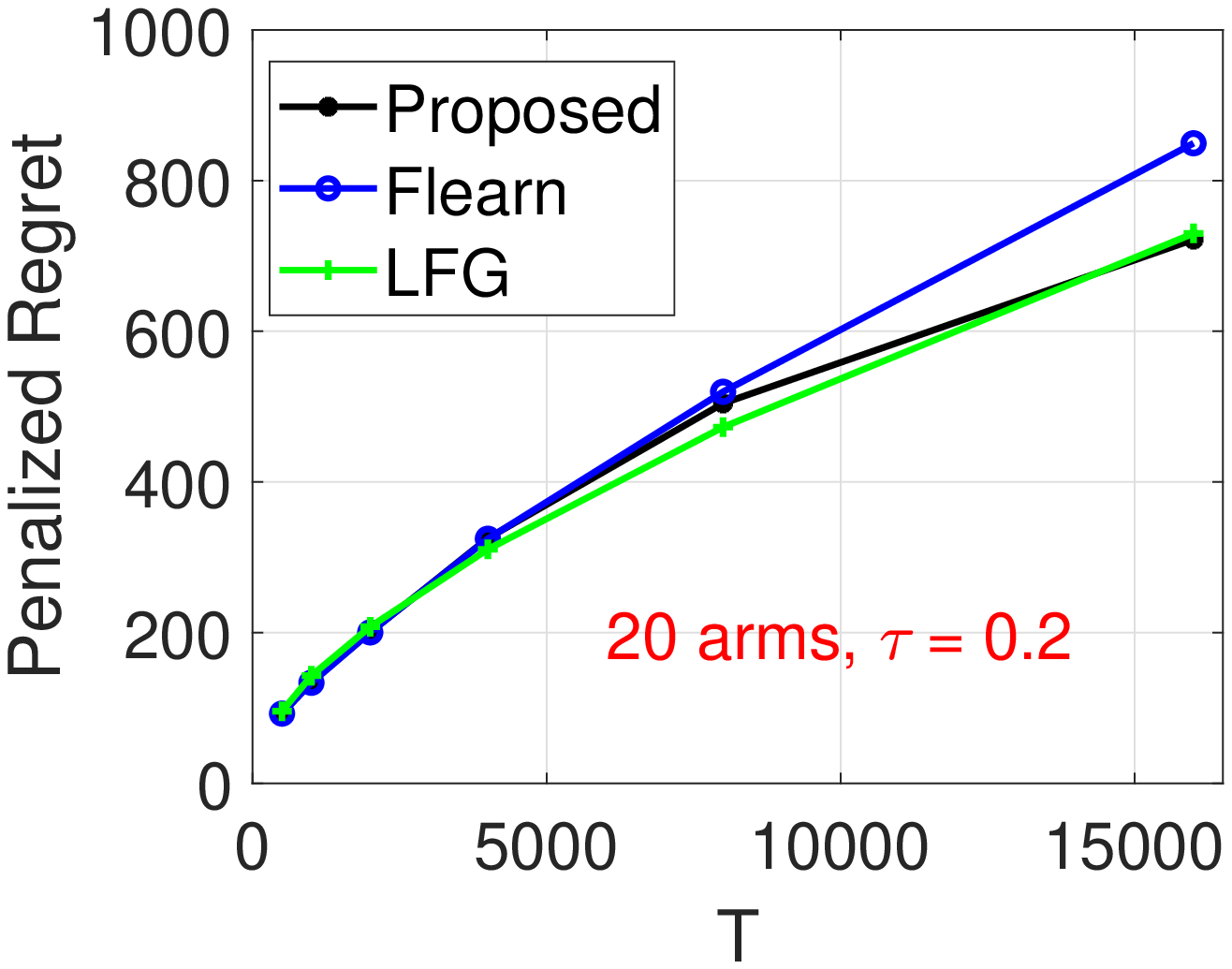}\hspace{-0.12in}
		\includegraphics[width=2.35in]{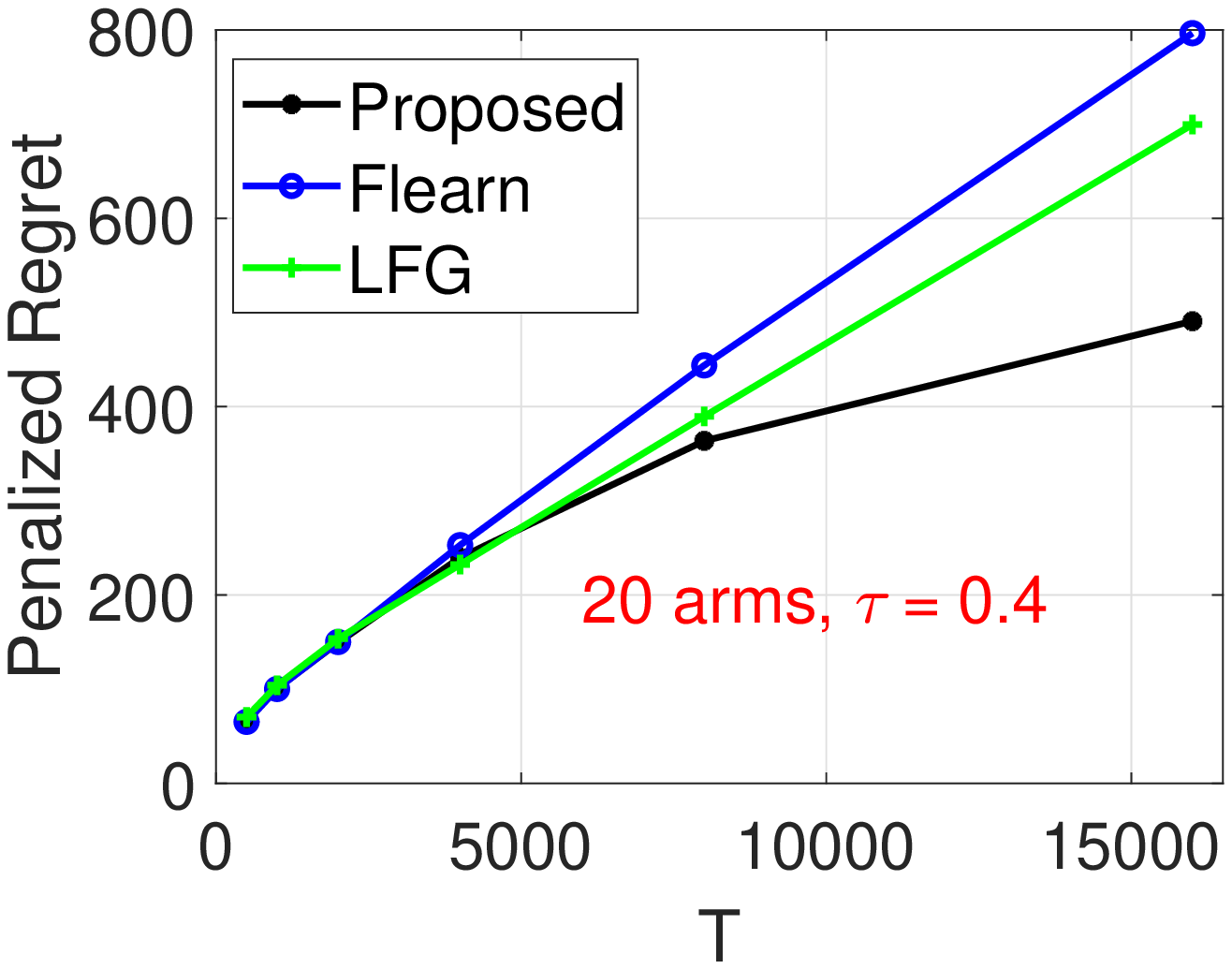}\hspace{-0.12in}
		\includegraphics[width=2.35in]{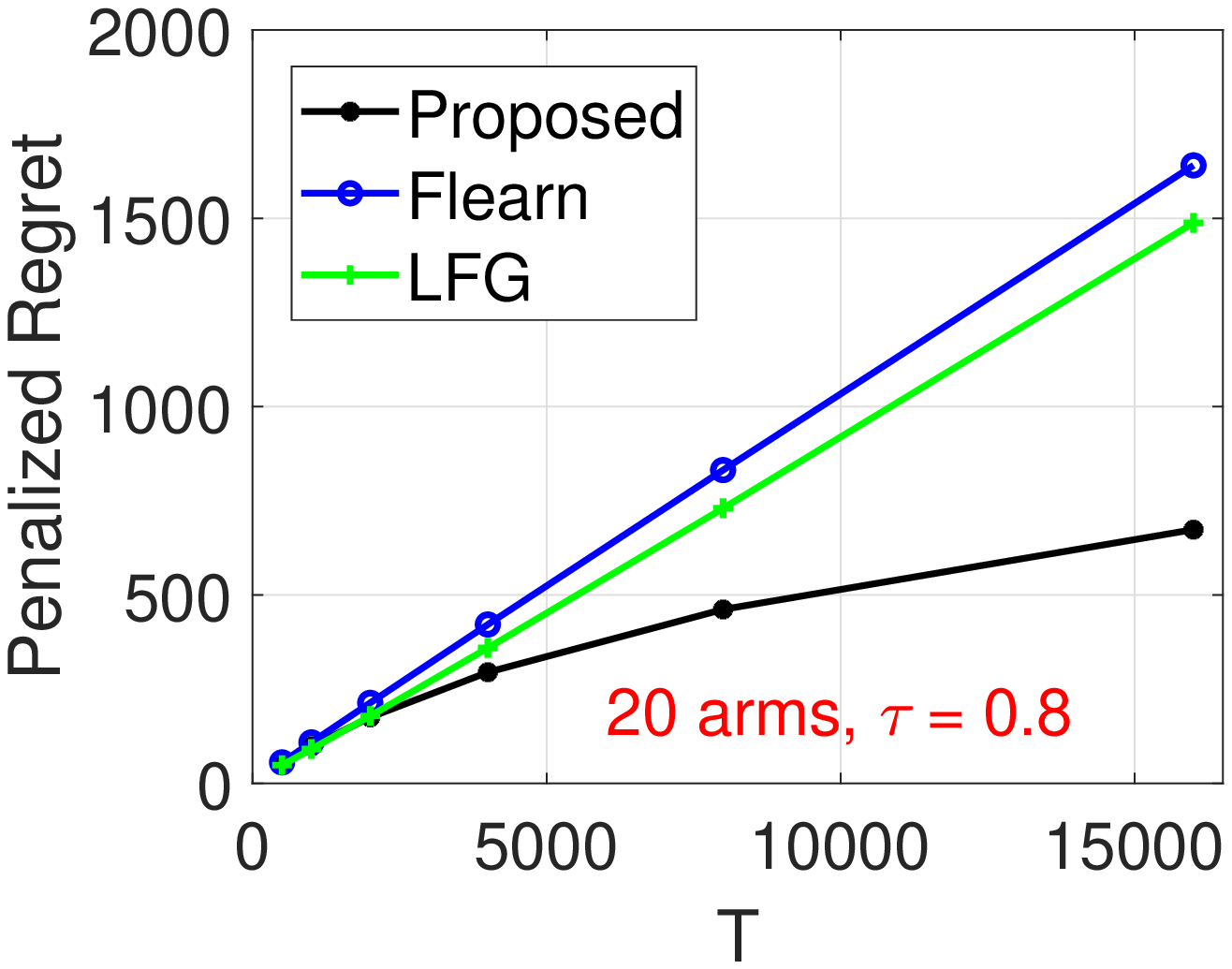}
	}

\vspace{-0.15in}

	\caption{Penalized Regret ($R_{\pi}(T)$) vs Different Time Horizon ($T$) under Setting 1 of different parameters.}\label{fig:setC}\vspace{-0.1in}
\end{figure*}

From Figure \ref{fig:setC}, we observe that the proposed method achieves smaller penalized regret compared with LFG and Flearn. This confirms that our method is indeed a good learning algorithm under penalization framework.

\newpage

\begin{figure*}[t]
	\mbox{\hspace{-0.2in}
		\includegraphics[width=2.35in]{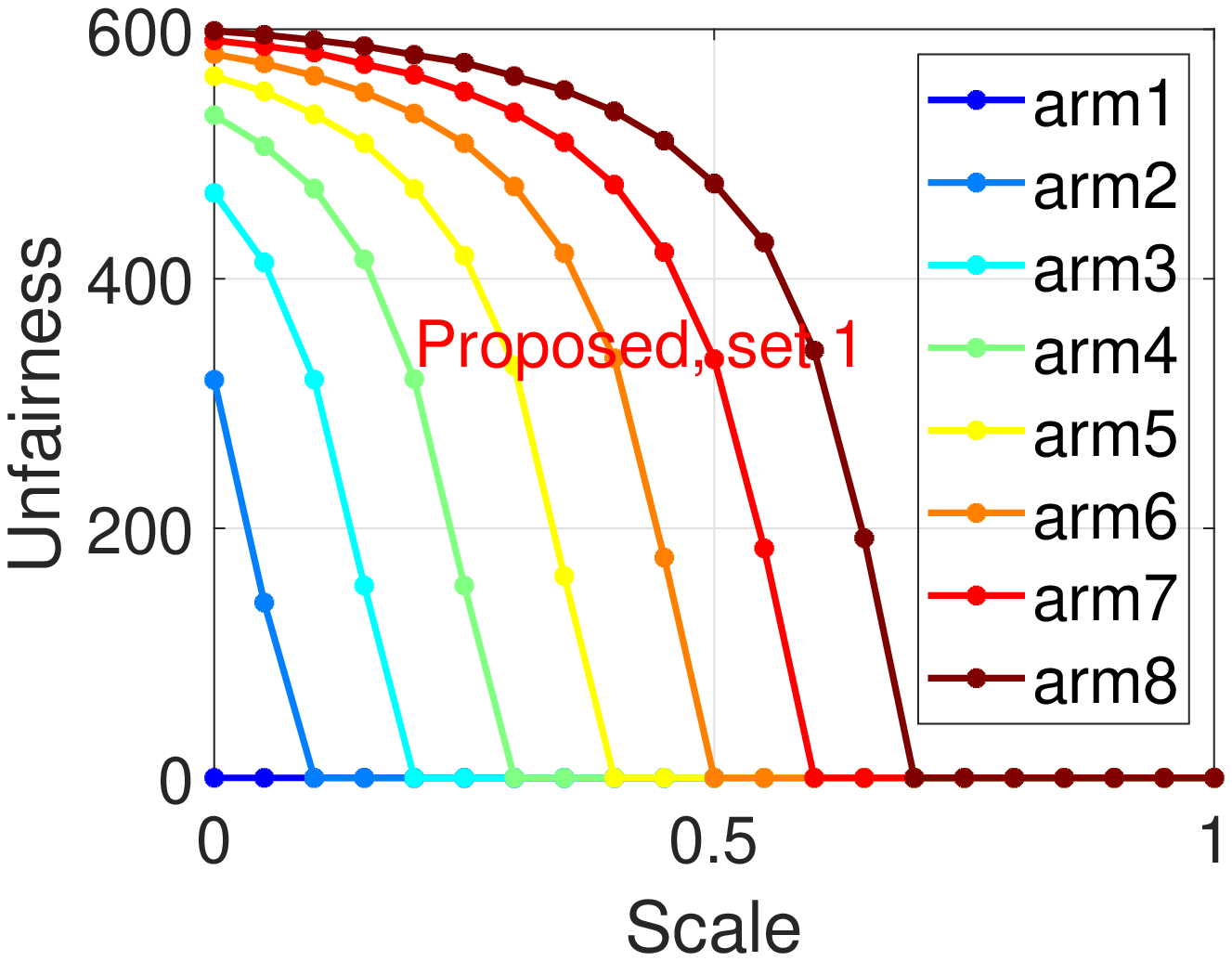}\hspace{-0.12in}
		\includegraphics[width=2.35in]{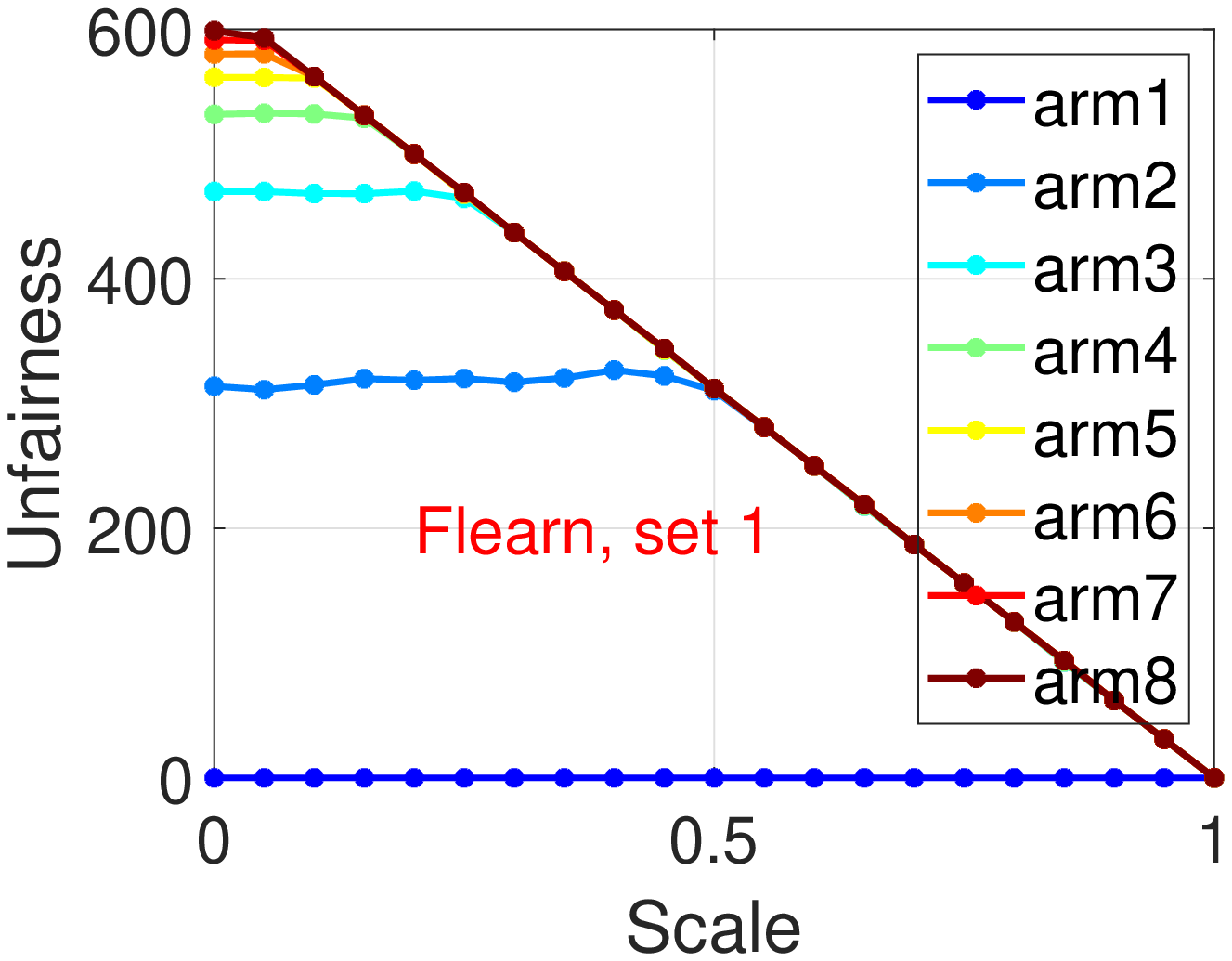}\hspace{-0.12in}
		\includegraphics[width=2.35in]{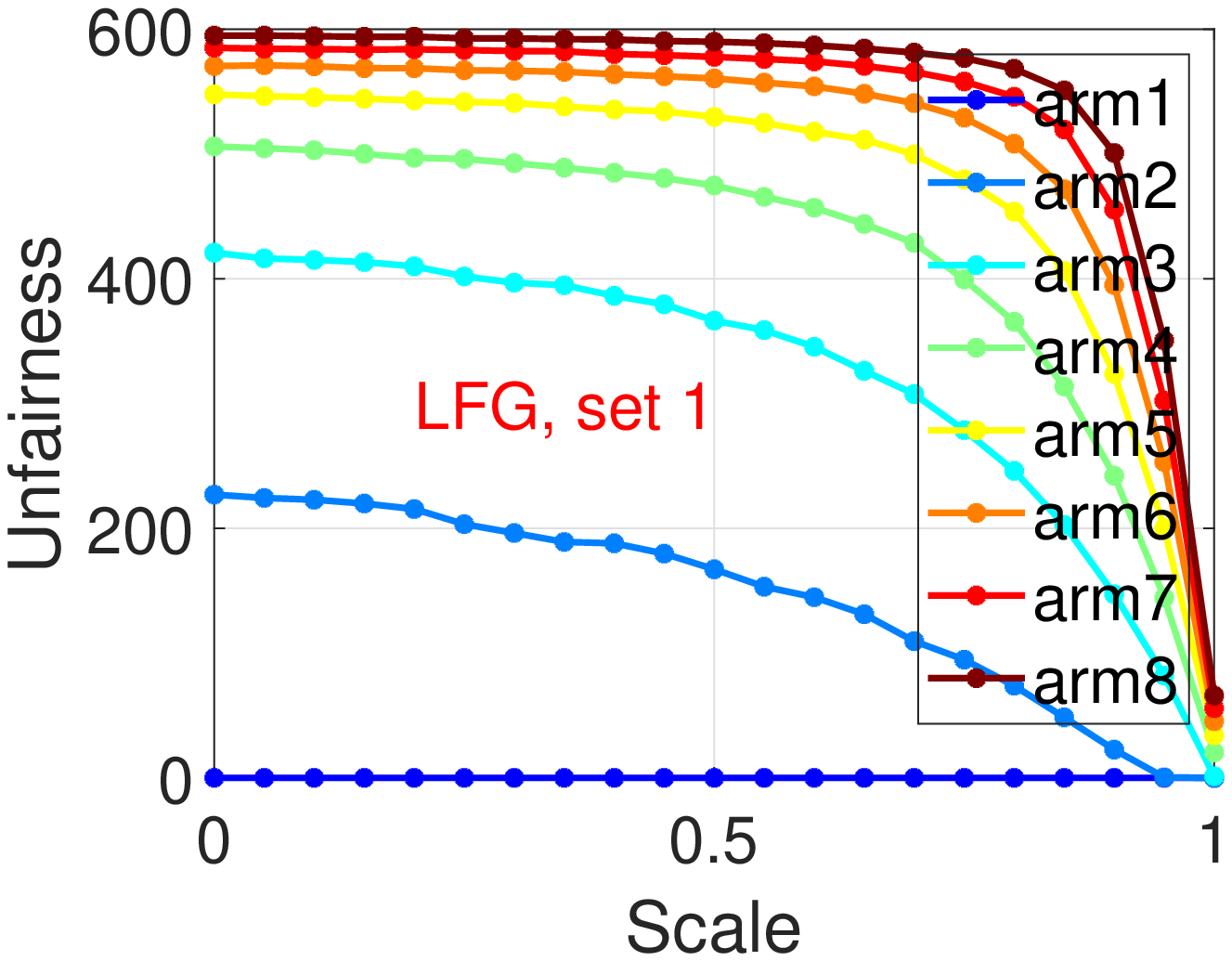}
	}
	
	\mbox{\hspace{-0.2in}
		\includegraphics[width=2.35in]{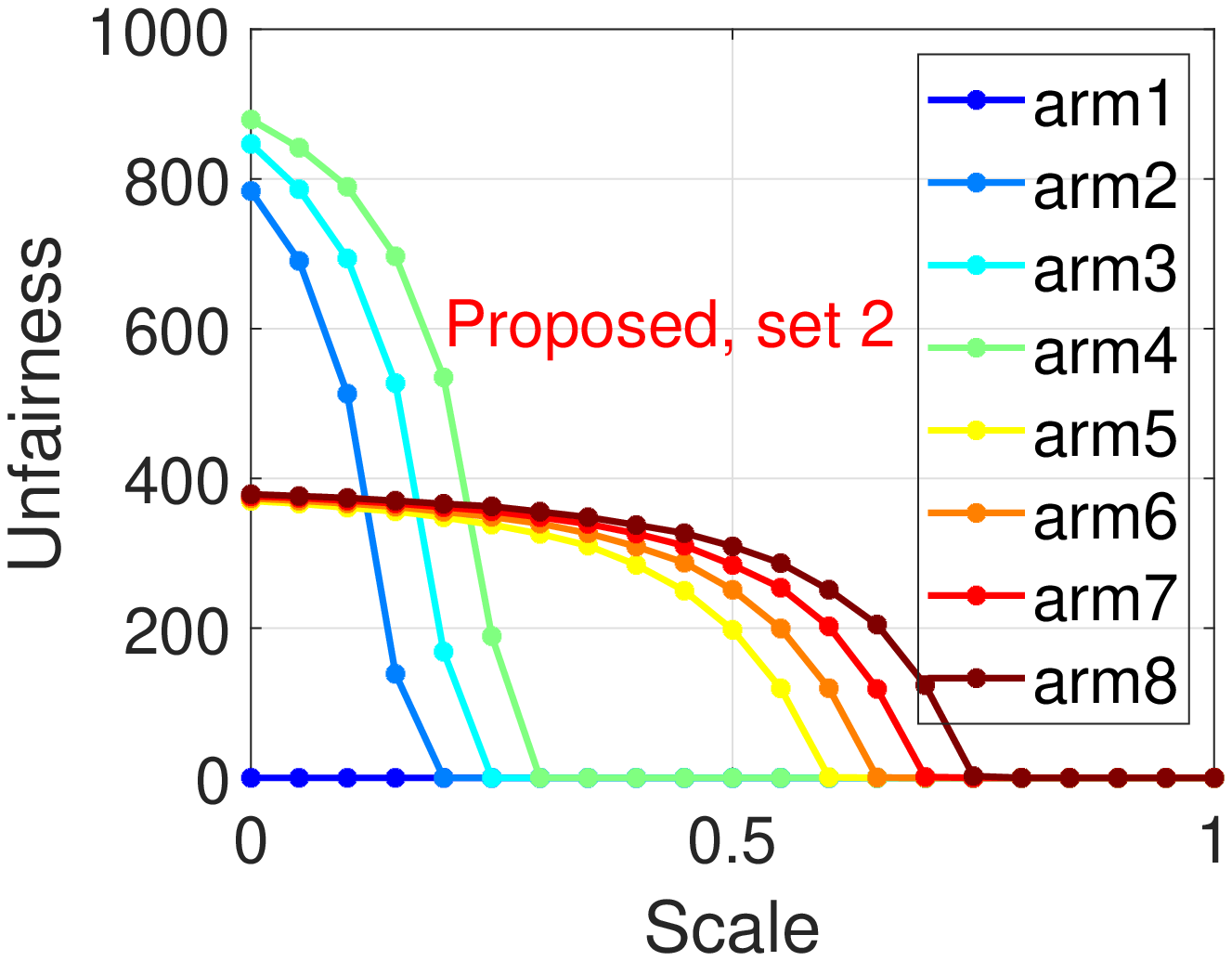}\hspace{-0.12in}
		\includegraphics[width=2.35in]{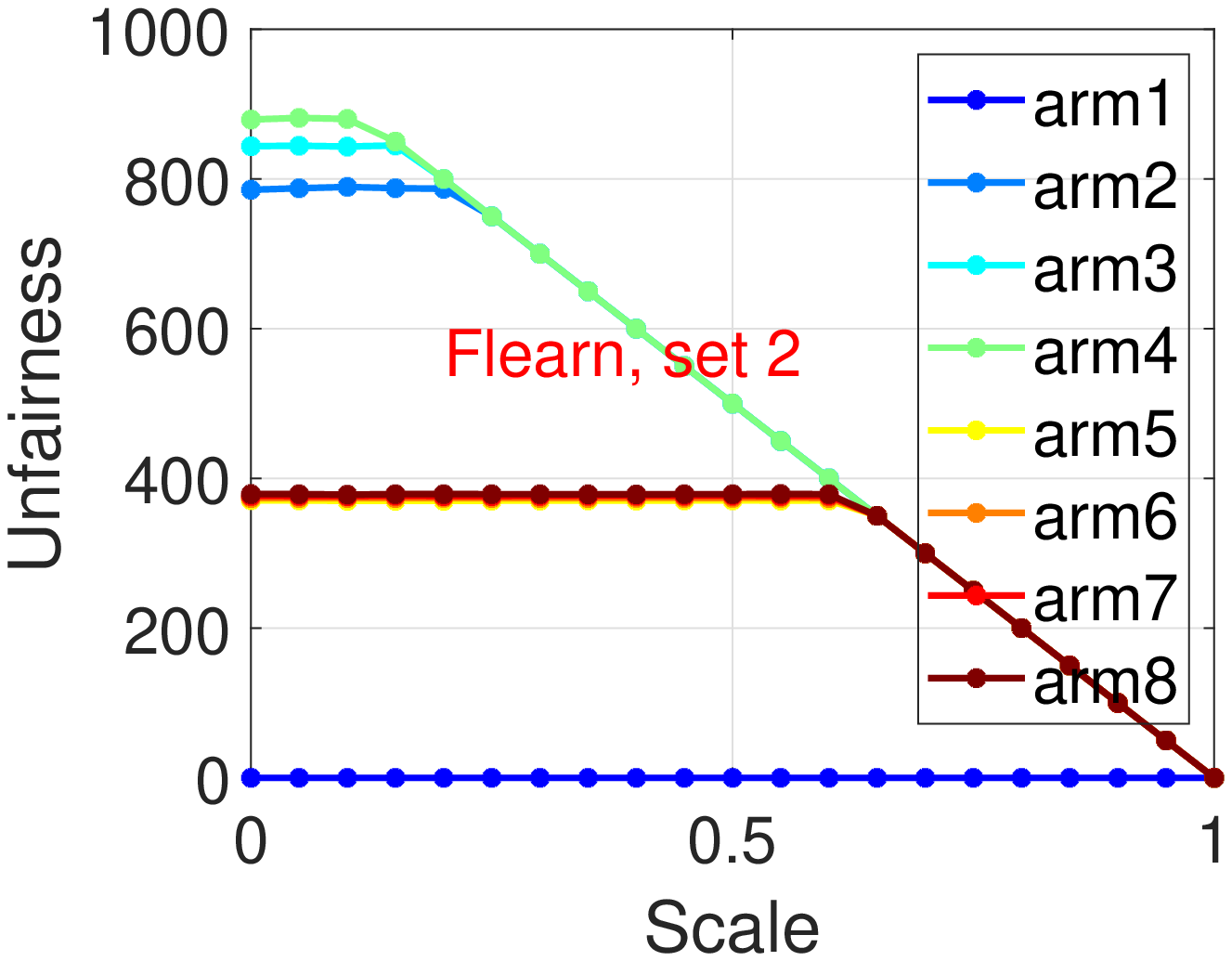}\hspace{-0.12in}
		\includegraphics[width=2.35in]{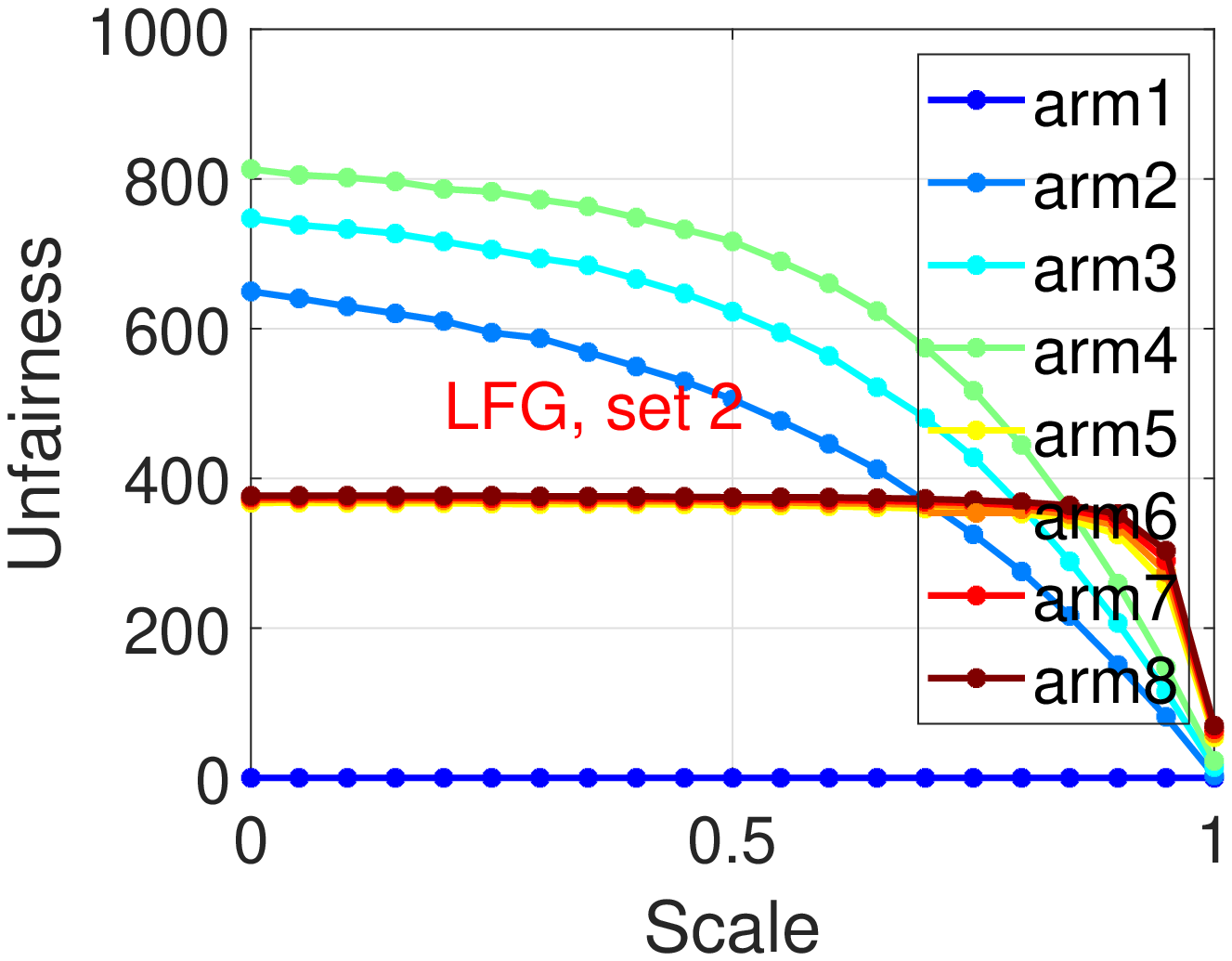}
	}

\vspace{-0.15in}

	\caption{Unfairness path ($(\tau_k T - N_k(T))_{+}$, $k \in [K]$) for three algorithms under two settings described in experiment of Setting 2.
		(Upper row is for Case 1 and bottom row is for Case 2.)
		For sub-optimal arms, the proposed method can guarantee the fairness with a wider range of tuning parameter. By contrast, Flearn and LFG can break the fairness easily.}\label{fig:setD}
\end{figure*}

From Figure \ref{fig:setD}, it can be seen that the paths of unfairness level show different behaviors under three algorithms. 
For our method, with scale parameter decreasing, each arm becomes unfair one by one. By contrast, all arms under both Flearn and LFG methods suddenly become unfair once scale parameter decreases from one. This suggests that our method has sparsity feature as LASSO does, i.e., making arms with small sub-optimality gap fair.   
Therefore, our method can be incorporated into a sparse learning problem for choosing the subset of best arms.

\vspace{0.1in}

\begin{figure*}[h!]	
	\mbox{\hspace{-0.2in}
		\includegraphics[width=2.35in]{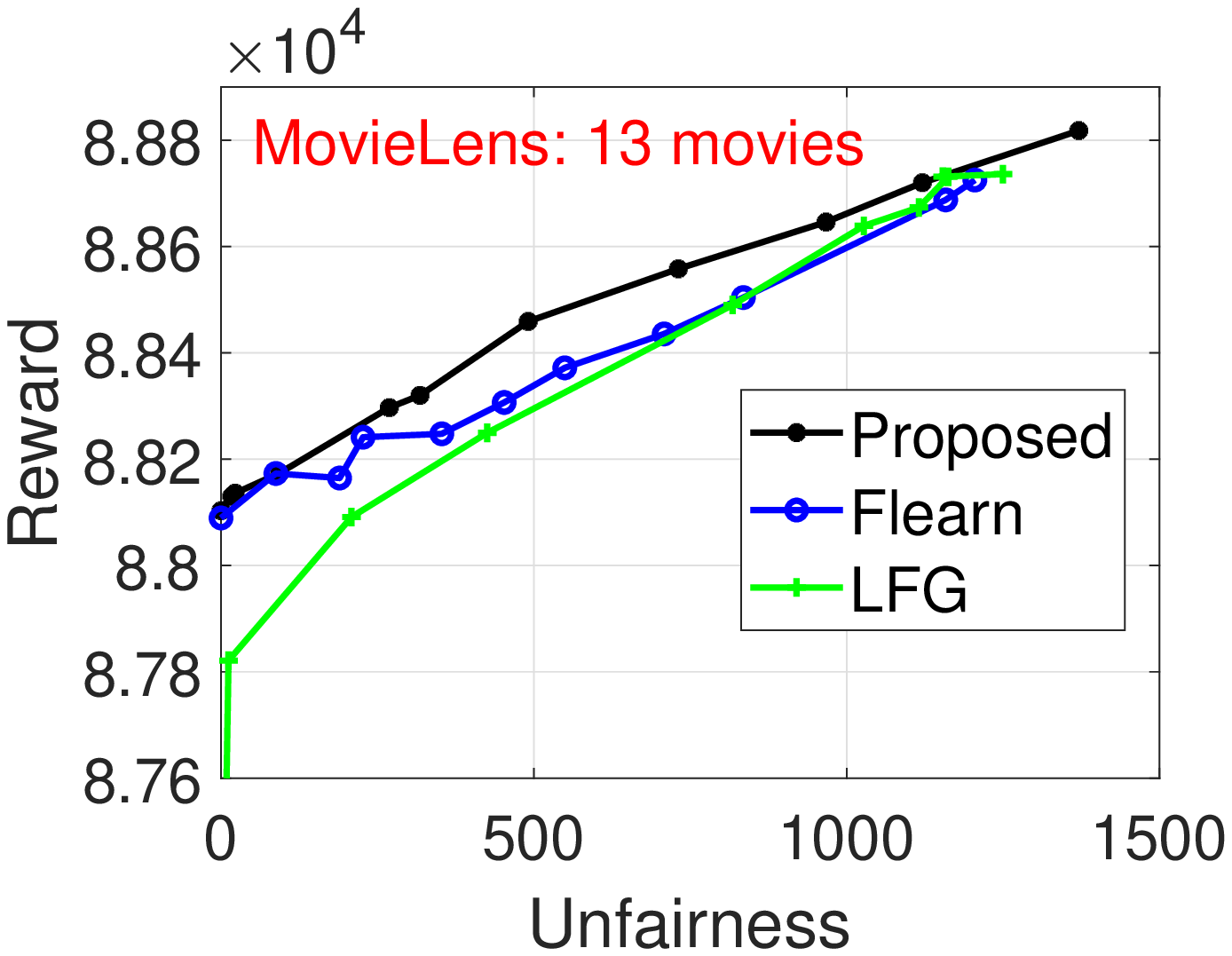}\hspace{-0.12in}
		\includegraphics[width=2.35in]{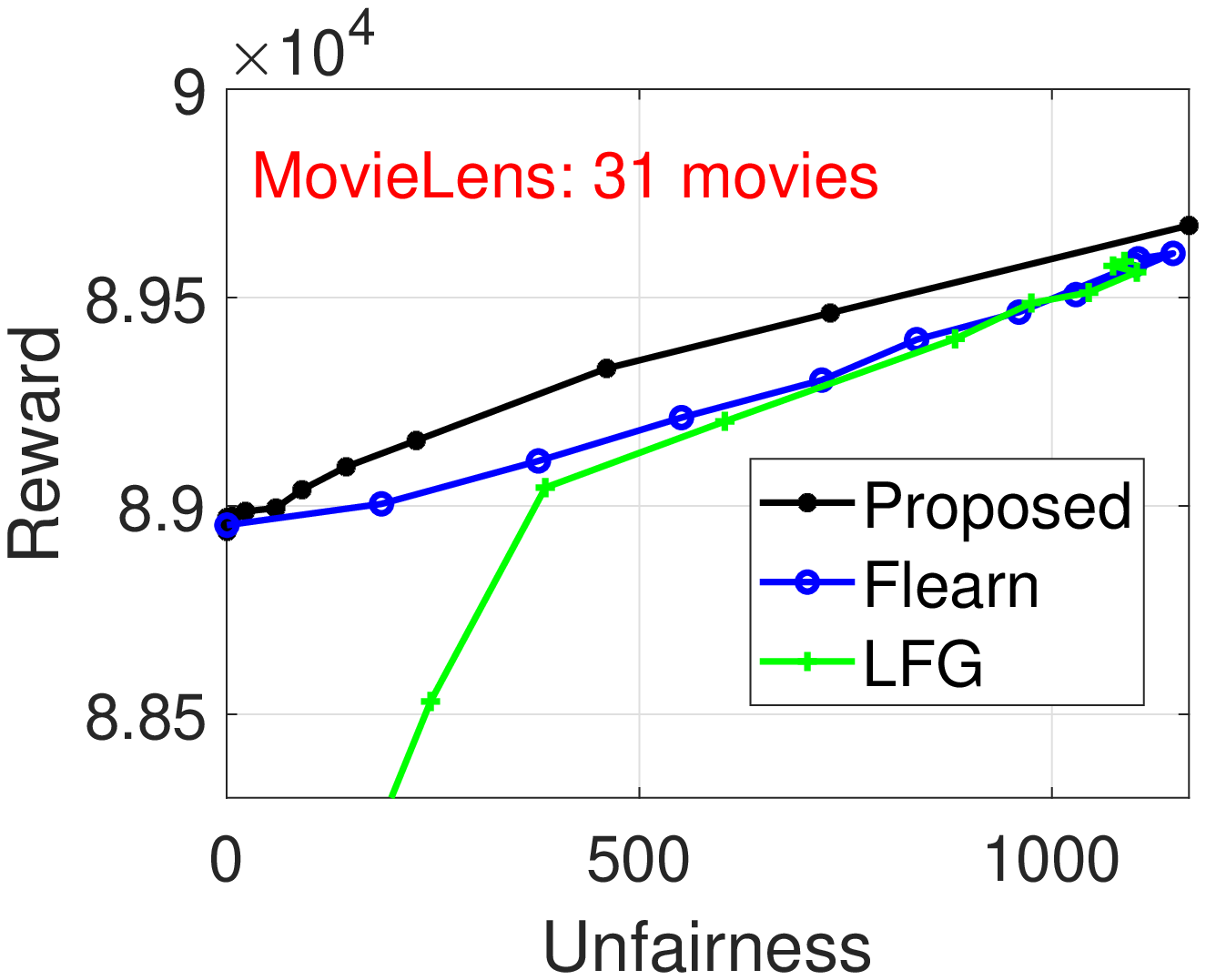}\hspace{-0.12in}
		\includegraphics[width=2.35in]{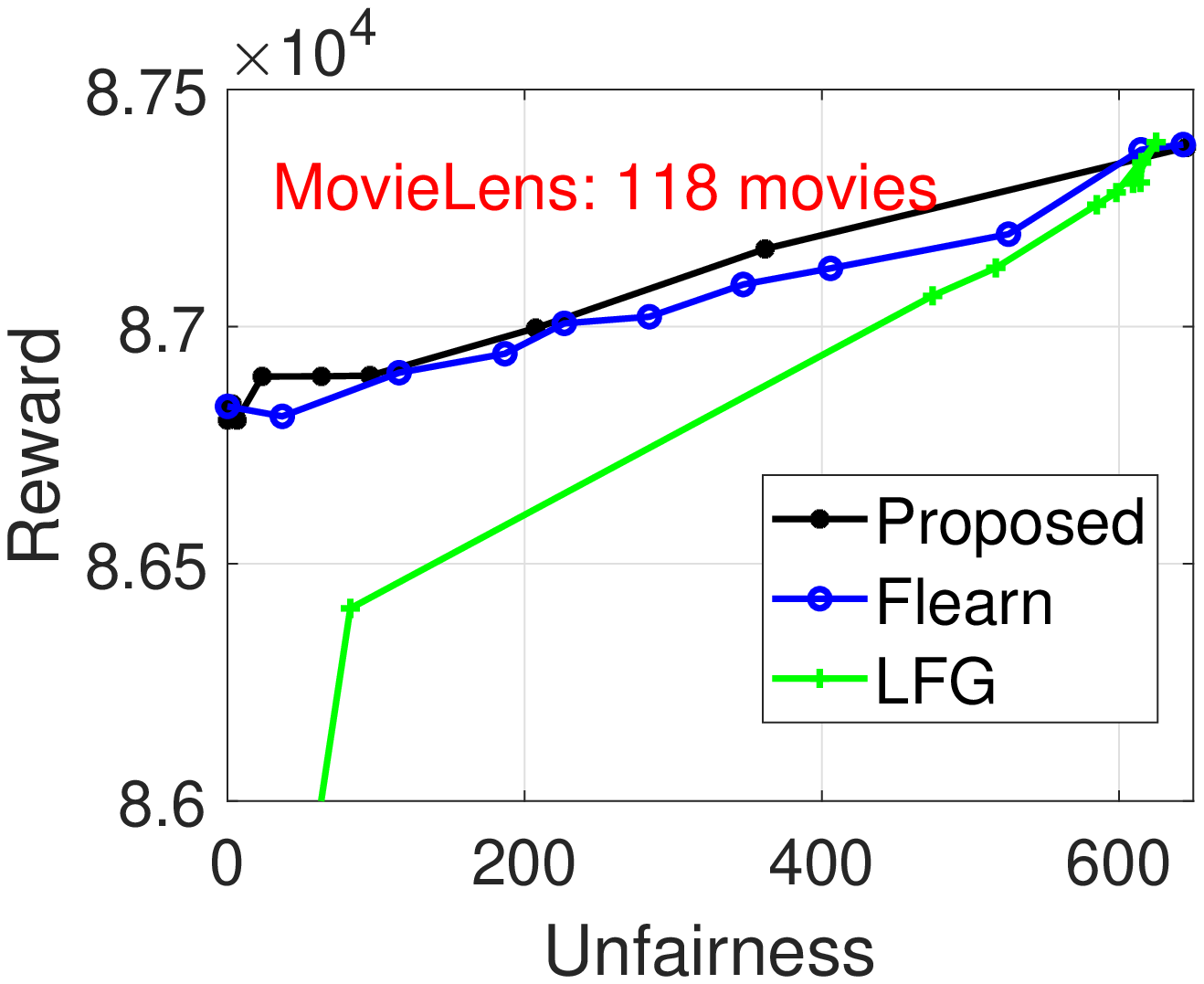}
	}
	
\vspace{-0.15in}

	\caption{Total reward vs unfairness level for three algorithms under different settings based on MoiveLens dataset.}\label{fig:setF}
\end{figure*}

From Figure \ref{fig:setF}, we can tell that the proposed method always achieves the highest reward given the same unfairness level for the MovieLens dataset with different number of movies. This gives the evidence that hard-threshold UCB algorithm makes better balance between total reward and fairness constraints compared with other competing methods.

\begin{figure*}[t]	
	\centering
	\mbox{\hspace{-0.1in}
		\includegraphics[width=2.35in]{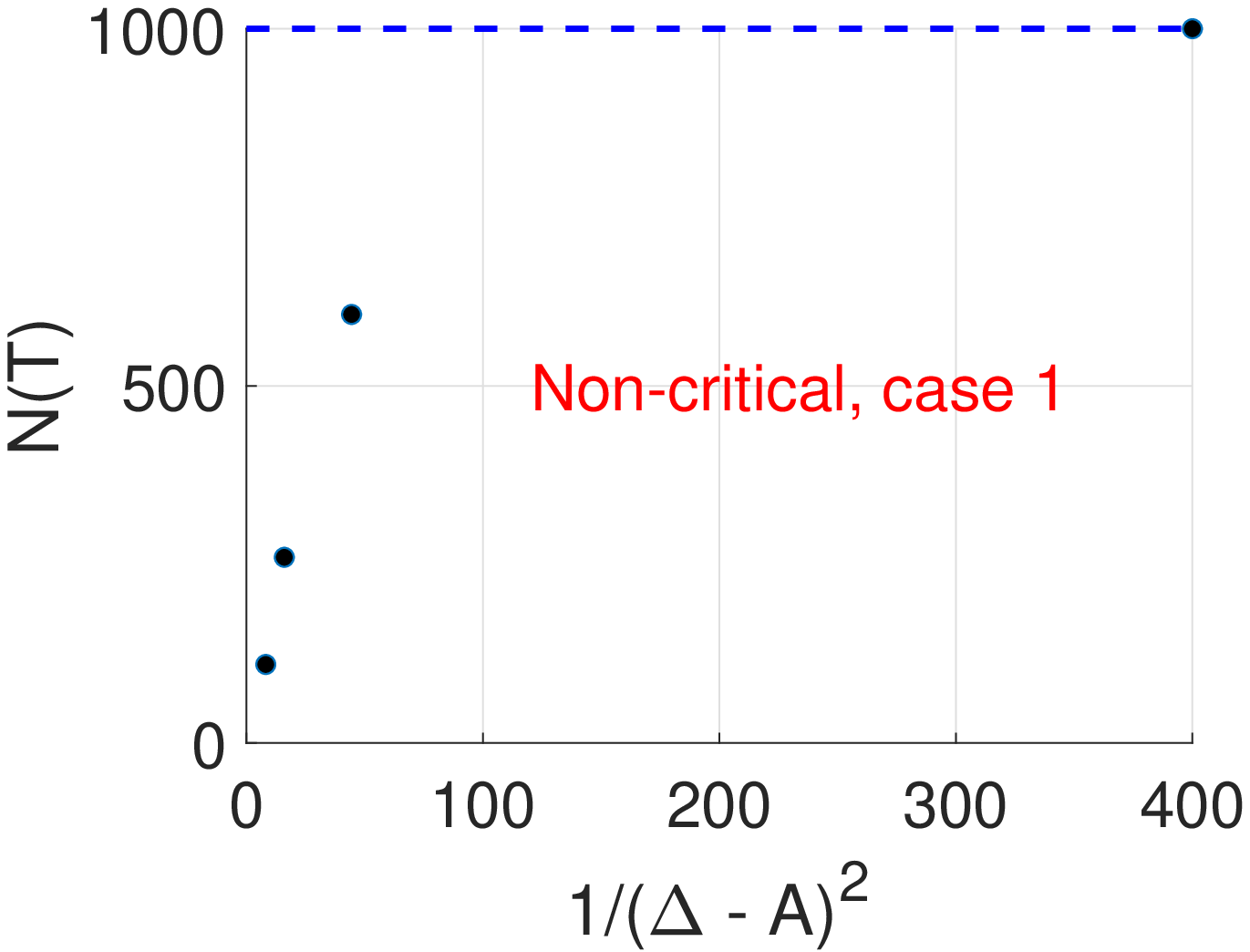}\hspace{-0.1in}
		\includegraphics[width=2.35in]{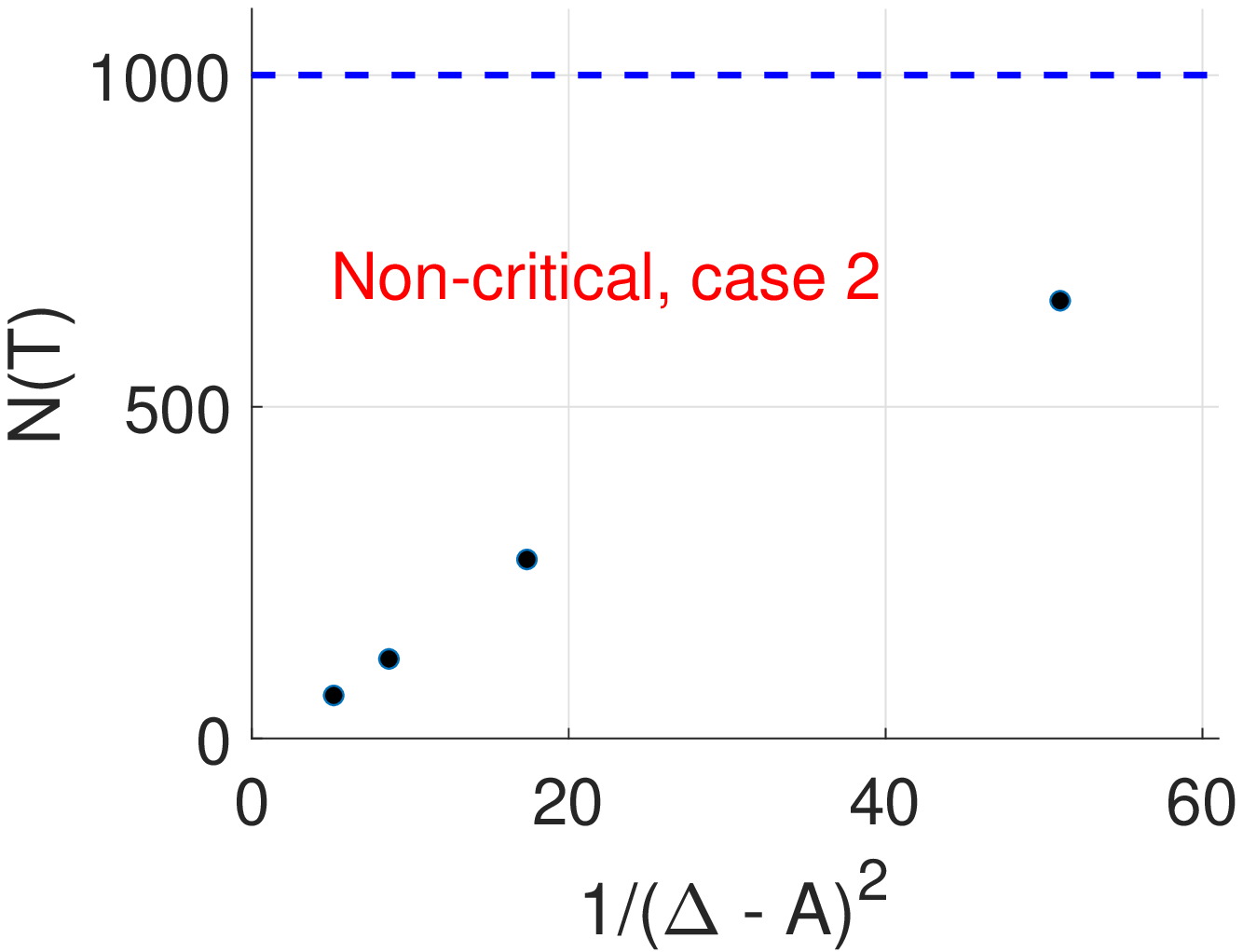}\hspace{-0.1in}
		\includegraphics[width=2.35in]{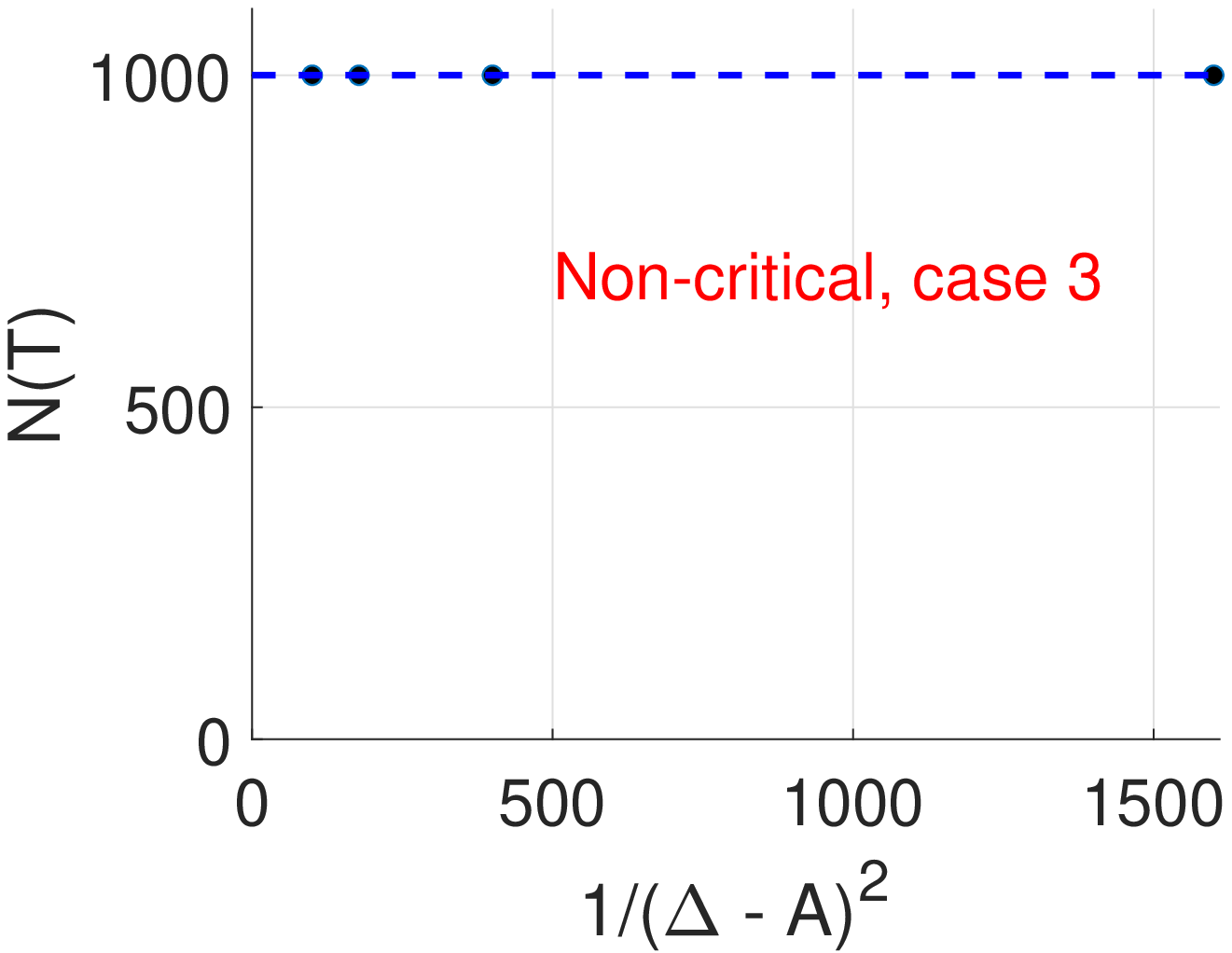}
	}
	
	\mbox{\hspace{-0.1in}
		\includegraphics[width=2.35in]{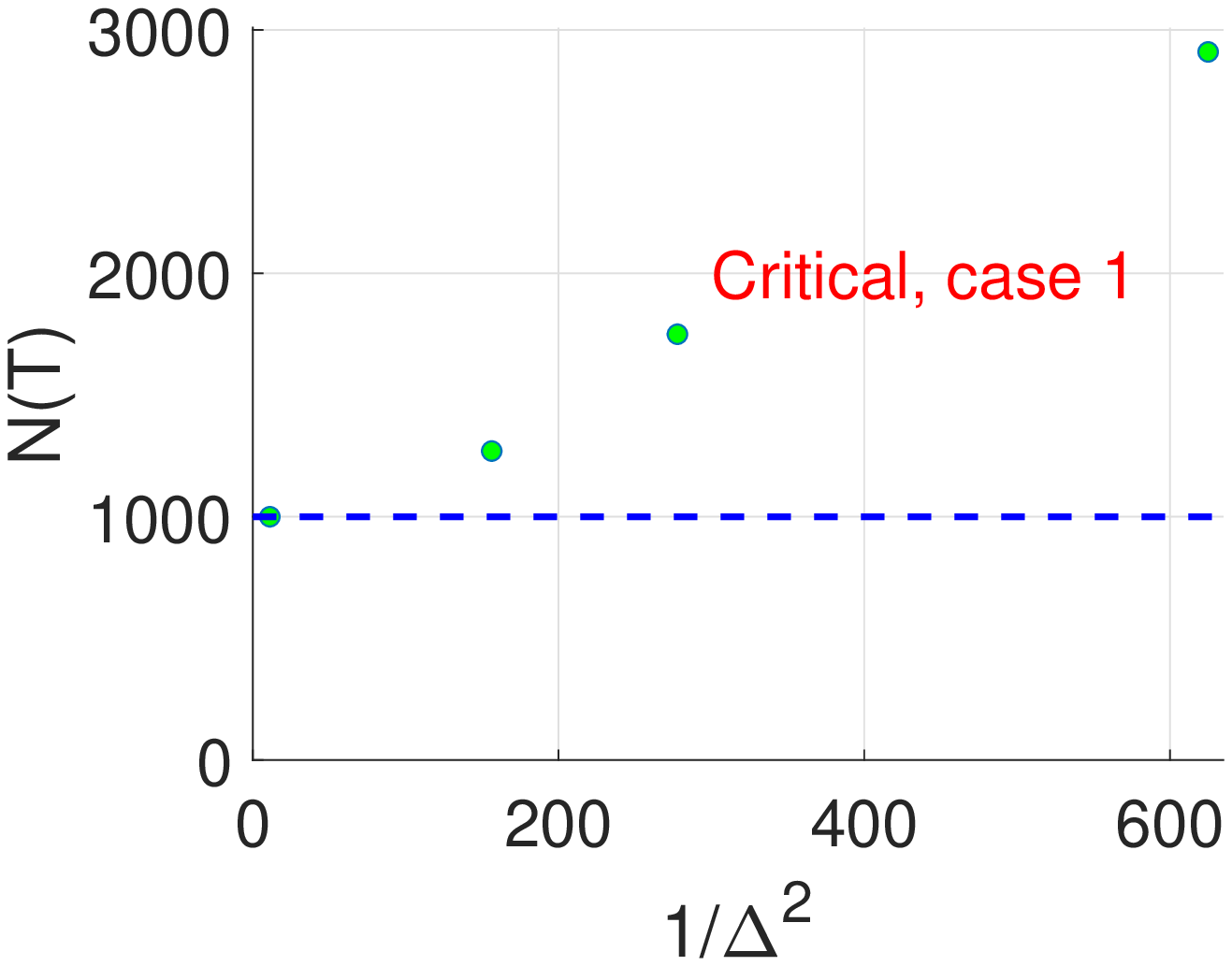}\hspace{-0.1in}
		\includegraphics[width=2.35in]{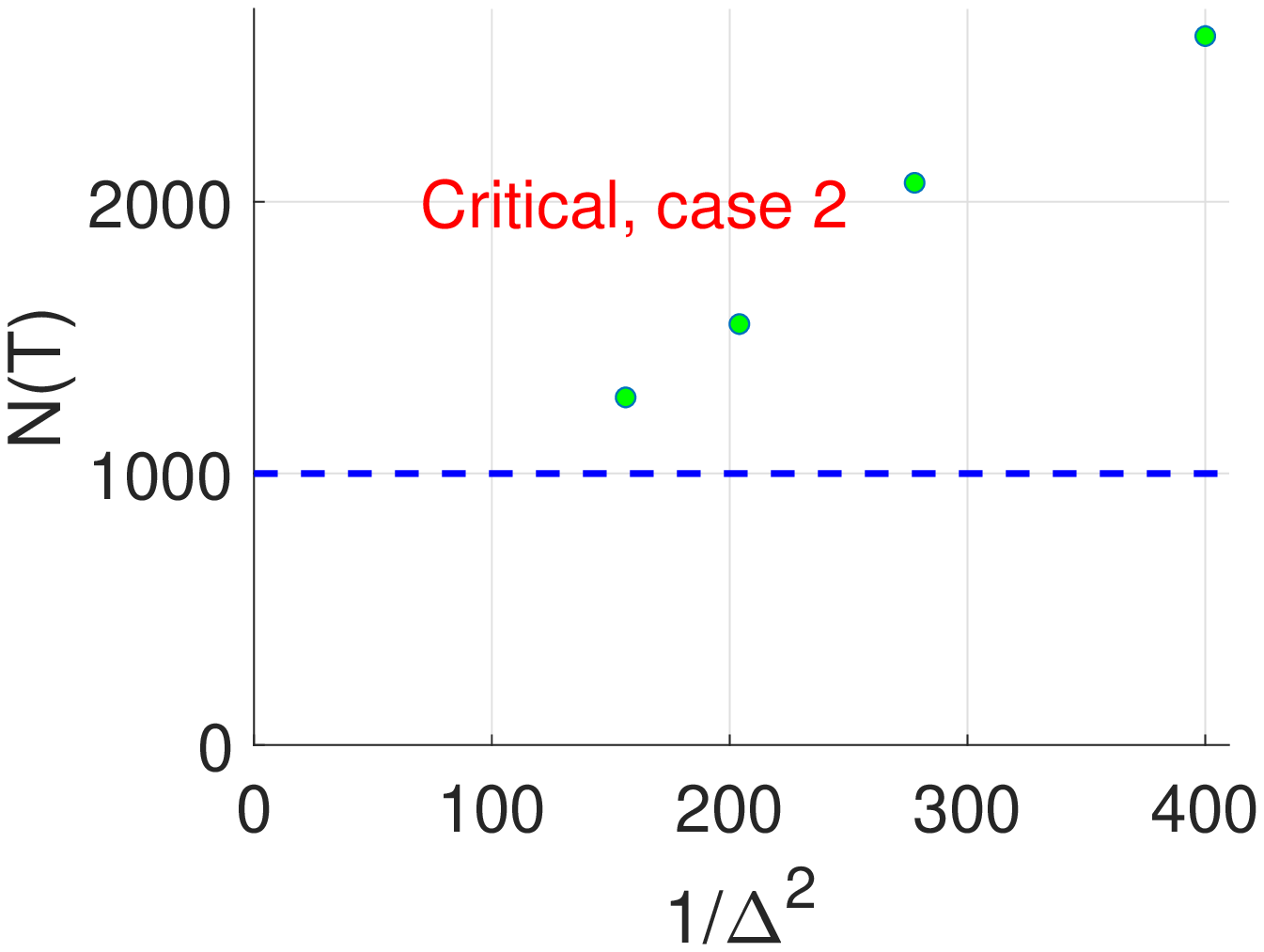}\hspace{-0.1in} 
		\includegraphics[width=2.35in]{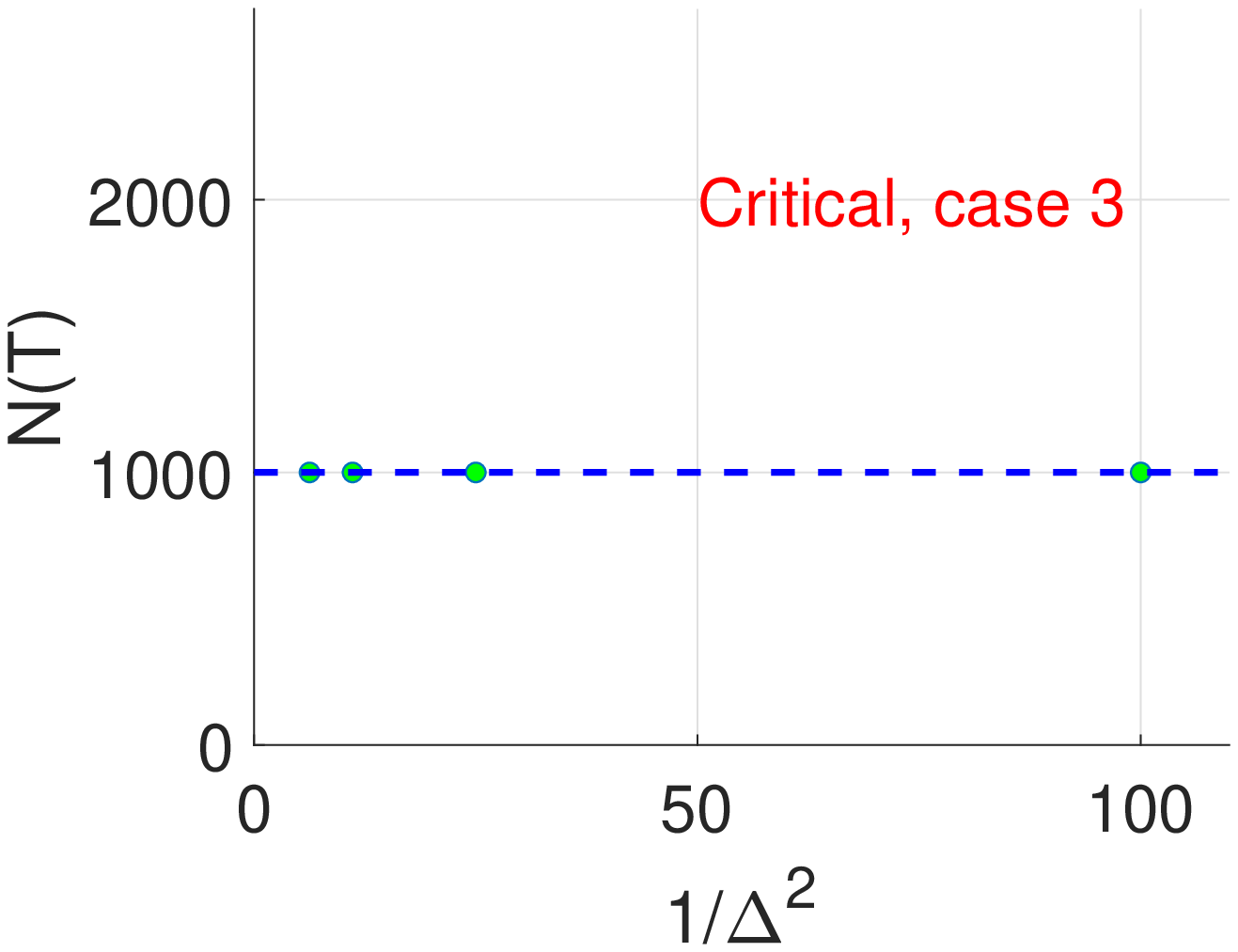}
	}
	\caption{Upper row: $N_k(T)$ vs $1 / (\Delta_k - A_k)^2$ for arm $k \in \mathcal A_{\text{non-cr}}$.
		Bottom row: $N_k(T)$ vs $1 / \Delta_k^2$ for arm $k \in \mathcal A_{\text{cr}}$. 
		In all plots, the blue horizontal line stands for fairness level $\tau_k T$.}\label{fig:setA}
\end{figure*}

From Figure \ref{fig:setA}, we can see that the pulling number $N_k(T)$ is proportional to $1 / (\Delta_k - A_k)^2$ for $k \in \mathcal A_{\text{non-cr}}$ when $N_k(T)$ does not reach fairness level $\tau_k T$.
We also see that $N_k(T)$ is proportional to $1 / \Delta_k^2$ for $k \in \mathcal A_{\text{cr}}$ when the pulling number is larger than fairness level $\tau_k T$. These phenomena match the regret bound obtained in Theorem \ref{thm:dep:upper} and empirically validate the tightness of our theoretical analysis.

\begin{figure*}[t!]
	\mbox{\hspace{-0.2in}
		\includegraphics[width=2.35in]{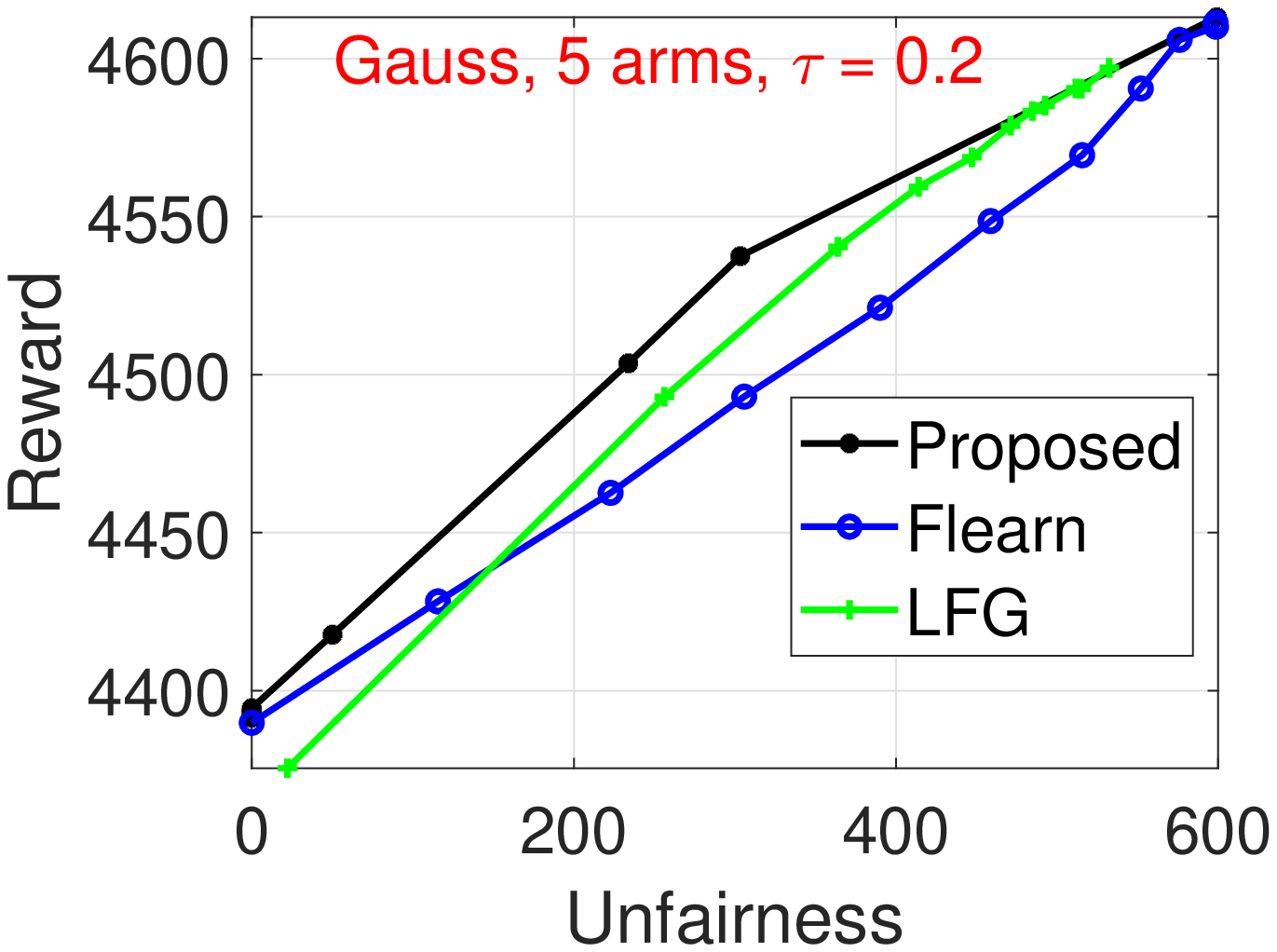}\hspace{-0.12in}
		\includegraphics[width=2.35in]{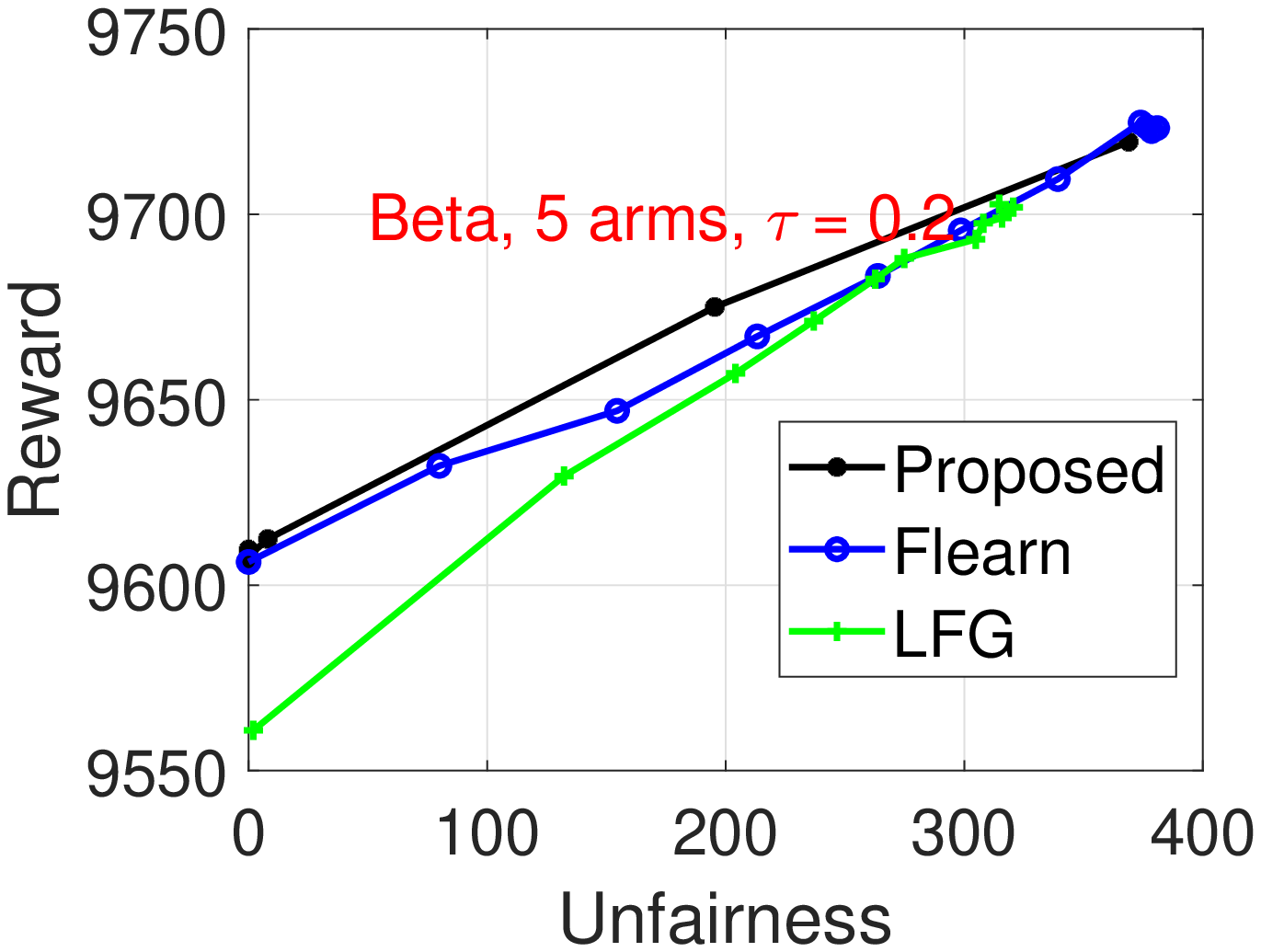}\hspace{-0.12in}
		\includegraphics[width=2.35in]{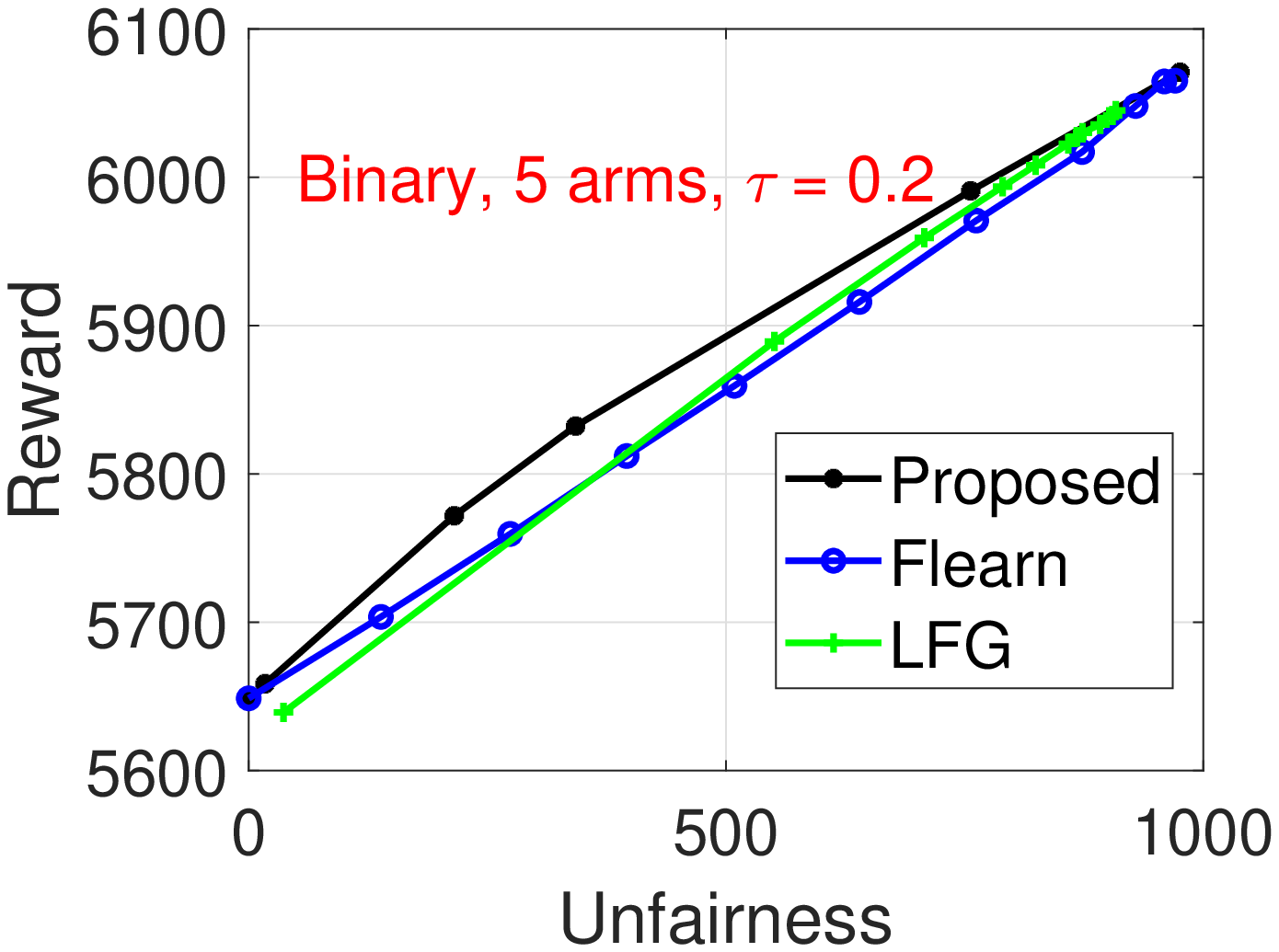}
	}
	
	\mbox{\hspace{-0.2in}
		\includegraphics[width=2.35in]{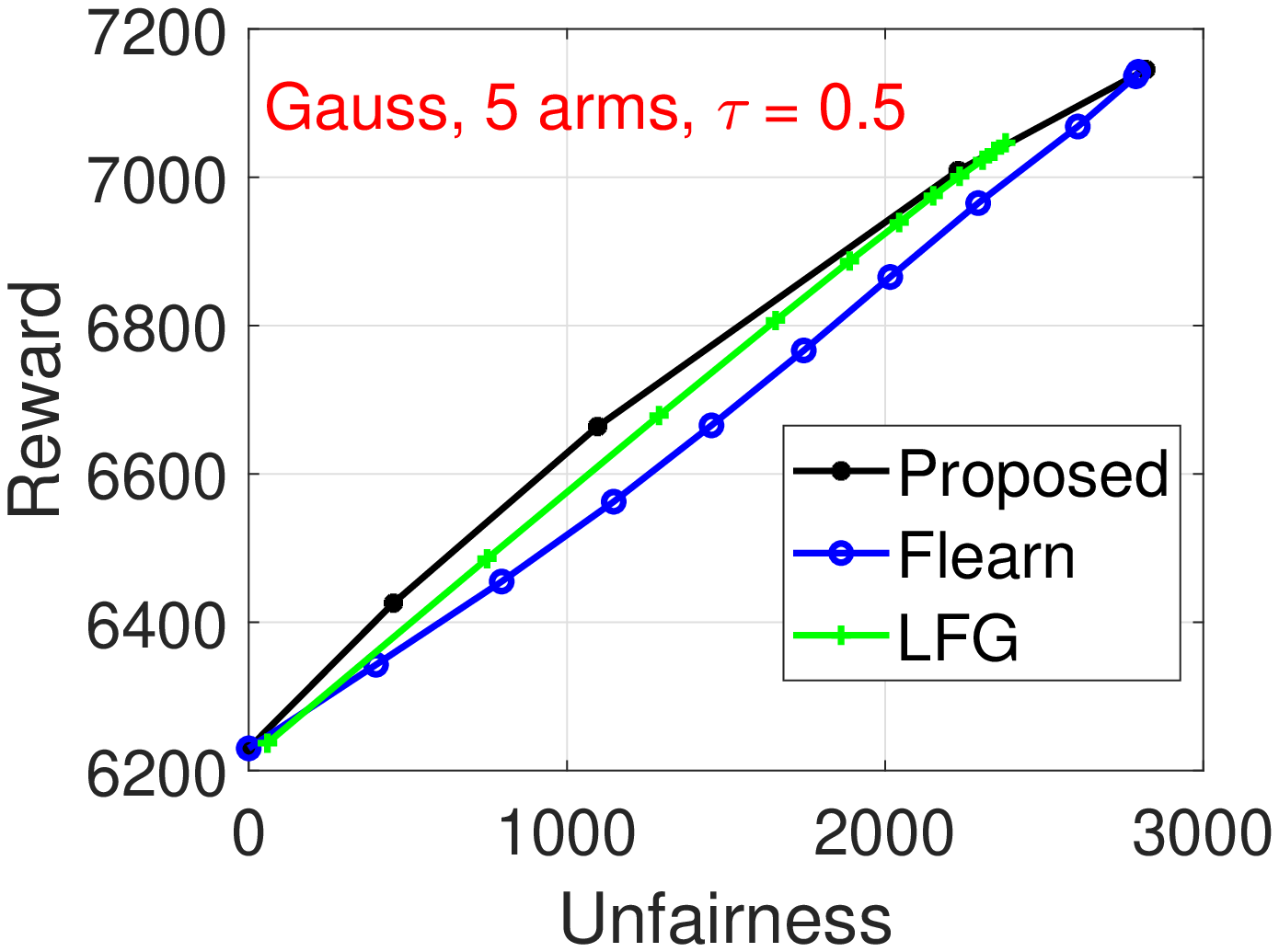}\hspace{-0.12in}
		\includegraphics[width=2.35in]{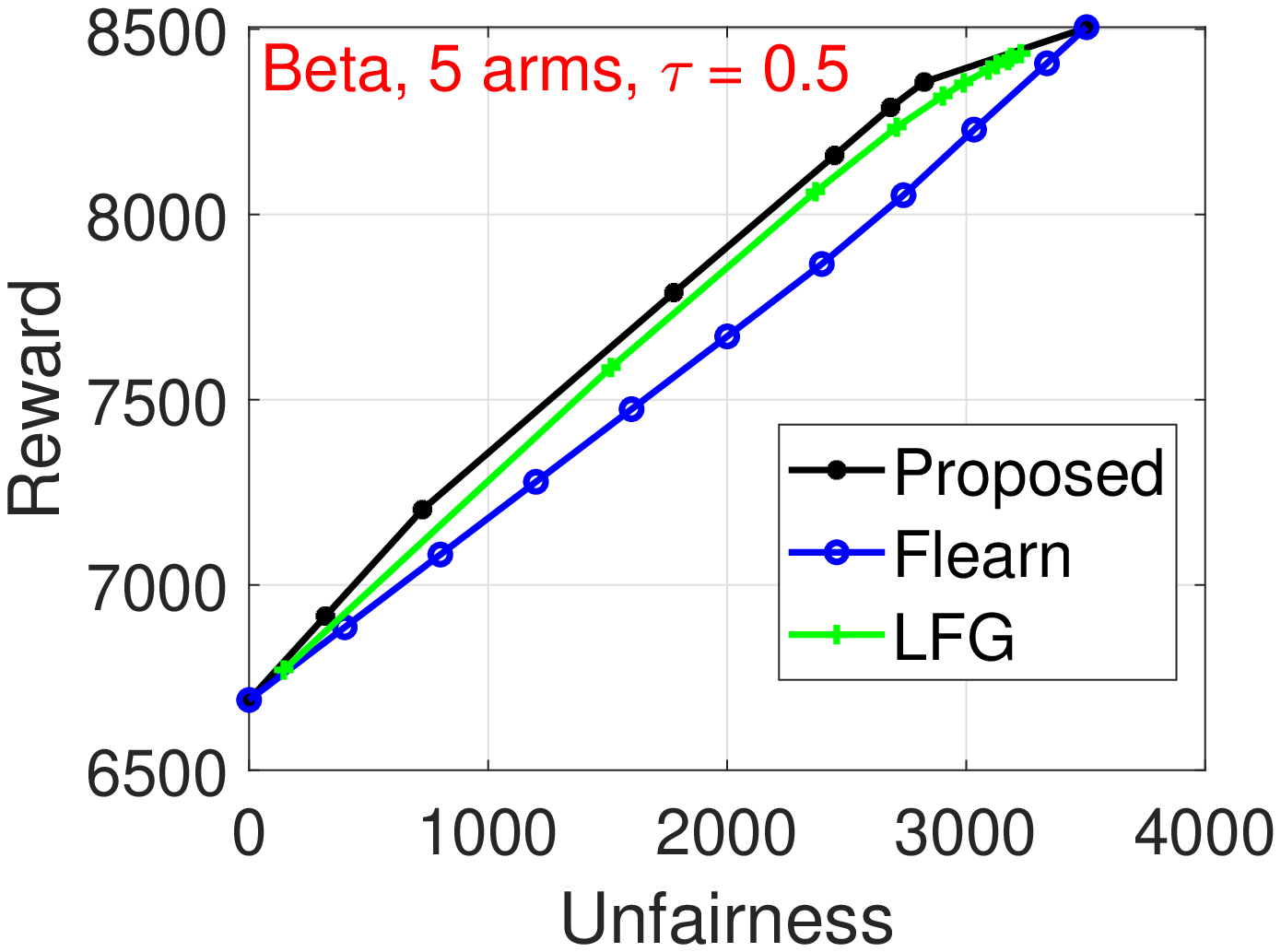}\hspace{-0.12in}
		\includegraphics[width=2.35in]{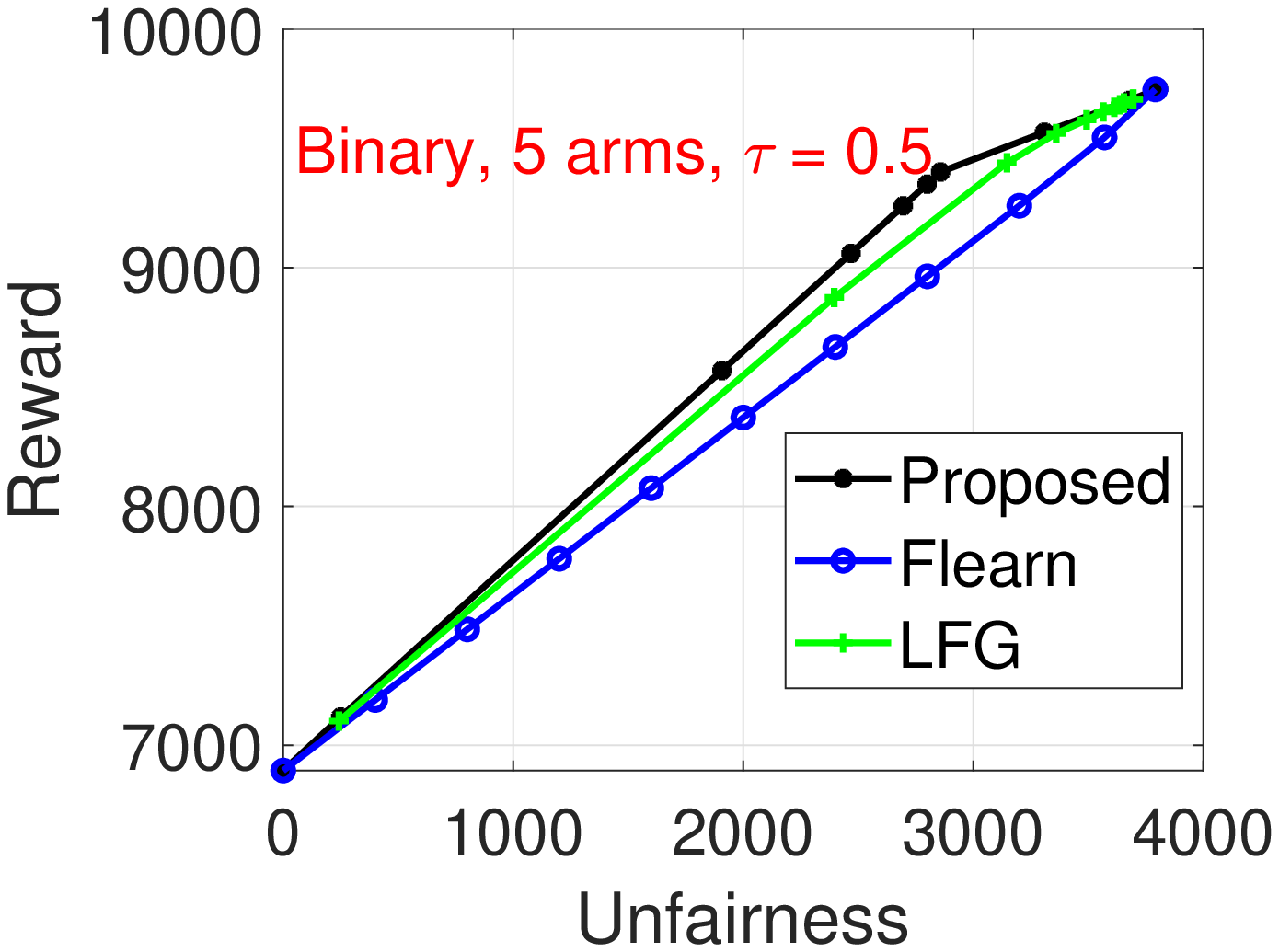}
	}

	\mbox{\hspace{-0.2in}
		\includegraphics[width=2.35in]{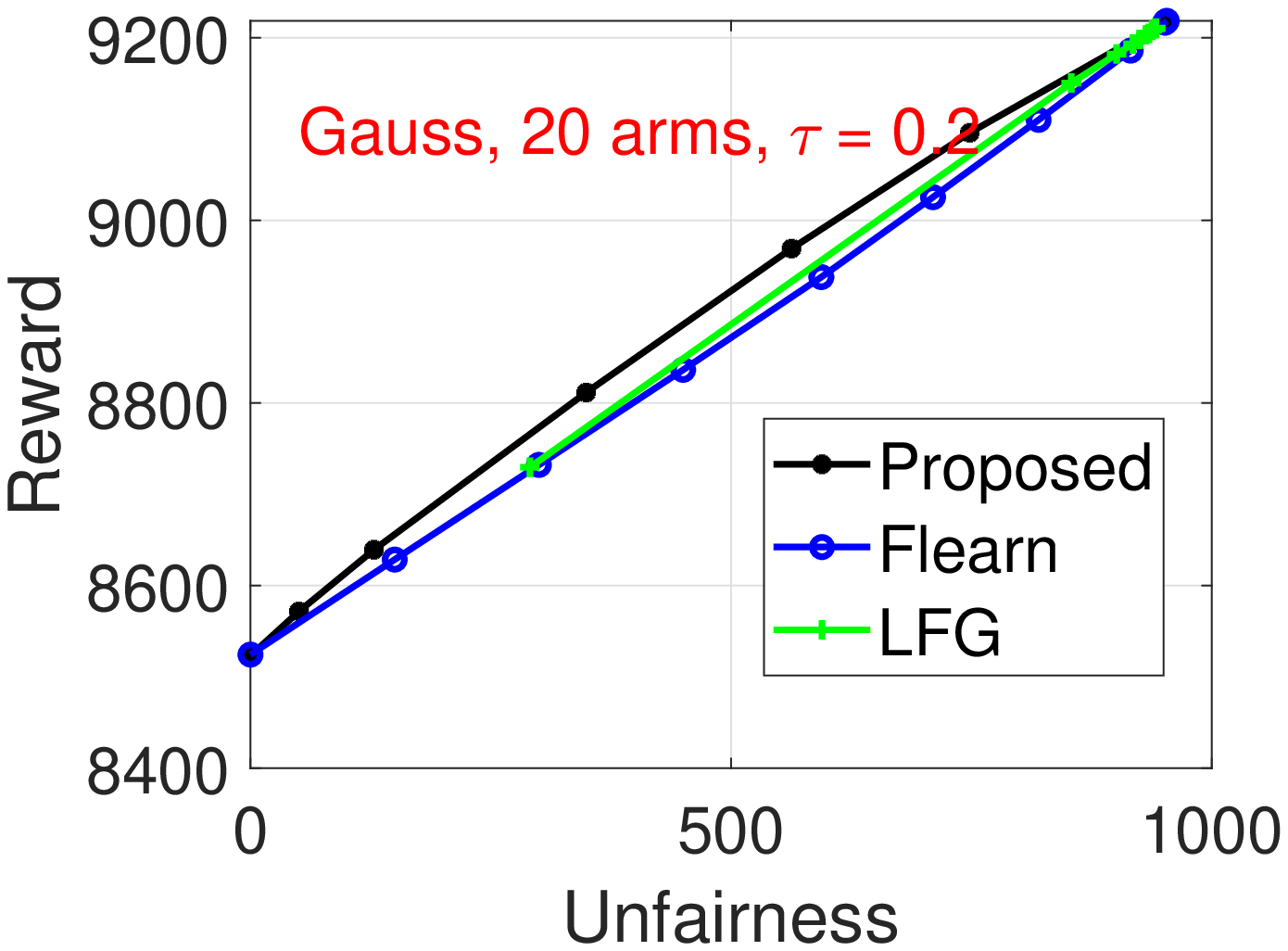}\hspace{-0.12in}
		\includegraphics[width=2.35in]{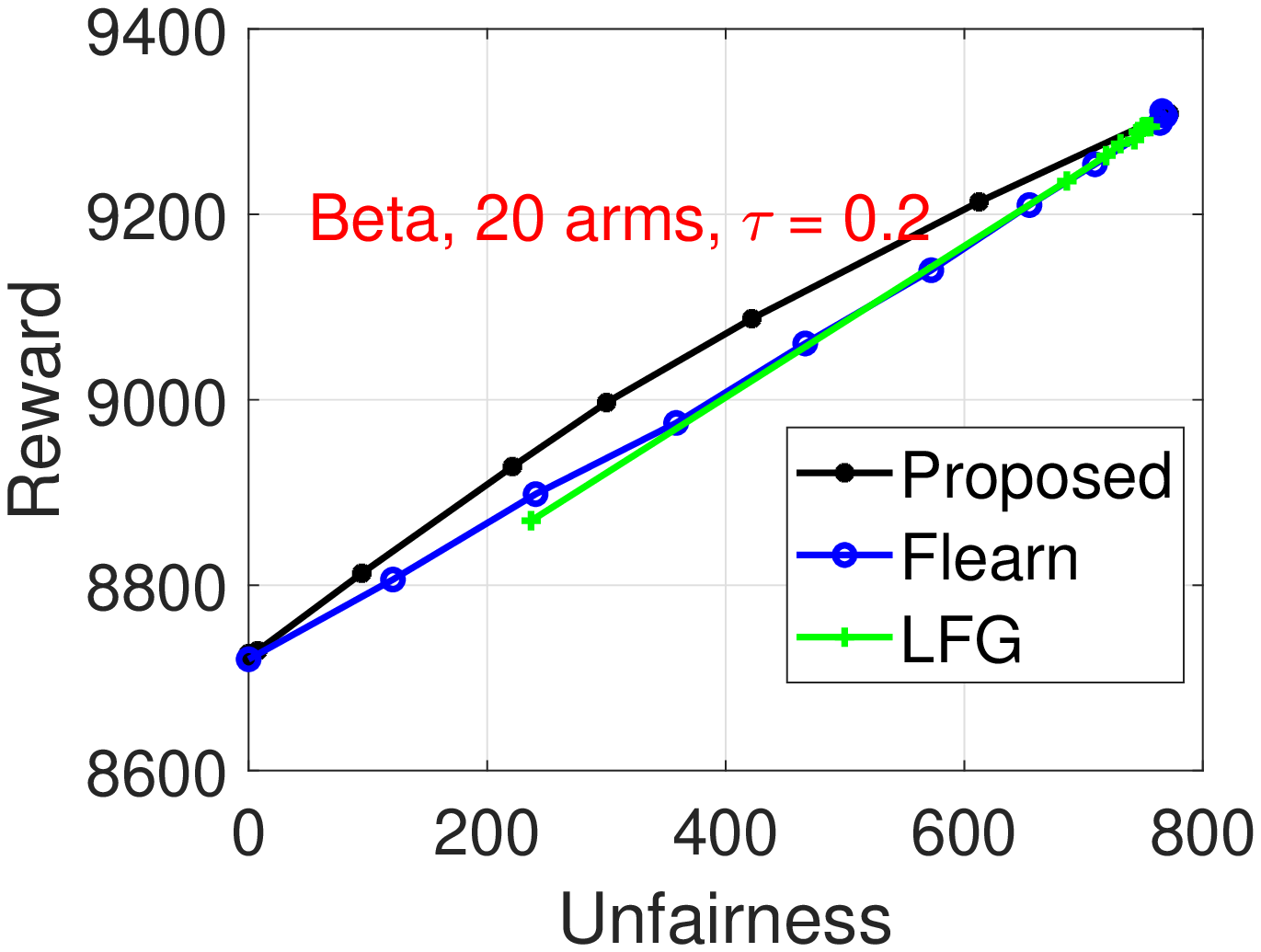}\hspace{-0.12in}
		\includegraphics[width=2.35in]{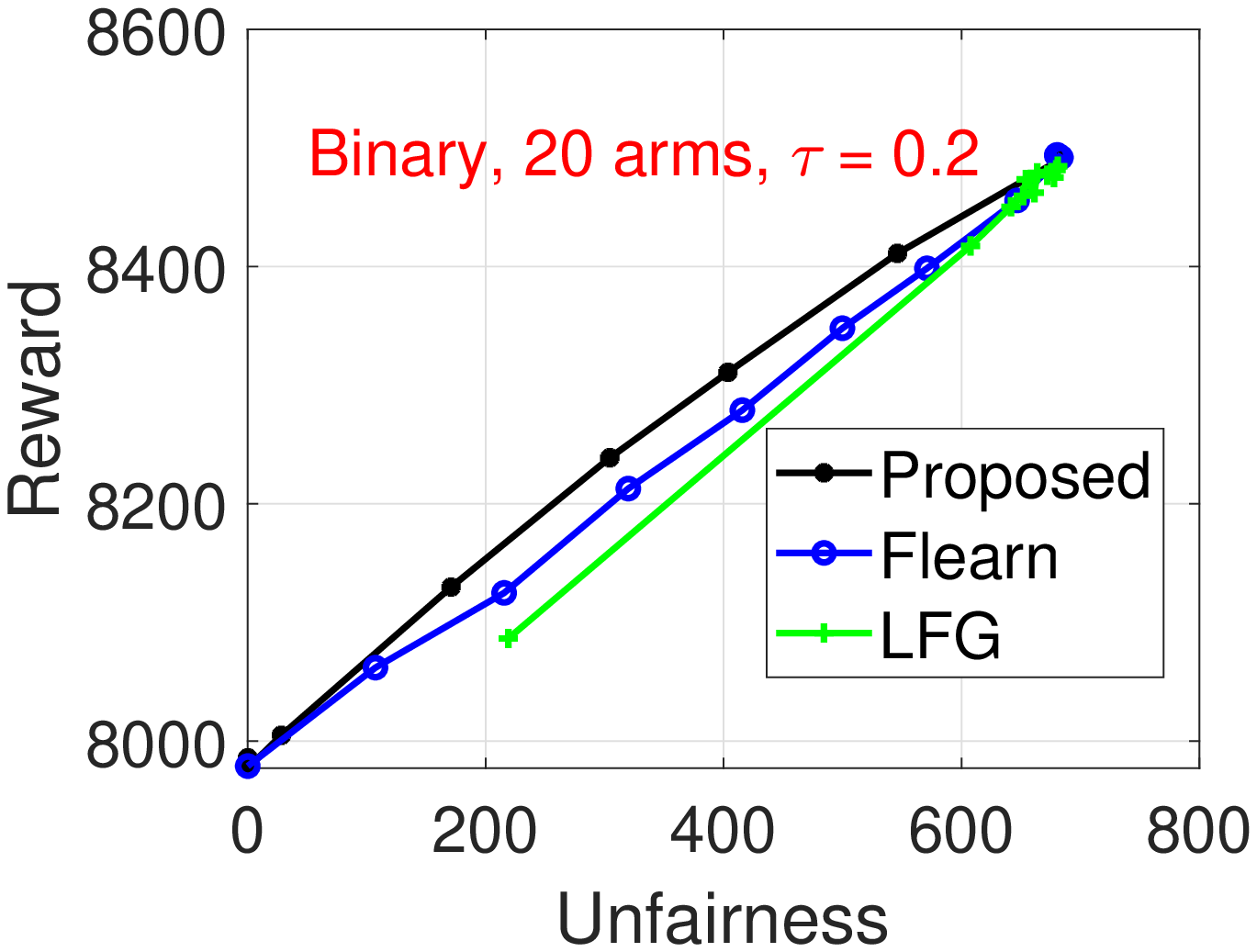}
	}

	\mbox{\hspace{-0.2in}
		\includegraphics[width=2.35in]{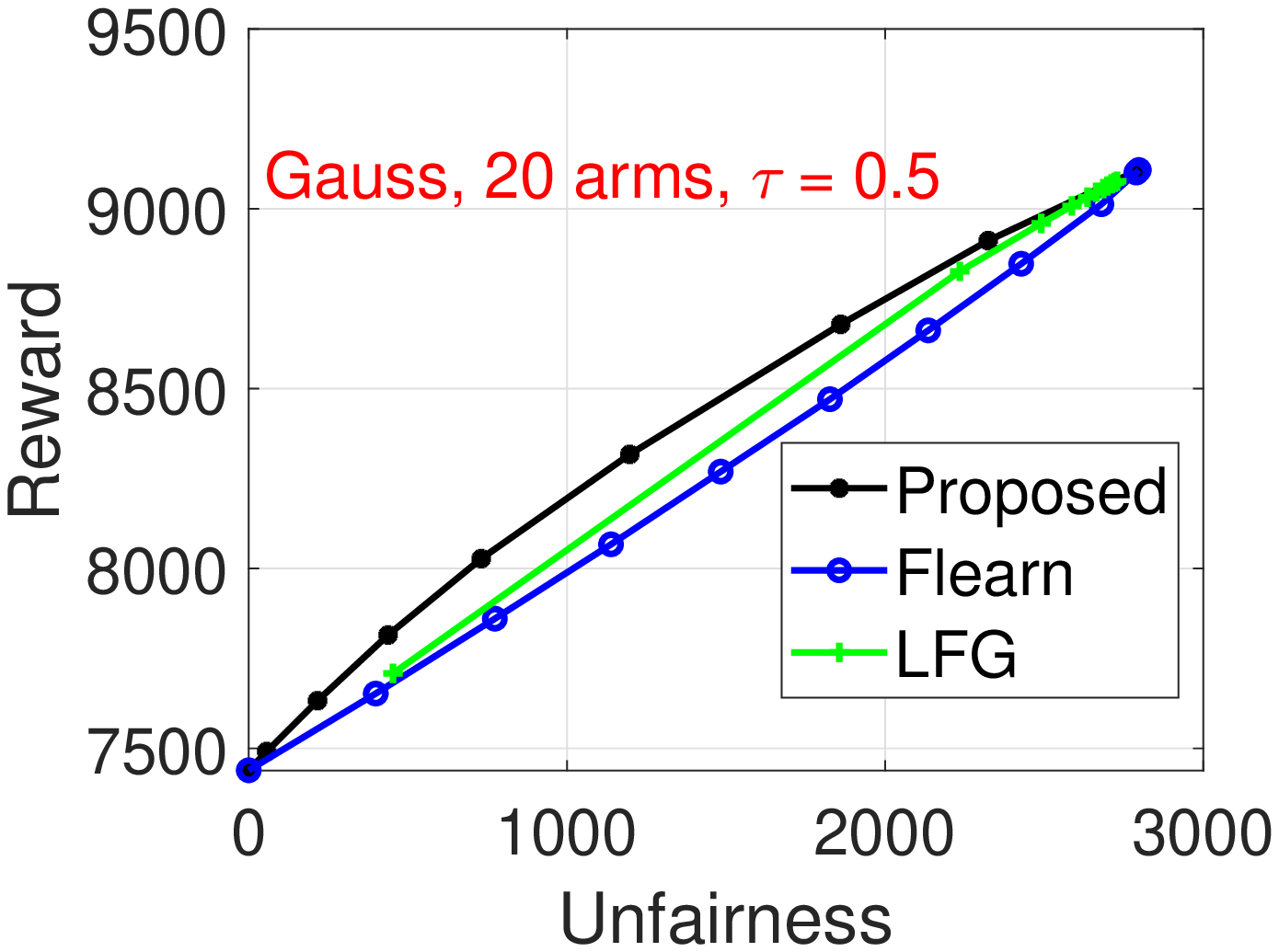}\hspace{-0.12in}
		\includegraphics[width=2.35in]{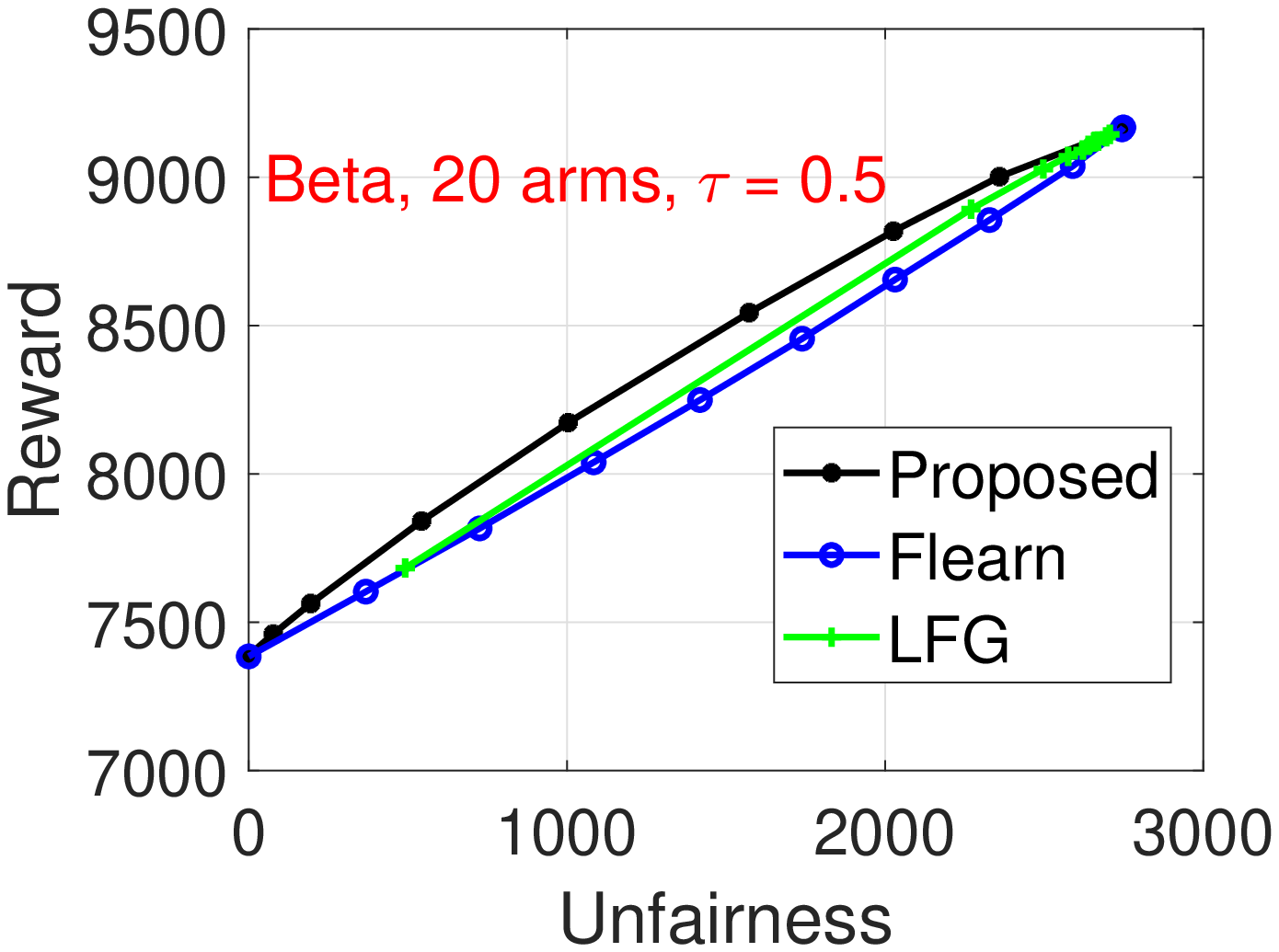}\hspace{-0.12in}
		\includegraphics[width=2.35in]{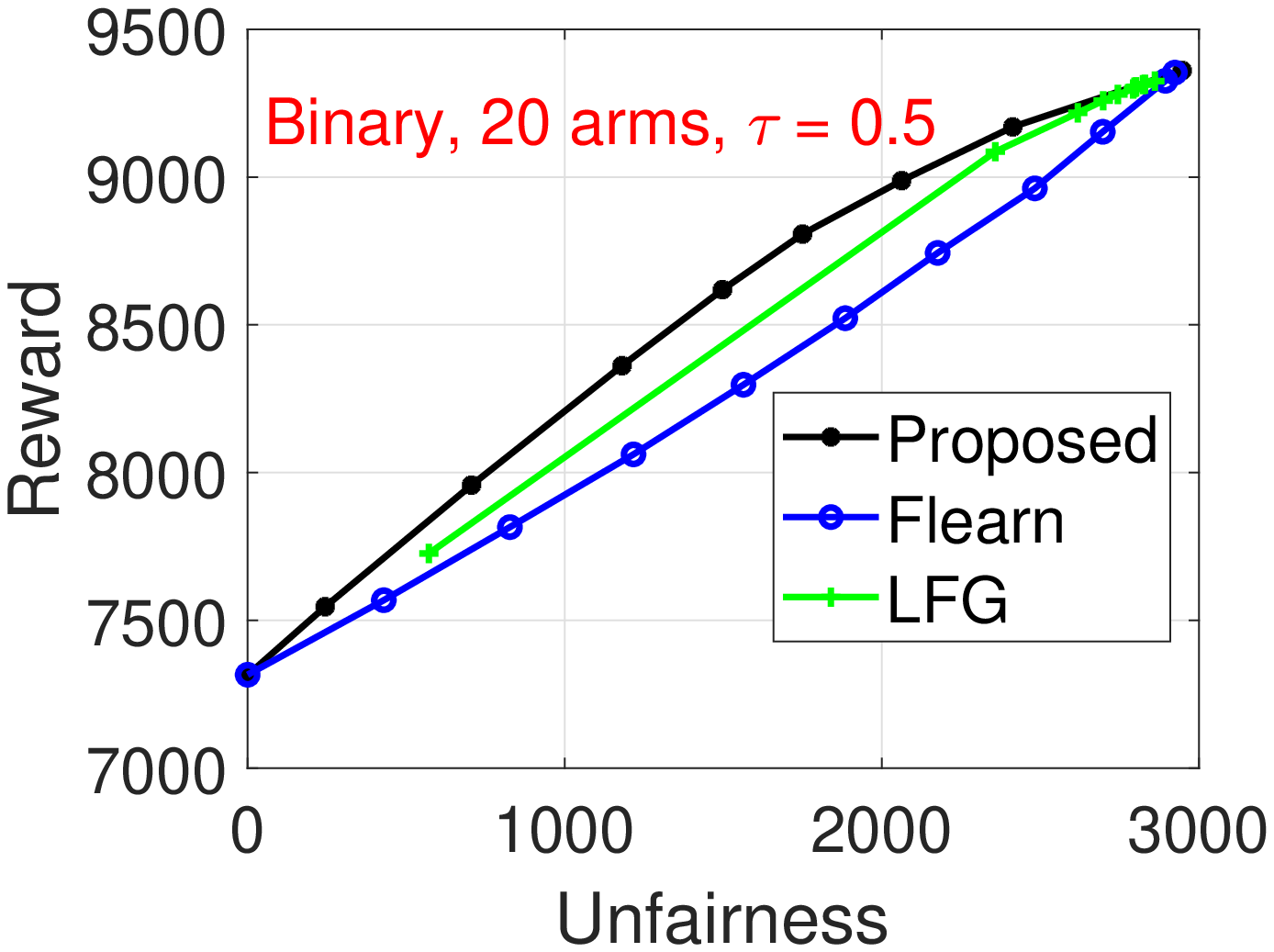}
	}
	
	\caption{Total reward vs unfairness level for three algorithms under different generation mechanisms as described in Setting 5.
		(The first row is for 5 arms with required fraction of times $\tau = 0.2$; The second row is for 5 arms with required fraction of times $\tau = 0.5$; The third row is for 20 arms with required fraction of times $\tau = 0.2$; The fourth row is for 20 arms with required fraction of times $\tau = 0.5$. The first column is for Gaussian reward distribution; the second column is for Beta distribution; and the third column is for Bernoulli distribution.)
		Given the fixed unfairness level, the proposed method can have larger total reward than other two methods consistently over all experimental settings.}\label{fig:setE}
\end{figure*}

The relationship between total expected reward ($\sum_{t=1}^T \mu_{\pi_t}$) and unfairness level ($\sum_{k \in [K]} (\tau_k T - N_k(T))_{+}$) for three algorithms are illustrated in Figure \ref{fig:setE}.
All results confirm that our proposed method is superior in balancing between cumulative rewards and fairness requirements.

\section{Conclusion}
In this paper, we provide a new framework of MAB problems by introducing regularization terms. 
The advantage of our new approach is that it allows the user to distinguish between arms for which is more important to sample an arm with required frequency level and arms for which it is less important to do so.
A hard-threshold UCB algorithm is proposed and has been shown to have good performance under this framework.
Unlike other existing algorithms, the proposed algorithm not only achieves the asymptotic fairness but also handles well in balancing between reward and fairness constraints. A relatively complete theory, including both gap-dependent / independent upper and lower bounds, has been established. 
Our new theoretical results significantly contribute to the bandit problems in machine learning field and bring better insights in how to play smartly in the exploitation and exploration games. 

\newpage

\clearpage

\bibliographystyle{plainnat}
\bibliography{refs_scholar}

\newpage

\appendix

\section*{Overview of Appendix}

We collect all technical proofs in this appendix. Specifically, the proofs of gap-dependent upper and lower bounds are given in Section \ref{app:gap:upper} and \ref{app:gap:lower}. 
The proofs of gap independent upper and lower bounds are given in Section \ref{app:ind:upper} and \ref{app:ind:lower}, respectively.
The proof of $\mathbb E[\max_{1 \leq t \leq T}(\tau_k t - N_k(t))]_{+}$ is given in Section \ref{app:max}. 

\section{Proof of Gap-dependent Upper Bounds}\label{app:gap:upper}

\textbf{Proof of Theorem} \ref{thm:fairness} Its proof is essentially same as the proof of first part in Theorem \ref{thm:dep:upper} by treating the non-critical set $\mathcal A_{\text{non-cr}}$ empty.

\vspace{0.2in}

\noindent\textbf{Proof of Theorem \ref{thm:dep:upper}}
We first prove the first part:
$\mathbb E[ (\tau_{k} T - N_{k}(T))_{+} ] = O(1)$ for any $k \in \mathcal A_{\text{opt}} \cup \mathcal A_{\text{cr}}$.

Suppose at time $n$ that a critical arm $k$ is played less than $\tau_k n$.  
We can prove that the algorithm pulls critical arm $k'$ at time $n$ such that $N_{k'}(n) \geq  8 \log T /c_a^2$ and $N_{k'}(n) > \tau_k n$ with vanishing probability.
This is because
\begin{eqnarray}
& & \mathbb P(A_k + m_k(n) + \sqrt{\frac{2 \log n}{N_k(n)}} \leq 
m_{k'}(n) + \sqrt{\frac{2\log n}{N_{k'}(n)}}) \nonumber \\
&\leq& \mathbb P(A_k + \mu_k \leq + \mu_k' + 2 \sqrt{\frac{2\log n}{N_{k'}(n)}}) + \frac{2}{n^2} \nonumber \\
&\leq& \mathbb P(A_k - \Delta_k \leq - \Delta_{k'} + 2 \sqrt{\frac{2\log n}{N_{k'}(n)}}) + \frac{2}{n^2} \nonumber \\
&=& 2 / n^2.
\end{eqnarray} 
By the same reason, the algorithm pulls non-critical arm $k''$ at time $n$ when $N_{k''}(n) \geq 8 \log T / c_a^2$ with vanishing probability.  

(Observation 3) In other words, it holds with high probability that once a critical arm $k$ is played with proportion less than required level $\tau_k$'s, it must be pulled in next round when all other arms is played with proportion greater than level $\tau_k$'s and is played more than $8 \log T/c_a^2$ times.

(Observation 4) It also holds with high probability that once a non-critical arm is played more than $8 \log T/c_a^2$, it can be only played when all critical arms are played with frequency more than the required level $\tau_k$'s.

Moreover, we can show that $N_{k'}(n) \geq 8 \log T /c_a^2$ at time $n = c_0 T / 2$ for each critical arm $k'$. If not, note that $8 \log T /c_a^2 \leq \tau_{k'} c_0 T / 4$, then $N_{k}'(n) < \tau_k' n$ for any $n \in \{\lceil c_0 T/4 \rceil, \ldots, \lfloor c_0 T / 2 \rfloor \}$. Hence, for any critical arm $k'$ can be played at most $\max\{\tau_{k'} c_0 T / 4, 8 \log T /c_a^2\}$ times between rounds $c_0 T/2$ and $c_0 T$; 
every non-critical arm $k''$ can be played at most $8 \log T /c_a^2$ times. 
Then, we must have 
\[c_0 T/2 - c_0 T/4 \leq \sum_k \tau_k c_0 T / 4 + \sum_k 8 \log T /c_a^2.\]
However, the above inequality fails to hold when $T/ \log T \geq 16 c_0^2 K / c_a^2$. This leads to the contradiction.
Thus, we have $N_{k'}(n) \geq 8 \log T /c_a^2$ for any critical arm $k'$ at time $n = c_0 T / 2$.

\newpage

Actually, this further gives us that we must have $N_{k'}(n) \geq \lfloor \tau_{k'} n \rfloor$ for all critical arms at some time $n \in [c_0 T / 2, T]$. 
To see this,
we observe the fact that for any arm $\bar k$, it will be played with probability less than $\frac{2}{T^2}$ at time $n$ once $N_{\bar k}(n) \geq \max\{\tau_k n, 8 \log T /c_a^2\}$ and one critical arm $k'$ is played less than $\tau_{k'} n$. 
(In other words, this tells us that once arm $\bar k$ has been played $\max \{\tau_{\bar k} n, 8 \log T /c_a^2\}$ times, then it can only be played at time when all critical arms $k'$s have been played for $\tau_{k'} n$ times or $\lfloor \tau_{\bar k} n \rfloor$ jumps by one with probability greater than $1 - 2 / T^2$.)

Let $n_1 (\geq c_0 T / 2)$ be the first ever time such that 
$N_{k'}(n_1) \geq \lfloor \tau_{k'} n_1 \rfloor$ for all critical arms $k'$s.
By straightforward calculation, it gives that $n_1$ must be bounded by 
\[n_1 \leq c_0 T / 2 + \sum_{k: \text{non-critical}} 8 \log T /c_a^2 + (\sum_{k: \text{critical}} \tau_k) T\]
with probability greater than $1 - 2 K/T$.
That is, $n_1$ is well defined between $c_0 T / 2$ and $T$. At time $n_1$, we have all critical arms $k'$ such that $N_{k'}(n_1) \geq \tau_{k'} n_1$.

Moreover, we consider the first time $n_2 (> n_1)$ such that every non-critical arm $k''$ has been played for at least $8 \log T /c_a^2$ times when $\Delta_{k''} - A_{k''} \leq c_a \sqrt{\frac{\log (c_0T/2)}{16 \log T}}$ (which is asymptotically $c_a / 4$).
We claim that $n_2 \leq n_1 + c_0 T / 2$.
This is because, between rounds $n_1$ and $n_2$,
the algorithm will choose non-critical arm $k''$ when $N_{k'}(n) \geq \tau_{k'}(n)$ for all critical arms $k'$s and $N_{k''}(n) \leq \min\{\log(c_0 T/2) / 2( \Delta_{k''} - A_{k''})^2, \tau_{k''} c_0 T / 2 \}$.
To see this, we know that 
\begin{eqnarray}
& & \mathbb P(m_{k'}(n) + \sqrt{\frac{2\log n}{N_{k'}(n)}} \geq m_{k''}(n) + A_{k''} + \sqrt{\frac{2 \log n}{N_{k''}(n)}}) \nonumber \\
& \leq & \mathbb P(\mu_k' + 2 \sqrt{\frac{2\log n}{N_{k'}(n)}} \geq \mu_{k''} + A_{k''} +  \sqrt{\frac{\log n}{N_{k''}(n)}}) + 2/n \nonumber \\
&\leq& \mathbb P(\mu_k' + 2 \sqrt{\frac{2\log T}{\tau_{k'}c_0 T/2}} \geq \mu_{k''} + A_{k''} + \sqrt{\frac{\log(c_0T/2)}{N_{k''}(n)}}) + 2 / n \nonumber \\
&\leq& \mathbb P(\Delta_{k''} - A_{k''} + 2 \sqrt{\frac{2\log T}{\tau_{k'}c_0 T/2}} \geq \sqrt{\frac{\log(c_0T/2)}{N_{k''}(n)}}) + 2 / n \nonumber \\
&\leq& \mathbb P(\Delta_{k''} - A_{k''}  \geq \sqrt{\frac{\log(c_0T/2)}{2 N_{k''}(n)}}) + 2 / n \nonumber \\
&\leq& 4 / c_0 T.
\end{eqnarray}
for $n \geq c_0 T / 2$.
That is, index of arm $k''$ is larger than $k'$ with high probability. 

In other words, for each round between $n_1$ and $n_2$, each critical arm $k'$ can be only pulled at most $\tau_{k'}(n_2 - n_1)$ before every non-critical arm $k''$ has been played for
$\min\{8 \log T/c_a^2, \log(c_0 T/2) / 2 ( \Delta_{k''} - A_{k''})^2\}$ ($\equiv 8 \log T/c_a^2$ when $\Delta_{k''} - A_{k''} < c_a / 4$).
Additionally, each non-critical arm $k''$ can be only played for at most $8 \log T / (\Delta_k - A_k)^2$ with high-probability.
Therefore, it must hold that 
\[ n_2 - n_1 \leq (\sum_k \tau_k)(n_2 - n_1) + \sum_k 8 \log T / (\Delta_k - A_k)^2.\]
However, the above inequality fails to hold when $n_2 - n_1 \geq c_0 T/2$ under assumption that $\Delta_k - A_k \geq \sqrt{\frac{8 K \log T}{c_0^2 T}}$.
This validates the claim $n_2 \leq n_1 + c_0T/2$.

\newpage

Starting from time $n_2$, by the observations 3 and 4, 
it can be seen that the maximum values of $(\tau_{k'} n - N_{k'}(n))_{+}$ for any critical arm $k'$ is always bounded by 1 with probability $1 - 2K/T$ ($n \in [n_2, T]$). This completes the proof of the first part.

For the second part, 
we need to prove $\mathbb E [N_{k}(T)] \leq \max\{\frac{8 \log T}{\Delta_{k}^2}, \tau_{k} T\} + O(1)$ for $k \in \mathcal A_{\text{cr}}$.

When $\frac{8 \log T}{\Delta_k^2} > \tau_k T$, we can calculate the probability 
\begin{eqnarray}
& & \mathbb P(\textrm{arm $k$ is pulled at round $n+1$}~ | N_k(n) \geq \frac{8 \log T}{\Delta_k^2}) \nonumber \\
&\leq& \mathbb P(i_k(n+1) \geq i_{k^{\ast}}(n+1)) \nonumber \\
&\leq& \mathbb P(\hat m_k(n+1) + \sqrt{\frac{2 \log (n+1)}{N_k(n)}} \geq \hat m_{k^{\ast}}(n+1) + \sqrt{\frac{2 \log (n+1)}{N_k(n)}})  \nonumber \\
&\leq& 1 / n^2 \leq 1 / (8 \log T/\Delta_k)^2 \leq 1 / (\tau_k T)^2.
\end{eqnarray}

When $\frac{8 \log T}{\Delta_k^2} \leq \tau_k T$, we can similarly calculate the probability
\begin{eqnarray}
& & \mathbb P(\textrm{arm $k$ is pulled at round $n+1$}~ | N_k(n) \geq \tau_k T) \nonumber \\
&\leq& \mathbb P(i_k(n+1) \geq i_{k^{\ast}}(n+1)) \nonumber \\
&\leq& \mathbb P(\hat m_k(n+1) + \sqrt{\frac{2 \log (n+1)}{N_k(n)}} \geq \hat m_{k^{\ast}}(n+1) + \sqrt{\frac{2 \log (n+1)}{N_k(n)}})  \nonumber \\
&\leq& 1 / n^2 \leq 1 / (\tau_k T)^2.
\end{eqnarray}
Hence we can easily obtain that $\mathbb E[N_k(T)] \leq \max\{\frac{8 \log T}{\Delta_k^2}, \tau_k T\} + O(1)$ by union bound.

For the third part that $\mathbb E[N_{k}(T)] \leq \min\{ \frac{8 \log T}{(\Delta_{k} - A_{k})^2}, \tau_{k} T \} + O(1)$ ($k_j \in \mathcal A_{\text{non-cr}}$), it follows from the fact that we can treat $\mu_k + A_k$ as new expected reward for arm $k \in \mathcal A_{\text{non-cr}}$. Thus the corresponding sub-optimality gap is $\Delta_k - A_k$. The result follow by using standard technique in the classical UCB algorithm. Hence we omit the details here.

Finally, by combining three parts and straightforward calculation, we obtain the desired gap-dependent upper bounds. This concludes the proof.

\newpage

\section{Proof of Gap-dependent Lower Bounds}\label{app:gap:lower}

\textbf{Proof of Theorem \ref{thm:lower:dep1}}.~
We consider the following setting, 
where arm 1 is the optimal arm with a deterministic reward $\Delta$ 
and arms $k$, $(k \geq 2)$ are sub-optimal arms with reward zero. 
Let penalty rate $A_k = A$ for all $k \in [K]$ with $\Delta > A$.
Assuming that 
$\frac{8 \log T}{(\Delta - A)^2} \leq \tau_k T / 2$,
we construct a lower bound as follows.

We claim that each arm $k \geq 2$ will be played at least
$n_1 := \frac{\log T}{(\Delta - A)^2}$ times.
If there exists an arm $k_0$ has not been played for $n_1$ times,
we then consider the time index $n_a = T/2 + 1 + (K-2) \frac{8 \log T}{(\Delta - A)^2} + n_1$.
At this time, we have that arm $1$ is the arm with largest index since that for each sub-optimal arm $k \neq k_0$, its index will never exceeds $\Delta$  once it has been played 
$\frac{8 \log T}{(\Delta - A)^2}$ times.
According to assumption that arm $k_0$ has been played less than $n_1$ times, thus arm 1 is the arm with largest index at time $n_a$.


However, the index of arm 1 at time $n_a$ is never larger than 
$\sqrt{\frac{2 \log T}{T/2}} + \Delta$.
The index of arm $k_0$ at time $n_a$ is always larger than 
$A + \sqrt{\frac{2 \log (T/2)}{n_1}}$.
It gives 
\begin{eqnarray}
i_1(n_a) \leq \sqrt{\frac{2 \log T}{T/2}} + \Delta < A + \sqrt{\frac{2 \log (T/2)}{n_1}} \leq i_{k_0}(n_a),
\end{eqnarray}
which leads to the contradiction of the mechanism of the proposed algorithm.
Hence, we have that each sub-optimal should have been played for at least $\frac{\log T}{(\Delta - A)^2}$ times.

\vspace{0.2in}
\noindent\textbf{Proof of Theorem \ref{thm:lower:dep2}}.~
We consider  another setting, 
where arm 1 is the optimal arm with deterministic reward $\Delta_1 + \Delta_2$, arm $k$'s $(k \in \mathcal A_{cr})$ are sub-optimal arms with reward being $\Delta_1$ and 
arm $k$'s $(k \in \mathcal A_{non-cr})$ ar sub-optimal arms with reward being $\Delta_2$.
Let penalty rate $A_k = A_2$ for all $k \in \mathcal A_{cr}$ with $\Delta_2 < A_2$ and 
penalty rate $A_k = A_1$ for all $k \in \mathcal A_{non-cr}$ with
$\Delta_1 > A_1$.
Assume that 
$\sum_{k \in \mathcal A_{non-cr}} \frac{8 \log T}{(\Delta_1 - A_1)^2}
+ \sum_{k \in \mathcal A_{cr}} \frac{8 \log T}{\Delta_2^2}  < T / 2$
and $\tau_k T \leq \frac{\log T}{\Delta_2^2}$ for $k \in \mathcal A_{cr}$,
we then have the following lower bound.

We claim that for each arm $k \in \mathcal A_{cr}$ will be played for at least $n_2 := \frac{\log T}{\Delta_2^2}$ times.
If not, there will be at least one arm $k_1 \in \mathcal A_{cr}$ has been played for less than $n_2$ times.
We consider the time stamp, $n_b = T/2 + 1 + \sum_{k \in \mathcal A_{non-cr}} \frac{8 \log T}{(\Delta_1 - A_1)^2} + \sum_{k \in \mathcal A_{cr}; k \neq k_1} \frac{8 \log T}{\Delta_2^2} + n_2$. 
At this time, we have that arm 1 is the arm with the largest index since that for each arm in $\mathcal A_{non-cr}$, its index is always smaller than $\Delta_1 + \Delta_2$ once it has been played for 
$\frac{8 \log T}{(\Delta_1 - A_1)^2}$ times.
For each arm $k \in \mathcal A_{cr}$ ($k \neq k_1$), its index is also smaller than $\Delta_1 + \Delta_2$ once it has been played for 
$\frac{8 \log T}{\Delta_2^2}$ times.
According to assumption that arm $k_1$ has been played less than $n_2$ times, thus arm 1 is the arm with largest index at time $n_b$.

However, on other hand, the index of arm 1 at time $n_b$ is never larger than $\sqrt{\frac{2 \log T}{T/2}} + \Delta_1 + \Delta_2$.
The index of arm $k_1$ is not smaller than $\Delta_1 + \sqrt{\frac{2 \log (T/2)}{n_2}}$.
It leads to 
\[i_1(n_b) \leq \sqrt{\frac{2 \log T}{T/2}} + \Delta_1 + \Delta_2 \leq \Delta_1 + \sqrt{\frac{2 \log (T/2)}{n_2}} \leq i_2(n_b),\] 
this contradicts with arm 1 is arm with largest index at time $n_b$.
Hence, any arm in $\mathcal A_{cr}$ should be played at least $\frac{\log T}{\Delta_2^2}$ times.

\section{Proof of Maximal Inequality (Proof of Theorem \ref{thm:maximal})}\label{app:max}

We can order $K$ arms according to the sums $\mu_k+A_k$'s. Specifically, let
the order $k_1,k_2,\ldots, k_K$ be defined by
\begin{equation} \label{e:order.arms}
\mu_{k_1}+A_{k_1}>\mu_{k_2}+A_{k_2}>\cdots >\mu_{k_K}+A_{k_K}.
\end{equation}
For simplicity we assume no ties in \eqref{e:order.arms}. We also
assume that $A_k>\Delta_k$ for all $k\in \calA_{\rm
	opt}\cup \calA_{\rm   cr}$.  


We now aim to bound expectations of the $\mathbb E \max_{t \in [T]}\bigl(
\tau_k t - N_k(t)\bigr)_+ $ for $k\in \calA_{\rm opt} \cup \calA_{\rm
	cr}$. 
We will use the ordering of the arms $k_1,k_2,\ldots, k_K$
defined in \eqref{e:order.arms}. 
Take any arbitrary $t \in [T]$ and let $k_j\in \calA_{\rm
	opt} \cup \calA_{\rm cr}$,
\begin{equation} \label{e:mT}
m_t^{(j)}=\sup\bigl\{ n=1,\ldots, t:\, \tau_{k_j}n\leq
N_{k_j}(n)\bigr\}.
\end{equation}
Suppose for a moment that $m_t^{(j)}<t$. 
We have
\begin{align} \label{e:split.deficit}
&\bigl(   \tau_{k_j}t -N_{k_j}(t)\bigr)_+ \leq \tau_{k_j} \\
\notag  +& \tau_{k_j} \#\bigl\{
n=m_t^{(j)}+1,\ldots, t:\, \tau_{k_d}n>N_{k_d}(n-1) \ \text{for some}
\ d=1,\ldots, j-1\bigr\} \\
\notag +& \tau_{k_j} \#\bigl\{
n=m_t^{(j)}+1,\ldots, t:\, \tau_{k_d}n\leq N_{k_d}(n-1) \ \text{for all}
\ d=1,\ldots, j-1, \\
\notag  & \hskip 3.1in \text{arm $k_j$ not pulled at time $n$}\bigr\} \\
\notag -&(1-\tau_{k_j}) \#\bigl\{
n=m_t^{(j)}+1,\ldots, t:\, \tau_{k_d}n\leq N_{k_d}(n-1) \ \text{for all}
\ d=1,\ldots, j-1, \\
\notag  & \hskip 3.1in \text{arm $k_j$  pulled at time $n$}\bigr\} \\
\notag =& \tau_{k_j} + \tau_{k_j} \#\bigl\{
n=m_t^{(j)}+1,\ldots, t:\, \tau_{k_d}n>N_{k_d}(n-1) \ \text{for some}
\ d=1,\ldots, j-1\bigr\} \\
\notag -&(1-\tau_{k_j}) \#\bigl\{
n=m_t^{(j)}+1,\ldots, t:\, \tau_{k_d}n\leq N_{k_d}(n-1) \ \text{for all}
\ d=1,\ldots, j-1\bigr\} \\
\notag +& \#\bigl\{
n=m_t^{(j)}+1,\ldots, t:\, \tau_{k_d}n\leq N_{k_d}(n-1) \ \text{for all}
\ d=1,\ldots, j-1, \\
\notag  & \hskip 3.1in \text{arm $k_j$ not pulled at time $n$}\bigr\} \\
\notag =& \tau_{k_j} +  \#\bigl\{
n=m_t^{(j)}+1,\ldots, t:\, \tau_{k_d}n>N_{k_d}(n-1) \ \text{for some}
\ d=1,\ldots, j-1\bigr\} \\
\notag +& \#\bigl\{
n=m_t^{(j)}+1,\ldots, t:\, \tau_{k_d}n\leq N_{k_d}(n-1) \ \text{for all}
\ d=1,\ldots, j-1, \\
\notag  & \hskip 3.1in \text{arm $k_j$ not pulled at time $n$}\bigr\} \\
\notag -& { (1-\tau_{k_j}) (t-m_t^{(j)})}. 
\end{align}
The final bound is, clearly, also valid in the case $m_t^{(j)}=t$. 

Next,
\begin{align} \label{e:count.1}
&\#\bigl\{
n=m_t^{(j)}+1,\ldots, t:\, \tau_{k_d}n>N_{k_d}(n-1) \ \text{for some}
\ d=1,\ldots, j-1\bigr\}\\
=\sum_{d=1}^{j-1} &\#\bigl\{
n=m_t^{(j)}+1,\ldots, t:\, \tau_{k_d}n>N_{k_d}(n-1), \,
\tau_{k_m}n\leq N_{k_m}(n-1) , \, m=1,\ldots, d-1\bigr\}.\notag 
\end{align}
For $d=1,\ldots, j-1$ denote
\begin{equation} \label{e:mT.2}
m_t^{(j,d)}=\sup\bigl\{ n=m_t^{(j)},\ldots, t:\, \tau_{k_d}n>
N_{k_d}(n-1)\bigr\}.
\end{equation}
Suppose, for a moment, that $m_t^{(j,d)}>m_t^{(j)}$. Then
\begin{align*}
0<\tau_{k_d}m_t^{(j,d)}-&N_{k_d}\bigl( m_t^{(j,d)}-1\bigr)
= \tau_{k_d}m_t^{(j)}-N_{k_d}\bigl( m_t^{(j)}-1\bigr)\\
+& \tau_{k_d}\Bigl( m_t^{(j,d)}-m_t^{(j)} -\#\bigl\{
n=m_t^{(j)}+1,\ldots, m_t^{(j,d)}:\ \text{arm $k_d$
	pulled}\bigr\}\Bigr)\\
-&(1-\tau_{k_d})\#\bigl\{
n=m_t^{(j)}+1,\ldots, m_t^{(j,d)}:\ \text{arm $k_d$
	pulled}\bigr\}\\
=&\tau_{k_d}m_t^{(j)}-N_{k_d}\bigl( m_t^{(j)}-1\bigr)
+\tau_{k_d}\bigl( m_t^{(j,d)}-m_t^{(j)} \bigr)\\
-& \#\bigl\{
n=m_t^{(j)}+1,\ldots, m_t^{(j,d)}:\ \text{arm $k_d$
	pulled}\bigr\}.
\end{align*}
We conclude that
\begin{align*} 
&\#\bigl\{
n=m_t^{(j)}+1,\ldots, m_t^{(j,d)}:\ \text{arm $k_d$
	pulled}\bigr\}\\
\leq &\max_{n=1,\ldots, t} \bigl( \tau_{k_d}n-N_{k_d}(n)\bigr)_+
+ \tau_{k_d}\bigl( m_t^{(j,d)}-m_t^{(j)} \bigr).  
\end{align*}
Therefore, 
\begin{align*}
&\#\bigl\{
n=m_t^{(j)}+1,\ldots, t:\, \tau_{k_d}n>N_{k_d}(n-1), \,
\tau_{k_m}n\leq N_{k_m}(n-1) , \, m=1,\ldots, d-1\bigr\}\\
=&\#\bigl\{
n=m_t^{(j)}+1,\ldots, m_t^{(j,d)}:\, \tau_{k_d}n>N_{k_d}(n-1), \,
\tau_{k_m}n\leq N_{k_m}(n-1) , \, m=1,\ldots, d-1\bigr\}\\
\leq& \#\bigl\{
n=m_t^{(j)}+1,\ldots, m_t^{(j,d)}:\ \text{arm $k_d$
	pulled}\bigr\} \\
+& \#\bigl\{
n=m_t^{(j)}+1,\ldots, t:\, \tau_{k_d}n>N_{k_d}(n-1), \,
\tau_{k_m}n\leq N_{k_m}(n-1) , \, m=1,\ldots, d-1, \\
&\hskip 4.5in   \text{arm $k_d$ not pulled} \bigr\}\\
\leq& \max_{n=1,\ldots, t} \bigl( \tau_{k_d}n-N_{k_d}(n)\bigr)_+
+\tau_{k_d}\bigl(t -m_t^{(j)} \bigr) \\
+& \#\bigl\{
n=m_t^{(j)}+1,\ldots, t:\, \tau_{k_d}n>N_{k_d}(n-1), \,
\tau_{k_m}n\leq N_{k_m}(n-1) , \, m=1,\ldots, d-1, \\
&\hskip 4.5in   \text{arm $k_d$ not pulled} \bigr\},
\end{align*}
and the final bound is clearly valid even if
$m_T^{(j,d)}=m_T^{(j)}$. Substituting this bound into
\eqref{e:count.1} we obtain 
\begin{align*}
&\#\bigl\{
n=m_t^{(j)}+1,\ldots, t:\, \tau_{k_d}n>N_{k_d}(n-1) \ \text{for some}
\ d=1,\ldots, j-1\bigr\}\\
\leq&  \bigl( t-m_t^{(j)} \bigr)\sum_{d=1}^{j-1}\tau_{k_d}
+ \sum_{d=1}^{j-1}\max_{t' =1,\ldots, t} \bigl(
\tau_{k_d}t'-N_{k_d}(t')\bigr)_+\\
+& \sum_{d=1}^{j-1} \#\bigl\{
n=m_t^{(j)}+1,\ldots, t:\, \tau_{k_d}n>N_{k_d}(n-1), \,
\tau_{k_m}n\leq N_{k_m}(n-1) , \, m=1,\ldots, d-1, \\
&\hskip 4.5in   \text{arm $k_d$ not pulled} \bigr\},
\end{align*}
Substituting this bound into \eqref{e:split.deficit} gives us
\begin{align} \label{e:split.deficit.2}
& \bigl(   \tau_{k_j}t -N_{k_j}(t)\bigr)_+ \leq \tau_{k_j}\\
\notag  +& \sum_{d=1}^{j-1}\max_{t' = 1,\ldots, t} \bigl(
\tau_{k_d}t' - N_{k_d}(t')\bigr)_+ - \bigl( t-m_t^{(j)}
\bigr)\left(1-\sum_{d=1}^{j}\tau_{k_d}\right) \\
\notag  +& \sum_{d=1}^{j} \#\bigl\{
n=m_t^{(j)}+1,\ldots, t:\, \tau_{k_d}n>N_{k_d}(n-1),     \\
\notag  & \hskip 1in   \tau_{k_m}n\leq N_{k_m}(n-1) , \, m=1,\ldots,
d-1, \ \text{arm $k_d$ not pulled} \bigr\} \\
\notag \leq&   \tau_{k_j} +   \sum_{d=1}^{j-1}\max_{t' = 1,\ldots, t} \bigl(
\tau_{k_d}t' - N_{k_d}(t')\bigr)_+\\
\notag  +& \sum_{d=1}^{j} \#\bigl\{
n = 1,\ldots, t:\, \tau_{k_d}n>N_{k_d}(n-1),     \\
\notag  & \hskip 1in   \tau_{k_m}n\leq N_{k_m}(n-1) , \, m=1,\ldots,
d-1, \ \text{arm $k_d$ not pulled} \bigr\}.
\end{align}

Taking the maximum over $t$ on both sides of above inequality, we then
have
\begin{align} \label{e:split.deficit.3}
&\max_{t=1,\ldots, T} \bigl(
\tau_{k_j}t-N_{k_j}(t)\bigr)_+
\leq   \tau_{k_j} +   \sum_{d=1}^{j-1}\max_{t=1,\ldots, T} \bigl(
\tau_{k_d}t-N_{k_d}(t)\bigr)_+\\
\notag  +& \sum_{d=1}^{j} \#\bigl\{
n=1,\ldots, T:\, \tau_{k_d}n>N_{k_d}(n-1),     \\
\notag  & \hskip 1in   \tau_{k_m}n\leq N_{k_m}(n-1) , \, m=1,\ldots,
d-1, \ \text{arm $k_d$ not pulled} \bigr\}.
\end{align}
Therefore, we arrive at 
\begin{align}\label{e:deficit}
&
\mathbb E \left(\max_{t=1,\ldots, T} \bigl(
\tau_{k_j}t-N_{k_j}(t)\bigr)_+\right)\\
\notag \leq& \tau_{k_j}+  \mathbb E \left(\sum_{d=1}^{j-1}\max_{t=1,\ldots, T} \bigl(
\tau_{k_d}t-N_{k_d}(t)\bigr)_+\right)\\
\notag +&\sum_{d=1}^{j} \mathbb E \biggl( \sum_{n=1}^T \one\bigl(
\tau_{k_d}n>N_{k_d}(n-1),     \\
\notag  & \hskip 1in   \tau_{k_m}n\leq N_{k_m}(n-1) , \, m=1,\ldots,
d-1, \ \text{arm $k_d$ not pulled} \bigr)\biggr).
\end{align}

We will prove that for $k_d\in \calA_{\rm opt} \cup \calA_{\rm
	cr}$
\begin{align}\label{e:b}
& \mathbb E \biggl( \sum_{n=1}^T \one\bigl(
\tau_{k_d}n>N_{k_d}(n-1),     \\
\notag  & \hskip 1in   \tau_{k_m}n\leq N_{k_m}(n-1) , \, m=1,\ldots,
d-1, \ \text{arm $k_d$ not pulled} \bigr)\biggr)\\
\notag \leq& b_d\log T +O(1)
\end{align}
for $b_d>0$ that we will compute. It is elementary that $k_j\in
\calA_{\rm opt} \cup \calA_{\rm   cr}$ implies $k_d\in \calA_{\rm opt}
\cup \calA_{\rm   cr}$ for $d=1,\ldots, j-1$. Therefore, it will
follow from \eqref{e:b}, \eqref{e:deficit} and a simple inductive
argument that for any $k_j\in
\calA_{\rm opt} \cup \calA_{\rm   cr}$, 
\begin{align}\label{e:a}
\mathbb E \left(\max_{t=1,\ldots, T} \bigl(
\tau_{k_j}t-N_{k_j}(t)\bigr)_+\right) \leq a_j\log T +O(1)
\end{align}
with $a_1=b_1$ and for $j>1$,
$$
a_j=\sum_{d=1}^{j-1} a_d +\sum_{d=1}^j b_d, 
$$
which means that
\begin{equation} \label{e:induction}
a_j=\sum_{d=1}^j (j-d+1)b_d.
\end{equation}

We now prove \eqref{e:b}. We have
\begin{align*}
&E_\pi \biggl( \sum_{n=1}^T \one\bigl(
\tau_{k_d}n>N_{k_d}(n-1),    \tau_{k_m}n\leq N_{k_m}(n-1) , \, m=1,\ldots,
d-1, \ \text{arm $k_d$ not pulled} \bigr)\biggr)\\
=&\sum_{m=1}^{d-1} E_\pi \biggl( \sum_{n=1}^T \one\bigl(
\tau_{k_d}n>N_{k_d}(n-1),  \tau_{k_m}n\leq N_{k_m}(n-1), \
\text{arm $k_m$ is pulled at time $n$}\bigr)\biggr)\\
+&\sum_{m=d+1}^{K} E_\pi \biggl( \sum_{n=1}^T \one\bigl(
\tau_{k_d}n>N_{k_d}(n-1),  \
\text{arm $k_m$ is pulled at time $n$}\bigr)\biggr). 
\end{align*}
Observe that a ``no-tie'' assumption imposed at the beginning of the
section implies that 
$$
\mu_{k_d}+A_{k_d}>\mu_*\geq \mu_{k_m}.
$$
Therefore, we can use once again the usual UCB-type argument to see
that for any $m=1,\ldots, d-1$, for any $B>0$,
\begin{align*}
&E_\pi \biggl( \sum_{n=1}^T \one\bigl(
\tau_{k_d}n>N_{k_d}(n-1),  \tau_{k_m}n\leq N_{k_m}(n-1), \
\text{arm $k_m$ is pulled at time $n$}\bigr)\biggr)\\
\leq& B\log T + \sum_{n=1}^T P_\pi\bigl( N_{k_m}(n-1)>B\log T, \\
&\hskip 1.5in \tau_{k_d}n>N_{k_d}(n-1),  \tau_{k_m}n\leq N_{k_m}(n-1), \
\text{arm $k_m$ is pulled at time $n$}\bigr)\\
\leq& B\log T + \sum_{n=1}^T P_\pi\bigl( N_{k_m}(n-1)>B\log T,   \\
&\hskip 1.5in \tau_{k_d}n>N_{k_d}(n-1),  \tau_{k_m}n\leq
N_{k_m}(n-1), i_{k_m}(n)\geq i_{k_d}(n)\bigr)   \\
\leq& B\log T + \sum_{n=1}^T P_\pi\left( N_{k_m}(n-1)>B\log T, \, \hat
m_{k_m}(n-1)+    \sqrt{  \frac{2\log n}{N_{k_m}(n-1)}} \right.\\
&\left. \hskip 3in \, \geq \hat m_{k_d}(n-1)+ A_{k_d} +
\sqrt{ \frac{2\log n}{N_{k_d}(n-1)}}\right).
\end{align*}
By carefully choosing
$$
B= \frac{8}{(\mu_{k_d}+A_{k_d}-\mu_{k_m})^2},
$$
we obtain the bound
\begin{align} \label{e:high.m}
&E_\pi \biggl( \sum_{n=1}^T \one\bigl(
\tau_{k_d}n>N_{k_d}(n-1),  \tau_{k_m}n\leq N_{k_m}(n-1), \
\text{arm $k_m$ is pulled at time $n$}\bigr)\biggr)\\
\notag \leq& \frac{8}{(\mu_{k_d}+A_{k_d}-\mu_{k_m})^2}\log T+O(1),
\end{align} 
$m=1,\ldots, d-1$. The same argument shows that for every
$m=d+1,\ldots, K$,
\begin{align} \label{e:low.m}
&E_\pi \biggl( \sum_{n=1}^T \one\bigl(
\tau_{k_d}n>N_{k_d}(n-1),  \
\text{arm $k_m$ is pulled at time $n$}\bigr)\biggr) \\
\notag \leq& \frac{8}{(\mu_{k_d}+A_{k_d}-\mu_{k_m}-A_{k_m})^2}\log
T+O(1). 
\end{align} 
Now \eqref{e:high.m} and \eqref{e:low.m} imply \eqref{e:b} with
\begin{equation} \label{e:b.val}
b_d=\sum_{m=1}^{d-1} \frac{8}{(\mu_{k_d}+A_{k_d}-\mu_{k_m})^2} +
\sum_{m=d+1}^K \frac{8}{(\mu_{k_d}+A_{k_d}-\mu_{k_m}-A_{k_m})^2}.
\end{equation}
Now it follows from \eqref{e:b.val} and \eqref{e:induction} that for
every $j$ such that $k_j\in
\calA_{\rm opt} \cup \calA_{\rm   cr}$,
\begin{equation} \label{e:a.val}
a_j= 8\sum_{d=1}^j (j-d+1) \left( \sum_{m=1}^{d-1} \frac{1}{(\mu_{k_d}+A_{k_d}-\mu_{k_m})^2} +
\sum_{m=d+1}^K
\frac{1}{(\mu_{k_d}+A_{k_d}-\mu_{k_m}-A_{k_m})^2}\right). 
\end{equation} 
We conclude by \eqref{e:a} that every $j$ such that $k_j\in
\calA_{\rm opt} \cup \calA_{\rm   cr}$,
\begin{align} \label{e:deficit.final.b}
E_\pi  \bigl(
\tau_{k_j}T-N_{k_j}(T)\bigr)_+ \leq a_j\log T +O(1),
\end{align}
with $a_j$ given in \eqref{e:a.val}.

%

\vspace{0.2in}

\noindent \textbf{Remark}. 
In the proof, we assume that there is no tie, i.e., $A_{k_{j_1}} + \mu_{k_{j_1}} \neq A_{k_{j_2}} + \mu_{k_{j_2}}$ for any $j_1 \neq j_2 \in [K]$. 
This assumption is not restrictive since the probability that event ``$A_{k_{j_1}} + \mu_{k_{j_1}} \neq A_{k_{j_2}} + \mu_{k_{j_2}}$ for some $j_1 \neq j_2 \in [K]$." is zero when we pick penalty rates $A_k$'s uniformly randomly.

\newpage

\section{Proof of Gap-independent Upper Bounds}\label{app:ind:upper}

\textbf{Proof of Lemma \ref{lem:independent}}
We first prove that the algorithm pulls arm $k'$ with $A_{k'} - \Delta_{k'} \leq \frac{1}{2}(A_k - \Delta_k)$ at time $n$ when $N_{k'}(n) \geq \frac{32 \log T}{(A_k - \Delta_k)^2}$ and $N_k(n) < \tau_k n$ with vanishing probability.
This is because
\begin{eqnarray}
& & \mathbb P(A_k + m_k(n) + \sqrt{\frac{2 \log n}{N_k(n)}} \leq 
A_{k'} + m_{k'}(n) + \sqrt{\frac{2\log n}{N_{k'}(n)}}) \nonumber \\
&\leq& \mathbb P(A_k + \mu_k \leq A_k' + \mu_k' + 2 \sqrt{\frac{2\log n}{N_{k'}(n)}}) + \frac{2}{n^2} \nonumber \\
&\leq& \mathbb P(A_k - \Delta_k \leq A_k' - \Delta_{k'} + 2 \sqrt{\frac{2\log n}{N_{k'}(n)}}) + \frac{2}{n^2} \nonumber \\
&=& 2 / n^2.
\end{eqnarray} 
Next, we say arm $k'$ is a \textit{very critical} arm if arm $k'$ satisfies $A_{k'} - \Delta_{k'} \geq \frac{1}{2}(A_k - \Delta_k)$. Otherwise $k'$ is a \textit{non-very critical} arm. In other words, each non-very critical arm can be only played at most $\frac{32 \log T}{(A_k - \Delta_k)^2}$ times with high probability. 

Furthermore, we can show that $N_{k'}(n) \geq \frac{32 \log T}{(A_k - \Delta_k)^2}$ at time $n = c_0 T / 2$ for each very critical arm $k'$. If not, note that $\frac{32 \log T}{(A_k - \Delta_k)^2} \leq \tau_{k'} c_0^2 T / 4$, then $N_{k}'(n) < \tau_k' n$ for any $n \in \{\lceil c_0^2 T/4 \rceil, \ldots, \lfloor c_0 T / 2 \rfloor \}$. Hence, for any arm $k''$ can be played at most $\max\{\tau_{k''} c_0 T/2, T^{2/3}\}$ times between rounds $c_0^2 T/4$ and $c_0 T/2$.
Then, we must have 
\[c_0 T/2 - c_0^2 T/4 \leq \sum_{k'} \tau_{k'} c_0 T/2 + \sum_{k'} {\frac{32 \log T}{(A_k - \Delta_k)^2}}.\]
However, the above inequality fails to hold when $T$ is large enough and $ (A_k - \Delta_k) \geq \sqrt{\frac{32 K \log T}{T}}$. This leads to the contradiction.
Thus, we have $N_{k'}(n) \geq \frac{32 \log T}{(A_k - \Delta_k)^2}$ for any very critical arm $k'$ at time $n = c_0 T / 2$.

This further gives us that we must have $N_{k'}(n) \geq \lfloor \tau_{k'} n \rfloor$ for all very critical arms at some time $n \in [c_0 T, T]$. 
To prove this,
we observe the fact that for any arm $\bar k$, it will be played with probability less than $\frac{2}{T^2}$ at time $n$ once $N_{\bar k}(n) \geq \max\{\tau_k n, \frac{32 \log T}{(A_k - \Delta_k)^2}\}$ and one critical arm $k'$ is played less than $\tau_{k'} n$. 
(In other words, this tells us that once arm $\bar k$ has been played $\max \{\tau_{\bar k} n, \frac{32 \log T}{(A_k - \Delta_k)^2}\}$ times, then it can only be played at time when all very critical arms $k'$s have been played for $\tau_{k'} n$ times or $\lfloor \tau_{\bar k} n \rfloor$ jumps by one with probability greater than $1 - 2 / n^2$.)

Let $n_1 (\geq c_0 T / 2)$ be the first ever time such that 
$N_{k'}(n_1) \geq \lfloor \tau_{k'} n_1 \rfloor$.
By straightforward calculation, it gives that $n_1$ must be bounded by 
\[n_1 \leq c_0 T / 2 + \sum_{k^{''}: \text{non-very critical}} \frac{32 \log T}{(A_k - \Delta_k)^2} + (\sum_{k'} \tau_{k'}) T \leq (c_0 + \sum_{k'} \tau_{k'}) T\]
with probability greater than $1 - 2 K/T$.

That is, $n_1$ is well defined between $c_0 T / 2$ and $T$. At time $n_1$, we have all very critical arms $k'$ such that $N_{k'}(n_1) \geq \tau_{k'} n_1$.
Therefore, starting from time $n_1$, the maximum difference between any non-fairness level $(\tau_{k'} n - N_{k'}(n))_{+}$'s with $k'$ in the set of very-critical arms is always bounded by 1 with probability $1 - 2K/T$ for all $n \in [n_1, T]$.

Lastly, suppose $n_2$ be the last time that arm $k$ is above fairness level.
We know at time $n = n_2$, each very critical arm $k'$ is played for at least $\tau_{k'} n_2 - 1$.
by previous argument. 
Then in the remaining $T - n_2$ rounds, we know that each very critical arm is played at most $ \tau_{k'}T - \tau_{k'}n_2 + 1$.
Then we must have 
$$T - n_2 \leq  (\sum_{k': \text{very critical}} \tau_{k'}) (T - n_2) + K + \sum_{k: \text{non-very critical}} \frac{32 \log T}{(A_k - \Delta_k)^2},$$
which implies $T - n_2 \leq (K \frac{32 \log T}{(A_k - \Delta_k)^2} + K)/c_0$.
This finally implies that $N_{k}(T) \geq N_{k}(n_2) \geq \tau_k T - \tau_k (K \frac{32 \log T}{(A_k - \Delta_k)^2} + K)/c_0 - 1$ with probability at least $1 - 2 K / T$.
That is, $\mathbb E[(\tau_k T - N_k(T))_{+}] = \tau_k(K \frac{32 \log T}{(A_k - \Delta_k)^2} + K)/c_0 + 1 = O(\tau_k K \frac{32 \log T}{(A_k - \Delta_k)^2})$.

We prove the gap-independent upper bound (Theorem \ref{thm:indep:upper}) by considering the following situations.

\noindent \textbf{Situation 1.a} 
For arm $k \in \mathcal A_{\text{non-cr}}$ and $\Delta_k \leq 4 \sqrt{\frac{\log T}{T}}$,
the regret on arm $k$ is upper bounded by 
\begin{eqnarray}
(\Delta_k - A_k) (\tau_k T - N_k(T))
\end{eqnarray}
if $0 \leq N_k(T) \leq \tau_k T$; or bounded by
\begin{eqnarray}
\Delta_k (N_k(T) - \tau_k T) + (\Delta_k - A_k)\tau_k T
\end{eqnarray}
if $N_k(T) \geq \tau_k T$.

\noindent \textbf{Situation 1.b} 
For arm $k \in \mathcal A_{\text{non-cr}}$ and $\Delta_k > 4 \sqrt{\frac{\log T}{T}}$,
\begin{itemize}
	\item
	if $\Delta_k - A_k > 4 \sqrt{\log T / \tau_k T}$
	the regret on arm $k$ is upper bounded by 
	\begin{eqnarray}
	(\Delta_k - A_k) (\frac{8 \log T}{(\Delta_k - A_k)^2} + O(1)). 
	\end{eqnarray} 
	\item if $\Delta_k - A_k \leq 4 \sqrt{\log T / \tau_k T}$
	the regret on arm $k$ is upper bounded by 
	\begin{eqnarray}
	(\Delta_k - A_k) \tau_k T + \Delta_k [(\frac{8 \log T}{\Delta_k^2} - \tau_k T)_{+} + O(1)]. 
	\end{eqnarray} 
\end{itemize}

In other words, for any arm $k \in \mathcal A_{\text{non-cr}}$, its regret is always bounded by
\begin{eqnarray}
4 \sqrt{\frac{\log T}{T}} \tau_k T + 4 \sqrt{\tau_k T\log T} + 4 \sqrt{\frac{\log T}{T}}(N_k(T) - \tau_k T)_{+}.
\end{eqnarray}

\noindent \textbf{Situation 2} 
We then split set $\mathcal A_{\text{opt}} \cup \mathcal A_{\text{cr}}$ into two subsets, 
$\mathcal A_{\text{cr, large}}$ and $\mathcal A_{\text{cr,small}}$,
where 
\[\mathcal A_{\text{cr, large}} := \{k: A_k - \Delta_k >  \sqrt{\frac{32 K \log T}{T}} \}\]
and 
\[\mathcal A_{\text{cr, small}} := \{k: A_k - \Delta_k \leq \sqrt{\frac{32 K \log T}{T}} \} \}.\]

For arm $k \in \mathcal A_{\text{cr,large}}$, we have 
$\mathbb E[(\tau_k T - N_k(T))_{+}] = O(\tau_k K \frac{32 \log T}{(A_k - \Delta_k)^2})$ by Lemma \ref{lem:independent}. 
The regret on arm $k$ is then bounded by 
\begin{eqnarray}
& & \Delta_k \mathbb E[N_k(T) - \tau_k T] + A_k \mathbb E[(\tau_k T - N_k(T))_{+}] \nonumber \\
&\leq& \max\{\Delta_k \min\{\frac{8 \log T}{\Delta_k^2} - \tau_k T, N_k(T) - \tau_k T\}, (A_k - \Delta_k) \mathbb E[(\tau_k T - N_k(T))_{+}] \} \nonumber \\
&\leq& \max\{\Delta_k \min\{\frac{8 \log T}{\Delta_k^2} - \tau_k T, N_k(T) - \tau_k T\}, \tau_k \sqrt{32 K T \log T} \}.
\end{eqnarray}

For arm $k \in \mathcal A_{\text{cr,small}}$, the regret on arm $k$ is then bounded by
\begin{eqnarray}
(A_k - \Delta_k) (\tau_kT - N_k(T)) \leq \tau_k \sqrt{32 K T \log T}
\end{eqnarray}
if $0 \leq N_k(T) \leq \tau_k T$,
or
\begin{eqnarray}
\Delta_k \min\{\frac{8 \log T}{\Delta_k^2} - \tau_k T + O(1), N_k(T) - \tau_k T\} 
\end{eqnarray}
if $N_k(T) \geq \tau_k T$.

In summary, for any arm $k \in \mathcal A_{\text{opt}} \cup \mathcal A_{\text{cr}}$, 
\begin{eqnarray}
\Delta_{k} \min\{\frac{8 \log T}{\Delta_k^2} - \tau_k T, N_k(T) - \tau_k T\} + \tau_k \sqrt{32 K T \log T}.
\end{eqnarray}

Combining above situations, the total regret is upper bounded by
\begin{eqnarray}
& & \sum_{k \in \mathcal A_{\text{non-cr}}} \max\{8 \sqrt{\tau_k T \log T} + 4 \sqrt{\frac{\log T}{T}}(N_k(T) - \tau_k T)_{+} \} \nonumber \\
& & + \sum_{k \in \mathcal A_{\text{opt}} \cup \mathcal A_{\text{cr}}} 
\Delta_{k} \min\{\frac{8 \log T}{\Delta_k^2} - \tau_k T, N_k(T) - \tau_k T\} + \tau_k \sqrt{32 K T \log T} \nonumber \\
&\leq & 8 \sqrt{T \log T} (\sum_{k \in \mathcal A_{\text{non-cr}}} \sqrt{\tau_k}) + \sqrt{32 K T \log T} (\sum_{k \in \mathcal A_{\text{cr}} \cup \mathcal A_{opt}} \tau_k)
+  4 \sqrt{\frac{\log T}{T}} \sum_{k \in \mathcal A_{\text{non-cr}}}(N_k(T) - \tau_k T)_{+} \nonumber \\
& & + \sum_{k \in \mathcal A_{\text{opt}} \cup \mathcal A_{\text{cr}}} 
\sqrt{8 (N_k(T) - \tau_k T)_{+} \log T} \nonumber \\
&\leq& 8 \sqrt{T \log T} (\sum_k \sqrt{\tau_k}) + \sqrt{32 K T \log T}
+ 4 \sqrt{\frac{\log T}{T}} (1 - \tau_{min}) T
+ \sqrt{8 \log T} \sqrt{K T(1 - \tau_{min})} \nonumber \\
& & \label{ineq:tau},
\end{eqnarray}
where \ref{ineq:tau} uses the fact that $\sum_{k \in \mathcal A_{\text{cr}} \cup \mathcal A_{opt}} \tau_k \leq \sum_{k} \tau_k \leq 1$;
$\sum_{k \in \mathcal A_{\text{non-cr}}}(N_k(T) - \tau_k T)_{+} \leq T(1 - \tau_{min})$ 
and
\begin{eqnarray*}
	& & \sum_{k \in \mathcal A_{\text{opt}} \cup \mathcal A_{\text{cr}}} 
	\sqrt{(N_k(T) - \tau_k T)_{+}} \leq \sum_{k}
	\sqrt{(N_k(T) - \tau_k T)_{+}} \leq \sqrt{K \sum_{k} (N_k(T) - \tau_k T)_{+}} \leq \sqrt{K T (1 - \tau_{min})}
\end{eqnarray*}
by Jensen's inequality.

\section{Proof of Gap-independent Lower Bounds}\label{app:ind:lower}

Consider a $K$-arm setting with $\mu_2 = \mu_3 = \ldots = \mu_K = 0$, $\mu_1 = \Delta$ ($0 < \Delta < 1/2$), $A_1, A_2, \ldots, A_K > 0$, $\Delta < A_k$ for $k \in [K]$, $\tau_1, \tau_2, \ldots, \tau_K \in [0,1]$. 

Since $\sum_{k = 2}^T N_k(T) \leq T$, then it holds 
$\mathbb E_{\pi} [N_{k_1}(T)] \leq T/(K-1)$ with 
$k_1 = \arg\min_{k > 1} \mathbb E_{\pi}[N_k(T)]$ for any policy $\pi$.
We then construct another $K$-arm setting with $\mu_{k_1} = 2 \Delta$ and all other parameters remain the same.

For policy $\pi$, the regret of the first setting is 
\[R_{1,\pi}(T) \geq A \mathbb E[(\tau_1 T - N_1(T))_{+}] 
+ \sum_{k \neq 1}\{ \Delta \mathbb E[N_{k}(T) - \tau_{k} T] + A \mathbb E(\tau_{k} T - N_{k}(T))_{+}\}\]
and the regret of the second setting is
\[R_{2,\pi}(T) \geq A \mathbb E[(\tau_{k_1} T - N_{k_1}(T))_{+}] 
+ \{ \Delta \mathbb E[N_1(T) - \tau_1 T] + A \mathbb E(\tau_1 T - N_1(T))_{+}\}\]

If $N_1(T) < (1 + \tau_1 - \sum_{k \neq 1}\tau_{k}) T/2$, then $R_{1,\pi}(T) \geq \Delta \frac{1 - \sum_k \tau_k}{2} T$.
While $N_1(T) > (1 + \tau_1 - \sum_{k \neq 1}\tau_{k})/2$, then $R_{2,\pi}(T) \geq \Delta \frac{1 - \sum_{k \neq 1}\tau_{k}}{2} T$.
In other words, for policy $\pi$,
\begin{eqnarray}
\text{worst regret} &\geq& \frac{1}{2}(R_{1,\pi}(T) + R_{2,\pi}(T)) \nonumber \\
&\geq& \frac{1}{2}(\Delta T \frac{1 - \sum_{k}\tau_{k}}{2} \mathbb P_1(N_1(T) < \frac{1 + \tau_1 - \sum_{k \neq 1}\tau_{k}}{2}) \nonumber \\
& & + \Delta T \frac{1 - \sum_{k}\tau_{k}}{2} \mathbb P_2(N_1(T) \geq \frac{1 + \tau_1 - \sum_{k \neq 1}\tau_{k}}{2})) \nonumber \\
&\geq& \frac{(1 - \sum_{k}\tau_{k}) \Delta T}{8} \exp\{-KL(P_1 \| P_2)\} \label{eq:kl} \\
&\geq& \frac{(1 - \sum_{k}\tau_{k}) \Delta T}{8} \exp\{- C T \Delta^2 / (K-1)\} \label{eq:gauss},
\end{eqnarray}
where $P_1$ and $P_2$ are two probability distributions under two settings associated with policy $\pi$; \ref{eq:kl} follows from the Bretagnolle–Huber inequality.  
Inequality \ref{eq:gauss} holds since KL-divergence $KL(P_1 \| P_2) \leq C T \Delta^2 / (K-1)$ for many probability distributions.
(E.g. $C = 1/2$ if the reward of each arm follows Gaussian distribution with variance 1.)

Taking $\Delta = \sqrt{\frac{K-1}{CT}}$, we have 
\begin{eqnarray}
\text{worst regret}
&\geq& \frac{(1 - \sum_{k}\tau_{k}) \Delta T}{8} \exp\{- C T \Delta^2 / (K-1)\} \nonumber \\
&\geq& \frac{(1 - \sum_{k}\tau_{k}) \sqrt{(K-1)T/C}}{8 e},
\end{eqnarray}
where $e = \exp\{1\}$. This completes the proof of Theorem \ref{thm:indep:lower}.

\end{document}